# Model validation in machine learning: A scenario-based guide from hold-out splits to nested group cross-validation in biomedical and applied research

Mehmet Baygin[1], Sengul Dogan[2], Turker Tuncer[2],

[1]Department of Computer Engineering, Faculty of Engineering and Architecture, Erzurum Technical University, Erzurum, Türkiye

mehmet.baygin@erzurum.edu.tr

[2]Department of Digital Forensics Engineering, Technology Faculty, Firat University, Elazig, Türkiye

sdogan@firat.edu.tr; turkertuncer@firat.edu.tr

**Abstract**

Model validation estimates the performance of a complete learning procedure on new data. However, an invalid split can produce an optimistic and stable result. This tutorial reviews hold-out validation, train/validation/test designs, repeated random subsampling, k-fold and repeated stratified cross-validation, leave-one-out and leave-p-out schemes, group-aware validation, and nested group cross-validation. General machine-learning principles are linked to EEG epochs, paired-eye OCT images, repeated clinical measurements, and multicenter data. Eight controlled scenarios compare flawed and leakage-safe designs: seven use locked confusion matrices with auditable metrics, and one uses a reproducible repeated-study simulation. The scenarios cover global feature selection, normalization leakage, dependent records, center mixing, repeated test-set use, and estimator instability. Bias, variance, metric aggregation, uncertainty, and computational cost are also examined. A data-size matrix, a decision tree, and reporting checklists are provided. Reproducible MATLAB templates and scikit-learn counterparts are included. The results show that no validation method is universally best. The independent unit must match the intended deployment target. Every data-dependent operation must also exclude the observations used for performance estimation.

**Keywords:** model validation; cross-validation; nested cross-validation; data leakage; group-aware validation; biomedical machine learning; confusion matrix; computational complexity; performance estimation.

## 1. Introduction

### 1.1. Validation as an estimate of future performance

Model validation estimates the performance of a complete learning procedure on new data. The assessment observations must not be used for model fitting or selection. Training performance cannot provide this estimate because the model has adapted to those observations. The target population, sampling process, and prediction task must be defined before the split. The predictive data-reuse principle established the basis of modern resampling [1, 2].

The term validation has several roles. Internal validation reorganizes one development dataset. External validation assesses transport to another population or setting. A subset used for early stopping or hyperparameter selection remains development data. A score used to choose a pipeline is also a selection criterion. Assessment information must not return to the learning process.

Generalization depends on the unit that will be new at deployment. An EEG epoch from a known subject is not equivalent to an epoch from a new subject. Two OCT eyes from one patient are also dependent. Mixed-center data cannot support a new-center claim unless centers are held out. Record-wise validation can overestimate new-subject performance. The independent unit must therefore match the deployment target [3, 4].

A validation estimate can be stable and still target the wrong task. Repeated sample-level cross-validation may return similar scores because subject-specific signals remain on both sides of each split. Repetition then reduces partition variation around an invalid target. It cannot convert a known-subject experiment into a new-subject experiment. The independent unit must be corrected before fold count, repetition, or classifier selection is considered.

### 1.2. Why optimistic estimates are common

Optimistic performance does not require misconduct or a defective classifier. It can arise when global feature screening or normalization precedes cross-validation, dependent records cross folds, or one test set guides repeated decisions. Prevalidation feature selection can create severe bias. Model selection can also overfit its own criterion. The resulting value may be computed correctly but answer an easier prediction question. Such leakage is a major obstacle to reproducible machine-learning research [5–9].

Small biomedical cohorts magnify these problems because a few cases can change performance substantially. Cross-validation error bars can be wide at neuroimaging sample sizes. Repetition describes partition sensitivity but creates neither new patients nor representativeness. Effective sample size follows independent subjects, patients, sessions, or centers rather than windows or slices. Thus, 600 EEG epochs from 30 subjects remain 30 independent units for a new-subject claim [10].

Transparent reporting is required. The independent unit, preprocessing scope, selection procedure, class counts, missing-data handling, and aggregation rule must be stated. Complete prediction-model reporting is supported by TRIPOD. Therefore, the statement that 10-fold cross-validation was used is insufficient. Folds must be reproducible, independence must be defensible, and every learned operation must be assigned to a training partition [11].

### 1.3. Scope and contributions

This article treats validation as the design of an estimator rather than the selection of a software option. The covered methods include hold-out, three-way splits, stratified and repeated hold-out, k-fold and repeated stratified k-fold, LOOCV, leave-p-out, subject- and group-aware schemes, and nested group cross-validation. Bootstrap estimators, out-of-bag error, specialized time-series resampling, and cross-dataset external validation are reserved for a short advanced-reading note. The emphasis is general machine learning with recurring examples from EEG, OCT, and hematology. Figure 1 organizes the included methods according to the unit and level at which data are withheld.

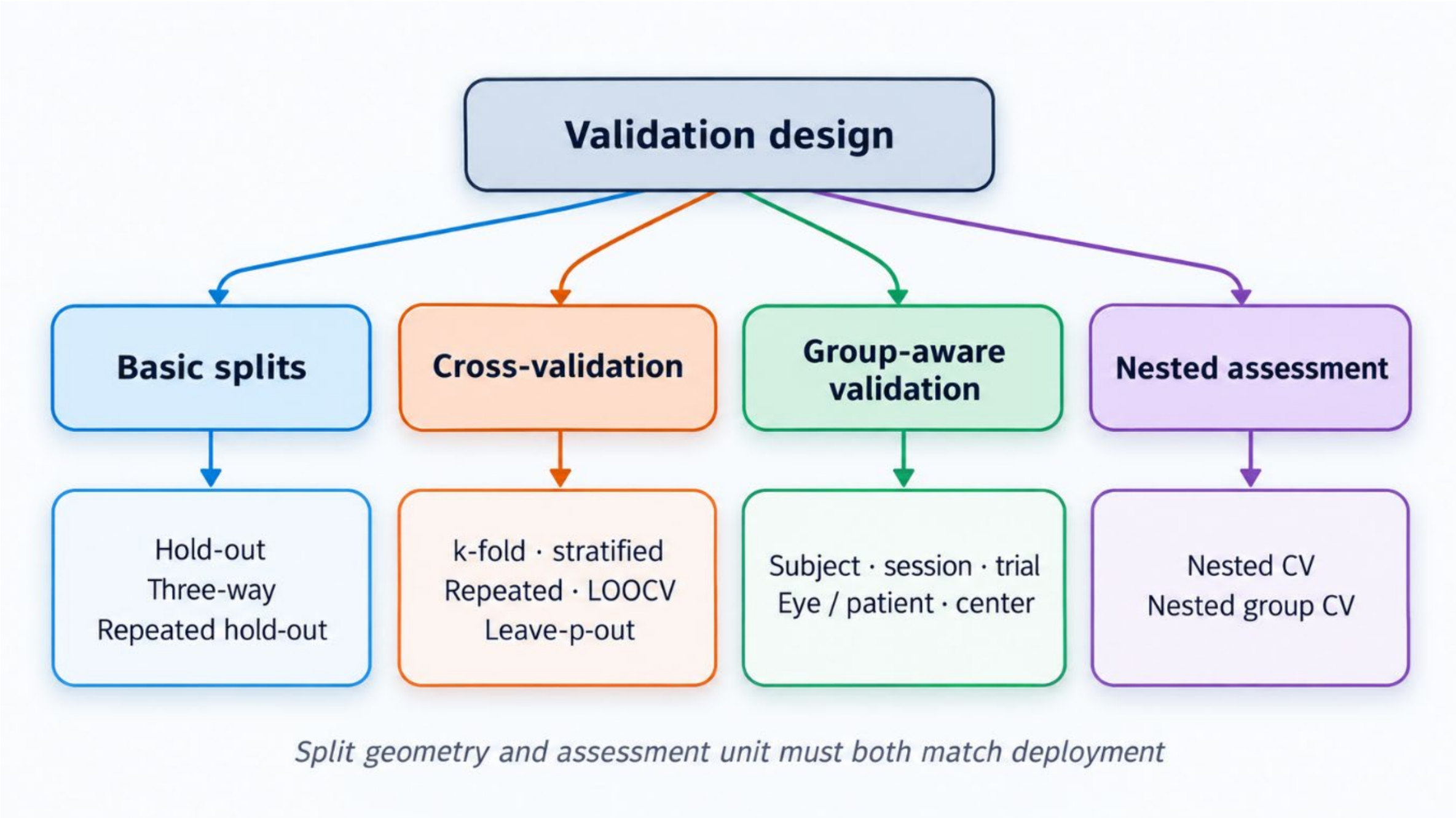

Figure 1. Overview of the validation designs discussed in this tutorial. Basic splits define one or more development-assessment partitions, cross-validation rotates the assessment subset, group-aware schemes preserve the chosen independent unit, and nested designs separate model selection from performance estimation. Bootstrap, out-of-bag, specialized time-series, and cross-dataset external validation are outside the main taxonomy and are mentioned only as advanced topics.

The taxonomy is built around two separate choices that are easy to mix together. One concerns the shape of the split: a single hold-out, several folds, one group at a time, or an inner/outer hierarchy. The other concerns what each column in that split represents: a record, subject, patient, session, device, or center. Even an elaborate fold structure is scientifically wrong if it is applied to the wrong unit. By contrast, a simple hold-out may be entirely defensible when it uses the right independent unit, includes enough cases, and is left untouched during development.

The four geometries may be compared before their statistical details are introduced. As shown in Figure 2, hold-out relies on one assessment block, k-fold rotates several blocks, LOOCV rotates single independent observations, and LOSO rotates complete subject groups. Rows represent split iterations and columns represent observations ordered by group.

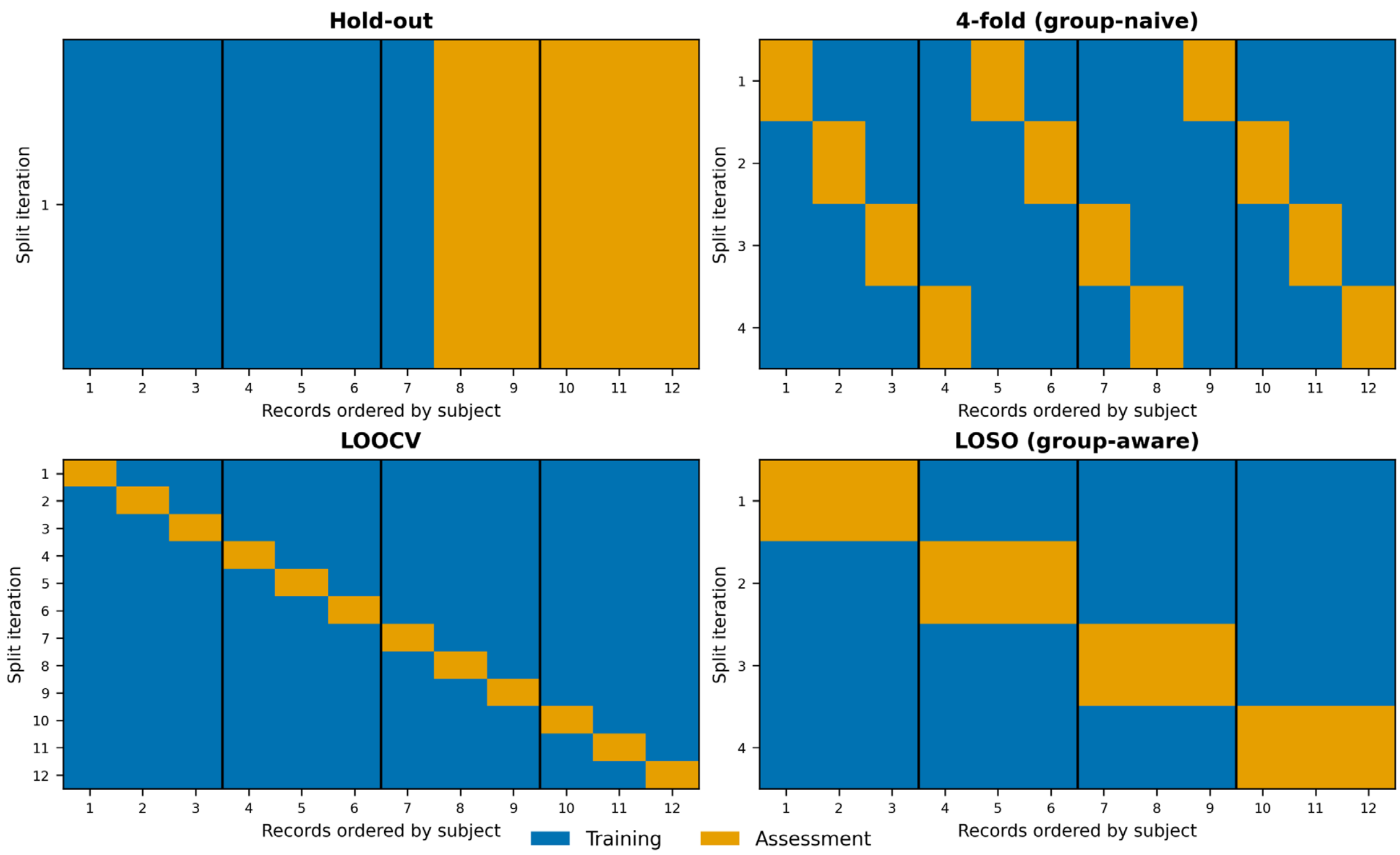

Figure 2. Schematic comparison of hold-out, k-fold, LOOCV, and LOSO. Rows represent split iterations, columns represent records ordered by subject, blue cells represent training membership, orange cells represent assessment membership, and thick vertical boundaries identify subjects. The hold-out block and the k-fold folds are assigned without reference to the subject identifier and therefore cross those boundaries; LOOCV withholds one record at a time and also crosses them. Only LOSO keeps every subject entirely on one side of the split, which is why it is the only panel that can support a new-subject claim.

Figure 2 makes the difference between LOOCV and LOSO visible. LOOCV withholds one independent observation, while LOSO withholds every record from one subject. They coincide only when each subject contributes a single row. With repeated measurements, leaving out one row may still leave information about that subject in training. The appropriate choice depends on what will be unfamiliar when the model is used.

Table 1 summarizes the validation methods considered in this article. The fit counts in the table refer to a fixed pipeline; a hyperparameter search adds extra fits unless that search is explicitly nested. The bias and variance columns should be interpreted as common tendencies of the performance estimator rather than guarantees. Their actual behavior still depends on the learner, loss function, dependence structure, and underlying data distribution.

Table 1. Practical overview of the validation designs discussed in this tutorial. N denotes the number of independent observations, G the number of groups, R the number of repetitions, K the ordinary fold count, K_o and K_i the outer and inner fold counts, and |Λ| the number of candidate hyperparameter configurations. The bias and variance entries describe the performance estimator for the stated target rather than the fitted learner in isolation.

| Method | Independent unit | Model fits | Typical estimator bias | Typical estimator variance |
|---|---|---|---|---|
| Hold-out | Record or chosen group | 1 | Can be appreciable because less data are used for training | High split sensitivity when the test set is small |
| Train/validation/test | Record or chosen group | At least 1 final fit; selection adds fits | Honest only if the test set remains untouched | Depends strongly on validation and test sizes |
| Stratified hold-out | Record or chosen group | 1 | Similar to hold-out | Usually lower class-composition variability |
| Repeated hold-out / Monte Carlo CV | Record or chosen group | $R$ | Depends on the training fraction | Lower descriptive split variability, but repetitions are dependent |
| k-fold CV | Independent unit | $K$ | Often decreases as $K$ increases because training folds are larger | Fold estimates are correlated; variance need not decrease monotonically |
| Stratified k-fold | Independent unit | $K$ | Similar to k-fold | Usually more stable for imbalanced classification |
| Repeated (stratified) k-fold | Independent unit | $RK$ | Similar target to its base k-fold design | Better description of partition sensitivity at added cost |
| LOOCV | Independent unit | $N$ | Usually low training-size bias | Can have high estimator variance; predictions are strongly coupled |
| Leave-p-out | Independent unit | $\binom{N}{p}$ | Depends on $p$ | Potentially stable when exhaustive, but often infeasible |
| LOSO / LOGO | Subject or group | $G$ | Targets new-group performance | Can be high when groups are heterogeneous or $G$ is small |
| GroupKFold | Subject or group | $K$ | Targets groups drawn from a similar group population | Depends on group balance and fold sizes |
| StratifiedGroupKFold | Indivisible group with class balancing | $K$ | Similar target to GroupKFold | Can reduce class imbalance, but perfect stratification may be impossible |

| Method | Independent unit | Model fits | Typical estimator bias | Typical estimator variance |
|---|---|---|---|---|
| Leave-one-session/trial/record-out | Session, trial, or recording | Number of groups | Targets the named acquisition unit | Reflects between-unit heterogeneity |
| Patient-level paired-eye split | Patient, with both eyes together | $K$ or number of patients | Targets new-patient performance | Wider uncertainty than an eye-level split with the same row count |
| Leave-one-site/center-out | Site or center | Number of sites | Estimates transport to an unseen acquisition environment | Very high uncertainty with few centers |
| Nested CV | Outer independent unit | K_o (K_i \|Λ\| + 1) | Reduces model-selection optimism | Higher variance and much higher compute |
| Nested group CV | Outer and usually inner groups | K_o (K_i \|Λ\| + 1) | Targets the selected group-aware pipeline | Sensitive to small numbers of groups and class infeasibility |

One lesson from Table 1 is that a larger number of fits should not be mistaken for a better estimate. Repetition can make the result less dependent on one arbitrary partition, but it cannot repair a split performed at the wrong unit. LOOCV trains on almost all observations in every iteration and can still be unstable in a small sample. Nested CV becomes useful when the learning procedure contains data-driven choices; for a completely prespecified model evaluated once on a sufficiently large untouched test set, that extra cost may add little. These are the distinctions developed more formally in the next section.

The central point is that validation estimates performance under a particular future-use condition; it is not a box to tick after model fitting. Decide first what will genuinely be new at deployment - a record, subject, patient, session, device, or center - and then choose the split. The term test should be reserved for data that play no role in model development.

## 2. Theoretical framework

### 2.1. Generalization error and empirical risk

A finite sample is all we observe from the population in which the model will eventually be used. Validation therefore asks how the learning procedure is likely to behave on new observations from that same population. We use the following notation: $X$ denotes the input, $Y$ the observed outcome, $P_\star$ the target joint distribution, $f$ a prediction rule, and $L$ the loss function. The generalization risk of $f$ is

$$\mathcal{R}_\star(f) = \mathbb{E}_{(X,Y)\sim P_\star}\big[L\big(Y, f(X)\big)\big] = \int L\big(y, f(x)\big)\, dP_\star(x,y). \tag{1}$$

Equation (1) writes $\mathcal{R}_\star(f)$ as the average loss under the deployment distribution. A pair $(X, Y)$ is drawn from $P_\star$; $f(X)$ is the model output, and $L\big(Y, f(X)\big)$ measures the resulting loss. Under zero-one loss, this is the misclassification probability; under squared loss, it is the expected squared prediction error. The estimate is meaningful only when the held-out observations reasonably represent draws from $P_\star$.

Since $P_\star$ cannot be observed in full, we work with the finite sample $D_N = \{(x_i, y_i)\}_{i=1}^N$ as its empirical counterpart. For a fixed rule $f$, the empirical risk is

$$\hat{\mathcal{R}}_N(f; D_N) = \frac{1}{N}\sum_{i=1}^{N} L\left(y_i, f(x_i)\right). \tag{2}$$

In Equation (2), $N$ counts the sampled independent units, $(x_i, y_i)$ is unit $i$, and $\hat{\mathcal{R}}_N$ is the average observed loss. If $f$ is fitted or chosen using that same $D_N$, the training risk is typically optimistic because the rule has already adapted to those data. Keeping fitting and scoring separate avoids this direct reuse. In practice, however, we usually evaluate more than a fixed rule $f$; the object of interest is the full learning procedure $\mathcal{A}$, which maps a training set to the fitted rule $\hat{f}_D = \mathcal{A}(D)$.

A stored row is not necessarily an independent unit. Twenty adjacent EEG epochs, for instance, do not provide twenty independent examples of unseen subjects. The same issue arises with two OCT eyes from one patient and with repeated hematology panels from the same person. The $N$ used in uncertainty calculations should therefore refer to the unit over which the study is intended to generalize.

### 2.2. Bias–variance–noise decomposition

For squared-error prediction, the familiar bias-variance-noise decomposition is a useful way to interpret expected test error. Suppose the regression model is $Y = f_\star(X) + \varepsilon$, where $f_\star(x) = \mathbb{E}[Y \mid X = x]$, $\mathbb{E}[\varepsilon \mid X] = 0$, and $\mathrm{Var}(\varepsilon \mid X = x_0) = \sigma^2(x_0)$. At a given input $x_0$, a learner trained on random datasets $D$ has expected test error

$$\mathbb{E}_{D,\varepsilon}\left[\left(Y_0 - \hat{f}_D(x_0)\right)^2\right] = \sigma^2(x_0) + \left(\mathbb{E}_D[\hat{f}_D(x_0)] - f_\star(x_0)\right)^2 + \mathrm{Var}_D\left(\hat{f}_D(x_0)\right). \tag{3}$$

Here, $Y_0$ is a future outcome at $x_0$, $\hat{f}_D$ is the model fitted from dataset $D$, and $f_\star$ is the conditional mean. The first term reflects variation in $Y$ that remains even when the mean is known; the second captures systematic error across training samples; and the third describes how fitted predictions change from one training sample to another. This additive decomposition is specific to squared loss. Classification has analogous ideas, but not a single universal three-term identity.

Figure 3 is intentionally schematic. It illustrates the trade-off rather than reporting an empirical result, so the curves should not be read as universal. The learner's bias and variance are shown separately from the bias and variance of the validation estimate, discussed next.

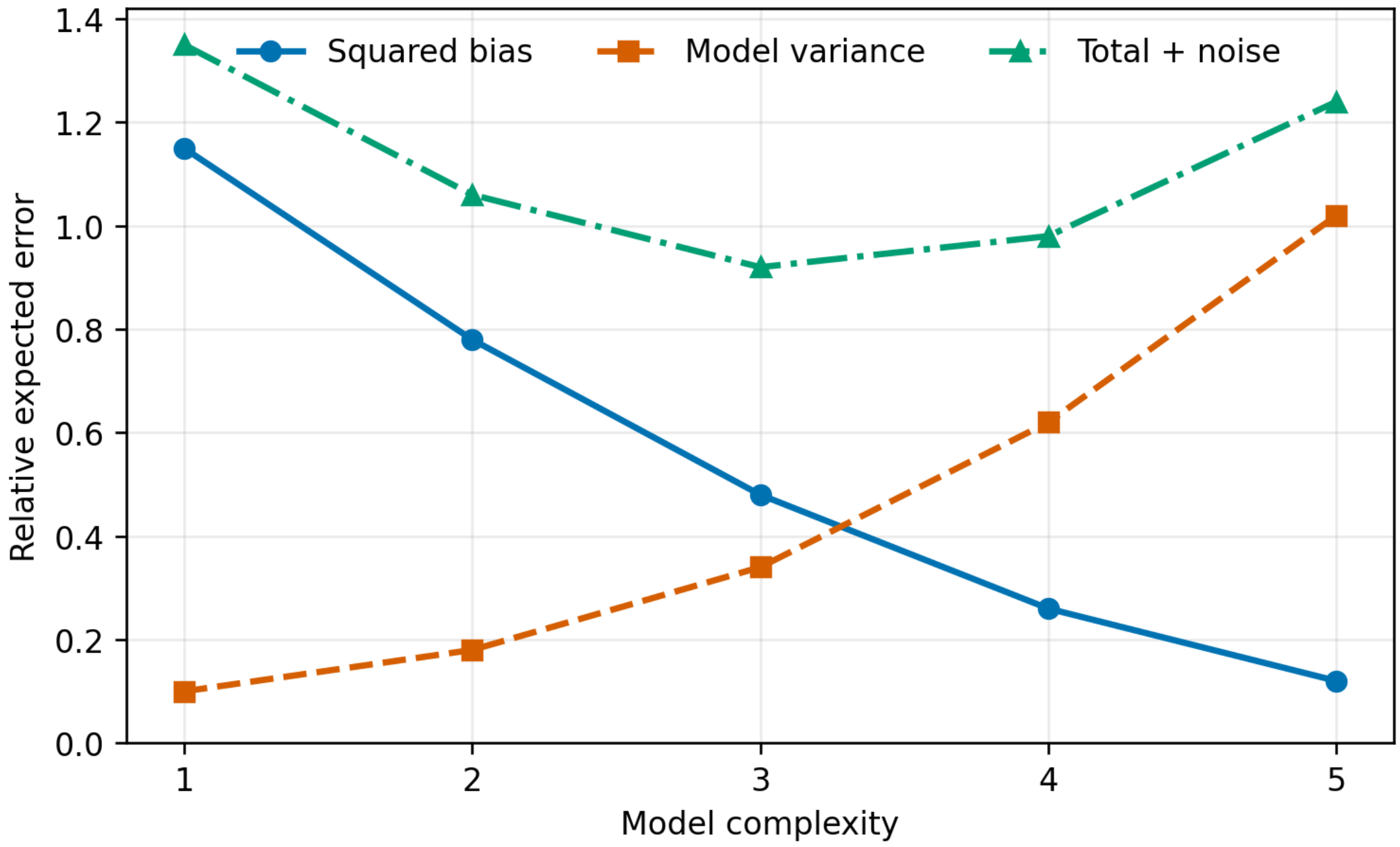

Figure 3. Schematic bias–variance curves as model complexity increases. The descending curve represents squared bias, the ascending curve represents model variance, and the U-shaped curve represents their sum plus a constant irreducible-noise term. Values are dimensionless teaching values and are not obtained from S1–S8.

The minimum of the total curve represents the best expected trade-off for the stated distribution and loss, not a model that is best under all conditions. A larger dataset may reduce the variance curve and shift the useful complexity region. Stronger regularization can move a learner toward higher bias and lower variance. Distribution shift can change all three curves because the target $P_\star$ has changed. Validation estimates the resulting risk only to the extent that its assessment data reproduce that target.

**2.3. The validation estimator has its own bias and variance**

The fitted learner and the validation estimator are different random objects. A learner may have low prediction variance while its assessed accuracy varies because the assessment set is small. Conversely, a flexible learner may have high model variance while a large test set estimates its average error precisely. Let R̂_V denote the estimate produced by validation procedure V, dataset D, randomized split specification S, and complete learning algorithm A. Relative to target risk R_target(A), estimator bias is defined in Equation (4). Recent work shows that cross-validation can target different risks according to training size and the learning procedure. The estimand must therefore be stated [12].

$$\mathrm{Bias}(\hat{\mathcal{R}}_V) = \mathbb{E}_{D,S}[\hat{\mathcal{R}}_V] - \mathcal{R}_{\mathrm{target}}(\mathcal{A}). \quad (4)$$

Equation (4) averages over newly sampled datasets $D$ and, when the procedure is randomized, over split assignments $S$. Importantly, the validation rule $V$ represents the full validation process - fitting, selection, and scoring - rather than only the fold indices. The target $\mathcal{R}_{\mathrm{target}}(\mathcal{A})$ is the learning procedure's expected risk at the intended training size in the deployment setting. For an error estimate, a positive bias is pessimistic and a negative bias optimistic. The interpretation reverses when performance is reported as accuracy.

The variance of this estimator is

$$\mathrm{Var}(\hat{\mathcal{R}}_V) = \mathbb{E}_{D,S}\left[\left(\hat{\mathcal{R}}_V - \mathbb{E}_{D,S}[\hat{\mathcal{R}}_V]\right)^2\right]. \tag{5}$$

Equation (5) concerns variation in the reported result across new datasets and permitted partitions. Repeating partitions within one dataset captures the part due to $S$ conditional on that dataset, but not the variation across fresh datasets $D$. Fold scores share much of their training data and are correlated. Their SD describes differences among folds; it is not automatically an uncertainty interval based on independent samples.

Validation bias can arise simply because the training size used during assessment differs from the one used for the final model. With a 50/50 hold-out, for instance, each assessment model is trained on only half of the available data, so its error can be pessimistic relative to a final model fitted to the full development set. The opposite problem appears when the same scores are used both to select and to evaluate a configuration, which introduces optimism. Substantial variation can still occur in a leakage-free analysis when groups are heterogeneous. Choosing a validation method is thus a trade-off among bias, variance, and computational cost rather than a progression from weak to strong.

### 2.4. Notation and the independent unit

The notation used throughout the article is brought together in Table 2. The table distinguishes row count $N_{\mathrm{row}}$ from the number of independent units $N$ and the number of groups $G$ since these quantities coincide only in an ordinary independent-row dataset. Confusion-matrix counts are introduced here; their metric equations and the macro/micro and fold-aggregation examples are derived after the validation estimators have been defined.

Table 2. Notation and symbols used throughout the tutorial. The distinction among N_row, N, G, and g(i) prevents repeated records from being silently interpreted as independent observations. TP, FN, FP, and TN use actual class in rows and predicted class in columns in every later confusion matrix.

| Symbol | Meaning |
|---|---|
| D_N | Dataset of N independent units |
| N_row | Number of stored rows, epochs, images, windows, or measurements |
| N | Number of units treated as exchangeable and independent for the target claim |
| D_tr, D_val, D_te | Training, validation, and test partitions |
| X, Y | Random input and outcome |
| P_tgt | Target deployment distribution |
| f, f_D, f* | Fixed rule f, rule f_D fitted on D, and target conditional mean or rule f* |
| A | Complete learning algorithm, including preprocessing and selection |
| L(y, ŷ) | Loss for outcome y and prediction ŷ |

| Symbol | Meaning |
|---|---|
| R(f), R_hat | Generalization risk and estimated risk |
| Π | Random split assignment or resampling specification |
| K, K_o, K_i | Ordinary, outer, and inner fold counts |
| R | Number of repetitions |
| G | Number of subjects or other indivisible groups |
| g(i) | Group identifier of row i |
| n_g | Number of rows contributed by group g |
| Λ, λ | Candidate hyperparameter set Λ and one configuration λ |
| λ_hat | Configuration selected from data |
| C | Number of outcome classes |
| π_c, π_c,k | Global and fold-specific proportions of class c |
| TP | Actual positive predicted as positive |
| FN | Actual positive predicted as negative |
| FP | Actual negative predicted as positive |
| TN | Actual negative predicted as negative |
| m_k | Metric computed in fold k |
| m_c, m_bar, s_m | Class-specific metric, replicate or fold mean, and sample standard deviation |
| P_micro, R_micro, F1_micro | Micro-averaged precision, recall, and F1 |
| m_pool | Metric computed after pooling held-out predictions |
| C_k | Confusion-count vector or matrix for fold k |
| d_r, d_bar, s_d | Paired algorithm difference in resample r, its mean, and its standard deviation |
| n_tr, n_te | Training and test sizes in one resample |
| SE, CI_95% | Standard error and 95% confidence interval |
| B, T_obs, T_b | Number of label permutations, observed statistic, and statistic in permutation b |

Table 2 provides a vocabulary for asking the main design question: which index should cross the split boundary? In independent tabular data, the row index $i$ may be sufficient. In EEG, $i$ may index an epoch while $g(i)$ identifies the participant. In paired-eye OCT, the two eye rows share a patient identifier and must remain together for new-patient assessment. In multicenter studies, nested identifiers may be needed because measurements are grouped within patients and patients are grouped within centers.

### 2.5. Dependence and effective sample size

The usual independent-and-identically-distributed argument no longer holds once rows are clustered. Adjacent EEG epochs from one recording, paired eyes from one patient, repeated laboratory panels from the same person, and images acquired on one scanner all share information. For a new-unit claim, the stored row count N can therefore give a misleading impression of how much independent evidence is actually available.

With roughly equal cluster sizes, this loss of independence can be illustrated by the following design-effect approximation:

$$N_{eff} \approx \frac{N}{1 + (\overline{m} - 1)\rho}. \quad (6a)$$

Equation (6a) takes m_bar as the average number of rows per independent unit and rho as the within-unit correlation. For N = 600 epochs, m_bar = 20, and rho = 0.60, the effective sample size is about 48.4, not 600. This is useful as a warning: row-wise counts can greatly overstate the amount of independent evidence. When rho is near zero, the rows carry more distinct information; as rho approaches one, N_eff moves toward the 30 underlying clusters. The value 48.4 therefore lies between those limits. Because rho is rarely known precisely and can differ across subjects, new-subject analyses are usually safer when the number of clusters is treated as the primary sample-size reference. Group counts and sizes should also be reported [13].

With unequal cluster sizes or dependence at several levels, the single-correlation approximation becomes less informative. In practice, keep the deployment-relevant unit intact, report how many rows each unit contributes, retain unit-level predictions, and base uncertainty resampling on independent units.

Generalization risk refers to the deployment distribution; empirical risk is calculated from the observed sample. The learner and the validation procedure each have their own bias and variance, and repeated folds within one dataset capture only part of the latter. When rows are clustered within subjects, patients, or centers, the independent evidence can be far smaller than the row count. For sample-size arguments, the number of independent target units is the more meaningful reference.

## 3. A taxonomy of data leakage

### 3.1. Leakage as an invalid information path

Data leakage occurs when model development receives information that is unavailable for the target prediction. The assessment procedure may be executed correctly, but the resulting score no longer represents independent performance. Leakage is defined by the prediction time, target population, and intended use case. It is not defined by a universal list of forbidden variables [14].

Some leakage is easy to recognize: a post-outcome treatment code cannot be used to predict an earlier outcome. Other routes are less visible. If assessment records contribute to imputation, z-scoring, PCA, intensity normalization, or batch correction before scoring, their covariates have already influenced the pipeline. Supervised ranking also exposes assessment labels to feature selection. Creating cross-validation folds afterward cannot remove information that the fitted transformation has already used.

When the split does not match the prediction target, leakage should be investigated directly. A high score may trigger the check, but the score itself says nothing about whether the pipeline is clean. If performance falls after grouping subjects, shuffling identifiers, or removing device-related variables, the original model may have used a shortcut. Label permutation is another useful check: rerun the entire pipeline inside each fold and expect performance to return toward the null. Table 3 summarizes the main leakage mechanisms, warning signs, and practical safeguards.

Table 3. Common sources of validation-related leakage, the warning signs they can produce, and practical controls for preventing them. Leakage is defined with respect to the intended prediction setting: an operation is unsafe whenever assessment information influences learning in a way that would not be available before prediction, even if no class label is used directly.

| **Leakage type** | **Mechanism** | **Common symptom or diagnostic clue** | **Prevention** |
|---|---|---|---|
| Feature-selection leakage | Outcomes from assessment rows influence ranking, filtering, dimensionality, or the selected subset | Very high CV accuracy in high-dimensional small-$N$ data; random-label accuracy exceeds chance | Refit every supervised selector inside each training fold; place selection inside the inner loop when its settings are tuned |
| Normalization and imputation leakage | Means, standard deviations, quantiles, missingness rules, PCA directions, or batch parameters use all rows | Performance drops when transformations are refitted per fold; fold statistics resemble the held-out distribution unusually well | Estimate all transformation parameters on the current training partition and apply them unchanged to held-out rows |
| Subject or group leakage | Correlated records from one identity occur in both training and assessment data | Row-level CV greatly exceeds subject-level CV; identifiers can be predicted from features | Split by subject, patient, family, eye-pair, device, recording, or another defensible independent group |
| Temporal leakage | Future values, future-derived summaries, overlapping windows, or post-index events inform earlier predictions | Performance collapses under chronological splitting; features use timestamps later than the prediction time | Enforce time-aware feature cutoffs and nonoverlapping temporal assessment; use specialized temporal CV when appropriate |
| Duplicate or augmentation leakage | Exact/near duplicates or augmented versions of one source appear on opposite sides | Near-perfect nearest-neighbor matches across partitions; image hashes or signal correlations are extreme | Group source records and all derivatives before splitting; deduplicate before partitioning |
| Site or device leakage | A mixed split lets the model exploit center/device signatures correlated with labels | High mixed-center CV but weak held-out-center performance | Hold out centers/devices according to the deployment claim; inspect class-by-center confounding |

| Leakage type | Mechanism | Common symptom or diagnostic clue | Prevention |
| --- | --- | --- | --- |
| Test-set erosion | Test feedback guides features, models, thresholds, stopping, or narrative choices | Test performance rises across analysis versions without a corresponding independent gain | Lock the test set, precommit decisions, use nested CV for selection, and obtain a fresh lockbox when reuse has occurred |
| Label leakage | Predictors contain the target, a deterministic proxy, or post-outcome information | Implausibly high performance from one or a few features | Define the prediction timestamp, audit provenance, and remove variables unavailable at that timestamp |

Although preprocessing leakage and statistical dependence are listed separately in the table, they often occur together in real studies. An OCT dataset, for example, may be normalized globally and then split eye by eye, allowing one patient to contribute both distributional information and a correlated fellow-eye record to training. EEG windows can create the same problem when overlapping segments cross a random split while retaining a subject-specific spectral signature.

### 3.2. Leakage from feature selection and preprocessing

High-dimensional data create many opportunities for chance associations. With 10,000 noise features and 60 random labels, some variables will appear related to the outcome simply because the most extreme correlations are selected. If that selection is performed globally, the assessment labels influence the representation itself, so subsequent CV evaluates features that were chosen using information from every fold. Ambroise and McLachlan [6] demonstrated this problem in microarray data. S2 recreates it by showing accuracy fall from 93.3% under the leaky pipeline to 51.7% when selection is nested.

The same information boundary applies even when $Y$ never enters the transformation. With global z-scoring, held-out observations contribute to the mean and SD that are later used for both training and assessment. In a large, homogeneous sample the effect may be small, but it can matter under distribution shift, small folds, imputation, or batch correction. A leakage-safe CV pipeline estimates each transformation from the current training fold, freezes those parameters, and only then transforms the held-out data. If the transformation itself must be chosen from data, that choice belongs in the inner loop.

Figure 4 expresses the same distinction as an information-flow problem. In the flawed pipeline, transformations and feature selection are learned before the split, which allows observations later treated as assessment data to influence development. In the leakage-safe pipeline, the split is defined first and every data-dependent operation is learned only from the current training data.

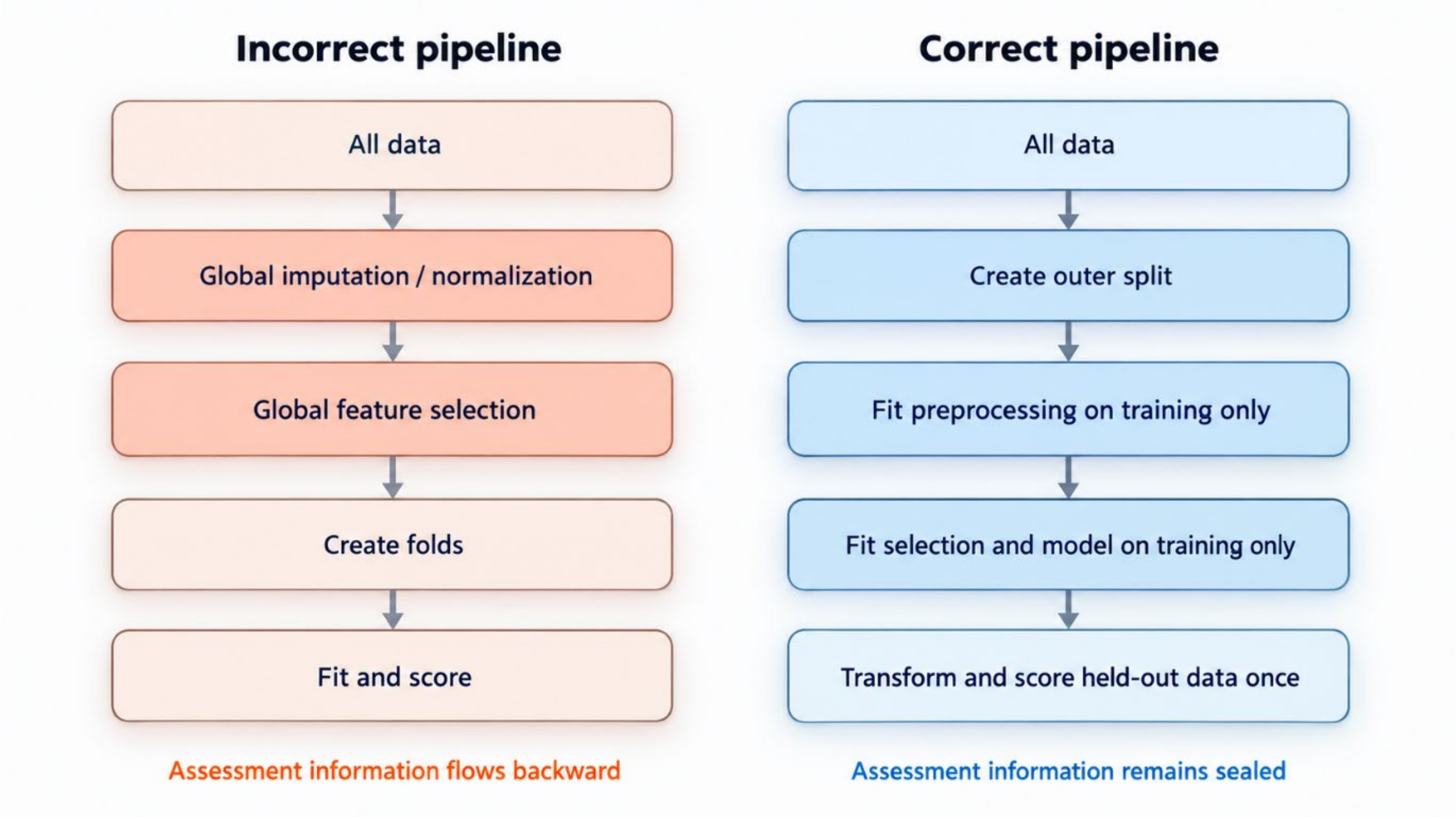

Figure 4. Information flow in an incorrect and a leakage-safe validation pipeline. In the incorrect path, global preprocessing and feature selection are performed before folds are created, allowing held-out covariates or labels to influence training. In the correct path, the split comes first and all learned transformations are fitted only on the corresponding training partition.

The word pipeline should therefore be interpreted broadly. It includes imputation, artifact rejection rules learned from data, normalization, feature construction with estimated parameters, feature selection, class rebalancing, augmentation, model fitting, probability calibration, threshold selection, and sometimes ensemble weighting. Fixed physical transformations that use no sample-derived parameter, such as a prespecified unit conversion, can be applied before splitting. However, a method does not become fixed merely because a software function has a default setting. If its state depends on the observed sample, that state must be fitted within the appropriate training boundary.

### 3.3. Subject, temporal, site, and test-set leakage

Group leakage is, in effect, inadvertent recognition of the unit being assessed. Other epochs from the same subject can expose stable anatomy, impedance, motion patterns, and device characteristics that are associated with that person's label. A classifier may then perform well without learning disease-related patterns that transfer to unseen people. The same shortcut can arise with fellow-eye OCT images because anatomy and acquisition conditions are shared. Saeb et al. [3] showed why record-wise splitting may be appropriate for personalized prediction while a new-subject diagnostic claim requires subject-wise splitting.

Temporal leakage introduces an additional ordering constraint. A model intended to predict at time $t$ cannot legitimately use measurements, summary windows, treatment decisions, or labels that arise after $t$. Random splits of heavily overlapping signal windows can likewise put near-identical time samples on opposite sides of the boundary. This tutorial does not develop forward chaining or purged k-fold in detail, but the leakage principle remains the same: reproduce the information set and delay structure available at the prediction time. For longitudinal clinical data, group and temporal constraints may need to be enforced simultaneously.

Site leakage is not merely the presence of a site variable. Device settings, reconstruction software, laboratory reference ranges, referral patterns, and class prevalence may encode site indirectly in many features. A mixed-site split can be appropriate when deployment will use the same mixture of established sites. It cannot support a claim about a new center unless centers are held out during assessment. When only three centers are available, leave-one-center-out provides three valuable transport experiments, but it cannot precisely characterize the population of all possible centers.

The final leakage mechanism is adaptive test reuse. Each inspection of test performance can influence the next model, feature set, threshold, or reporting choice, even when test rows are never directly added to training. With enough alternatives, the largest observed test score tends to select positive noise. Cawley and Talbot [8] describe the analogous overfitting of a model-selection criterion, and the remedy is to treat selection as part of training. Figure 5 previews the nested structure that contains this adaptivity.

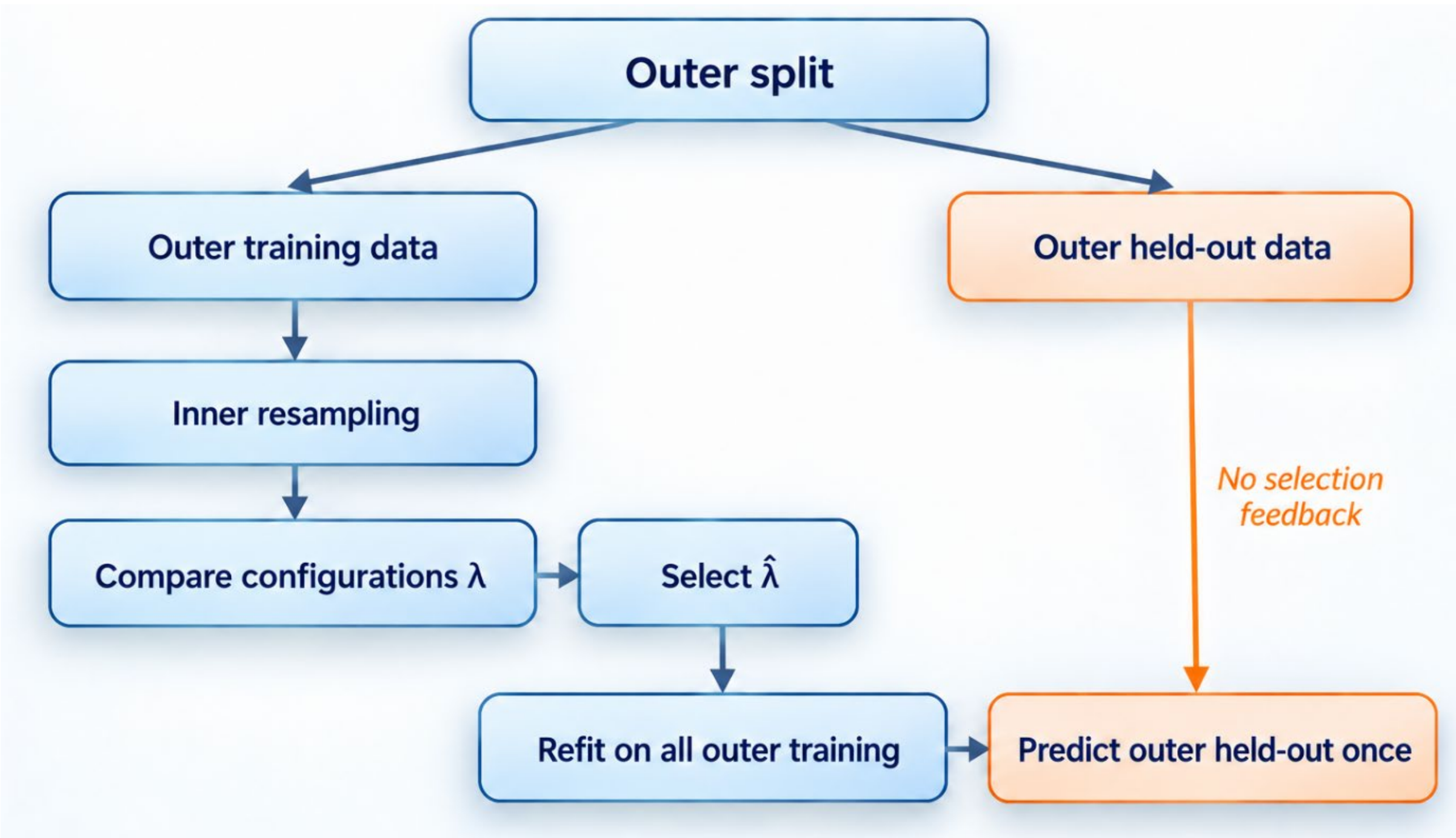

Figure 5. Information containment in nested cross-validation. Each outer training partition contains its own complete model-selection experiment. The selected configuration $\lambda_o$ is then refitted on the full outer training partition and used once to predict the corresponding outer held-out data; outer outcomes do not influence inner selection.

Nested CV does not make leaked features clean if they were computed globally before the outer loop. The raw or minimally fixed data must enter each outer iteration, and all sample-dependent operations must be reconstructed inside it. If records are grouped, the outer split must respect groups; the inner split should also respect them whenever the tuned pipeline could exploit within-group dependence.

Section 7 gives the complete formulas and manual nested group loops, while S2 and S3 quantify the consequences of violating these rules.

Table 4 places common data-dependent operations at the earliest stage where they can legitimately be learned. Here, 'allowed' means statistically permissible for that role rather than required. The final-test column is deliberately strict: the test data may be used to compute the final reported metrics, but they must not be used to alter the fitted system.

Table 4. Containment rules for common stages of machine-learning development and assessment. A learned operation is fitted only on the training data available at its level. Inner validation may choose among candidate operations; outer held-out data and the final test set may be used for assessment but must not influence selection, preprocessing state, stopping, calibration, or thresholds.

| **Operation** | **Training partition** | **Inner validation** | **Outer held-out fold** | **Final untouched test** |
|---|---|---|---|---|
| Fixed unit conversion or prespecified deterministic transform | Apply | Apply identically | Apply identically | Apply identically |
| Imputation, normalization, PCA, batch parameters | Fit here | Apply stored state | Apply outer-training state | Apply final-training state |
| Supervised feature selection | Fit here for a fixed rule | Evaluate candidate rule | Never fit | Never fit |
| Hyperparameter or model-family selection | Train candidates | Choose here | Never choose | Never choose |
| Class weights or resampling ratios estimated from data | Estimate here | Evaluate their effect | Never estimate | Never estimate |
| Probability calibration | Fit within training or a training-only calibration split | Choose calibration settings if needed | Apply only | Apply only |
| Classification threshold | Fit a prespecified threshold or tune inside training | Choose here when tuned | Apply only | Apply only |
| Early stopping | Fit on a training subset | Monitor here | Do not monitor | Do not monitor |
| Performance metric computation | Optional training diagnostic | Selection diagnostic | Unbiased outer assessment | Final locked assessment |
| Manual inspection that changes the pipeline | Permitted and documented | Permitted as model selection | Prohibited | Prohibited |
| Final refit after model specification is frozen | Fit on all available development data | Not applicable | Not applicable | Predict once |

Table 4 also shows that a train/validation/test split and nested CV rely on the same information-boundary principle even though their geometries differ. A validation subset or inner loop may guide model choices; the test subset or outer loop is reserved for estimating the performance of the resulting choice procedure. Once a result influences another development decision, that assessment data has effectively become part of development.

### 3.4. Diagnostic signs and negative controls

The mechanisms in Table 3 are easier to detect when the information boundary is reconstructed explicitly. Several quick checks can signal that a result deserves a deeper audit, although none of them is conclusive on its own.

• Label permutation: shuffle the labels, then rerun the entire analysis, including preprocessing and selection. With a properly isolated workflow, performance should return toward the null. If it remains clearly above chance, revisit feature selection, preprocessing, duplicate records, and group assignments [15].

• Unit escalation: repeat the evaluation at the level implied by the claim—for example, patients rather than images or centers rather than patients. A marked drop indicates that the original split was testing an easier problem.

• Pipeline reconstruction: create the folds before learning anything from the data. Fit imputation, scaling, harmonization, feature selection, oversampling, calibration, and threshold selection only on the training portion available at that stage.

• Heterogeneity display: report subject-, patient-, or center-level results when those units matter. A strong pooled score can otherwise hide a poorly performing center or subgroup.

• Duplicate and provenance audit: search for exact and near-duplicate files and trace augmented or segmented records to their source. Records derived from the same source should never be split across training and assessment.

A high score can be credible. The evidence is easier to assess when readers can inspect the negative controls, group assignments, held-out predictions, and order of operations used to obtain it.

Leakage includes any route by which unavailable information influences learning, whether or not a row is duplicated or a post-outcome variable is obvious. Define the assessment boundary before fitting transformations, selectors, tuning procedures, samplers, or calibrators. Label permutation and analyses at the intended unit can uncover hidden routes, but careful design prevents them in the first place.

## 4. Basic splitting approaches

### 4.1. Hold-out and train/validation/test designs

A hold-out design reserves one subset for assessment and fits the model on the remainder. It is inexpensive and easy to interpret when the held-out cases represent the classes and subgroups that matter. The trade-off is split sensitivity: changing the random allocation changes both the fitted model and the cases used for assessment. Reporting only a percentage split is therefore not enough; the corresponding absolute counts should also be given.

In a three-way split, the three subsets have distinct roles. Training data estimate model parameters; validation data support choices such as architecture, regularization, early stopping, and thresholds; and the test set is kept for the selected, frozen procedure. Training and validation data may be combined for a final refit after selection, but that refitted model has not been independently assessed unless another untouched test set exists. A subset used to monitor a deep network and decide when to stop is still development data, whatever it is called.

Stratification stabilizes class composition, but it does not change the unit that must remain independent. In S1, a 15% unstratified hold-out contains only 18 observations, so the number of positives can vary markedly around its expected value of three. A favorable split may contain only one positive and still report 17 correct predictions out of 18, or 94.4% accuracy, while offering very weak evidence about sensitivity. Another split can produce 14/18, or 77.8%. Stratification reduces these swings in minority-class counts, but it cannot rescue an assessment set that is simply too small.

**4.2. Repeated hold-out**

Repeated hold-out, also called random subsampling or Monte Carlo cross-validation, draws several training-assessment splits with a fixed training fraction. Let $I_{\text{te}}^{(r)}$ be the assessment indices in repetition $r$, let $D_{\text{tr}}^{(r)}$ contain the remaining training observations, and let $\hat{f}^{(r)} = \mathcal{A}\left(D_{\text{tr}}^{(r)}\right)$. Its average loss is

$$\hat{\mathcal{R}}_{\text{RH}} = \frac{1}{R}\sum_{r=1}^{R}\frac{1}{\left|I_{\text{te}}^{(r)}\right|}\sum_{i\in I_{\text{te}}^{(r)}} L\left(y_i, \hat{f}^{(r)}(x_i)\right). \tag{6}$$

In Equation (6), $R$ is the number of repetitions, $\left|I_{\text{te}}^{(r)}\right|$ is the assessment-set size in repetition $r$, and $L$ is the chosen loss. The learning algorithm $\mathcal{A}$ includes every fold-local preprocessing and selection operation. Assessment sets can overlap across repetitions, so their losses are dependent. The average is useful for describing sensitivity to partitioning, but $R$ repetitions do not create $R$ independent experiments.

The locked S1 example uses 20 repetitions. Unstratified accuracies of 77.8%, 83.3%, 88.9%, and 94.4% occur 2, 5, 8, and 5 times, yielding a mean of 87.8% and an SD of 5.3 percentage points. Stratified repeated 5-fold CV gives a similar mean (88.3%) but a much smaller SD (1.2 percentage points). Figure 6 plots these fixed teaching distributions; they are not observations from a clinical study.

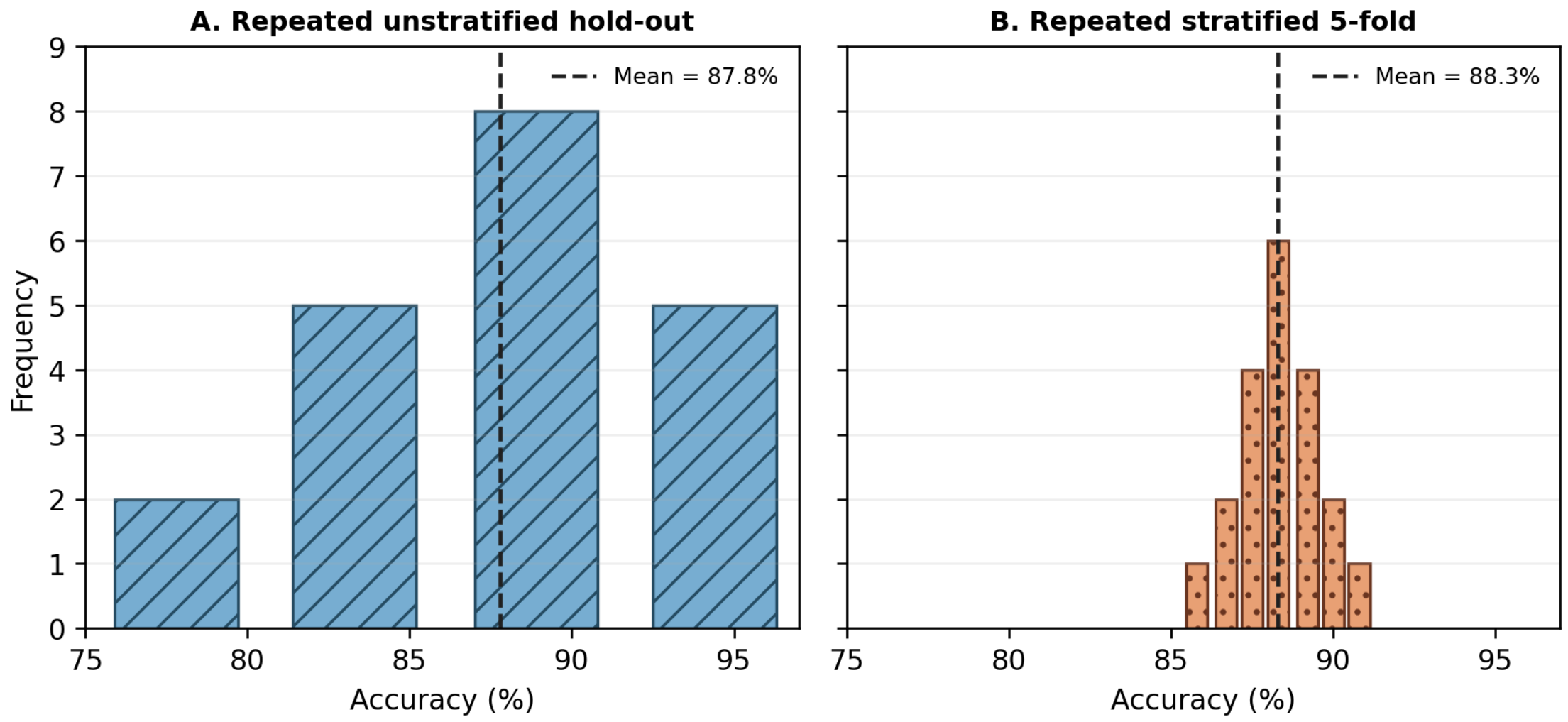

Figure 6. Accuracy distributions for the locked S1 example. Panel A shows 20 repeated 15% unstratified hold-outs (77.8%-94.4%); Panel B shows 20 repetitions of stratified 5-fold CV (85.8%-90.8%). Values come from the fixed repetition tables indexed in Appendix Table A1 and reproduced in Listing A3.

The tighter stratified distribution should not be interpreted as proof that every stratified estimate lies closer to the unknown truth. In this constructed example, stratification removes one avoidable source of variation: random changes in the number of minority-class cases assigned to assessment. The sample still contains only 120 observations, including 20 minority-class observations. Repeated estimates should therefore be presented as a distribution of dependent results, not as if repetition had enlarged the cohort.

Table 5 brings together the main choices in the basic-splitting family. Its 'reasonable use' column assumes that the independent unit is aligned with deployment and that learned transformations are fitted only after the split. Before accepting any percentage, the resulting test counts should be checked for each class, subgroup, and relevant group level.

Table 5. Comparison of basic splitting strategies. A split ratio is useful only when it leaves enough absolute observations in the target classes and groups. Stratification stabilizes composition, whereas grouping preserves independence; neither can replace the other.

| Method | Subset roles | Strength | Main limitation | Reasonable use |
|---|---|---|---|---|
| Single hold-out | Train / test | Minimal cost; simple audit trail | High dependence on one split | Large independent datasets or a locked external-like test subset |
| Train/validation/test | Fit / choose / assess | Clear separation of selection and testing | Uses fewer cases at each role | Large datasets, especially deep learning with early stopping |
| Stratified hold-out | Train / test with class control | Stabilizes class counts | Does not enforce group independence | Imbalanced independent observations or stratified groups when feasible |

| Method | Subset roles | Strength | Main limitation | Reasonable use |
|---|---|---|---|---|
| Repeated hold-out | Repeated train / test | Describes partition sensitivity | Repeated scores and test rows overlap | Small-to-medium independent data when k-fold coverage is not required |
| Repeated stratified hold-out | Repeated class-balanced train / test | Controls class composition and split sensitivity | Still dependent; minority count remains finite | Imbalanced classification with adequate minority cases per assessment split |

Table 5 also makes clear why a percentage alone is a poor design rule. An 80/20 split can be entirely adequate with 100,000 independent samples yet unusable with 50 patients and only five positive cases. When observations are repeated within a patient, a group-aware 80/20 split should allocate patients rather than individual images or epochs. The next section turns from repeated random partitions to fold-based schemes that rotate the assessment role across independent units.

### 4.3. Stratification, grouping, and the importance of absolute test counts

Stratification and grouping solve different problems. Stratification aims to keep class proportions similar across folds; grouping prevents dependent records from being separated. When both are required, group integrity comes first. A fold may be perfectly balanced yet still be invalid for a new-patient claim if records from the same patient appear in both training and assessment.

Absolute class counts often matter more than the split percentage alone. With 120 observations but only 20 positives, a 15% hold-out contains about three positive cases. One extra false negative would then shift sensitivity by roughly 33 percentage points. Stratification and repetition can reduce or reveal this allocation instability, but they cannot create additional independent positive cases.

Begin with the unit that must be unseen at deployment. Count positives and negatives at that level, keep related records together, and stratify only when the group structure allows it. The split ratio and fold count should then follow from the actual class and group counts.

Hold-out validation is most useful when the assessment subset is sufficiently large, representative, group-correct, and genuinely untouched. Stratification can stabilize class composition, and repetition can show how sensitive the estimate is to partitioning, but neither adds independent evidence. Split ratios should therefore be chosen from absolute class and group counts after the independent unit has been established.

## 5. The cross-validation family

### 5.1. K-fold, stratified k-fold, and repetition

k-fold cross-validation uses $K$ folds. Let the independent indices be partitioned into disjoint folds $I_1, \dots, I_K$, with $n_k = |I_k|$ and $\cup_k I_k = \{1, \dots, N\}$. For fold $k$, the complete learning procedure is fitted on $D_{-k} = D_N \setminus I_k$. The held-out fold loss is

$$\hat{\mathcal{R}}^{(k)} = \frac{1}{n_k} \sum_{i \in I_k} L\left(y_i, \mathcal{A}(D_{-k})(x_i)\right). \tag{7}$$

In Equation (7), $I_k$ is the assessment index set, $n_k$ is its size, and $\mathcal{A}(D_{-k})$ is fitted without any observation in fold $k$. The loss is computed only from predictions made while the corresponding observation is held out. Every row receives one held-out prediction in ordinary k-fold CV. If fold sizes differ, the observation-weighted estimator is

$$\hat{\mathcal{R}}_{K\text{-CV}} = \sum_{k=1}^{K} \frac{n_k}{N} \hat{\mathcal{R}}^{(k)} = \frac{1}{N} \sum_{k=1}^{K} \sum_{i \in I_k} L\left(y_i, \mathcal{A}(D_{-k})(x_i)\right). \tag{8}$$

In Equation (8), the weights n_k/N ensure that each observation contributes equally. When every fold has the same size, the expression reduces to (1/K) Σ_k R_k. For nonlinear metrics such as F1 or MCC, averaging fold metrics and computing one metric from pooled held-out predictions are different operations. Section 9 gives a numerical example in which unequal group-fold sizes make that distinction large.

Stratified k-fold seeks to keep the class proportion in each fold close to the global proportion. If $n_c$ observations belong to class $c$ and $n_{c,k}$ of them occur in fold $k$, a formal tolerance condition is

$$\left|\pi_{c,k} - \pi_c\right| = \left|\frac{n_{c,k}}{n_k} - \frac{n_c}{N}\right| \le \delta_c, \qquad c = 1, \ldots, C, \quad k = 1, \ldots, K. \tag{9}$$

In Equation (9), $C$ is the number of classes, $\pi_c = n_c/N$ is the global class proportion, $\pi_{c,k} = n_{c,k}/n_k$ is the fold proportion, and $\delta_c$ is an allowed discrepancy. Discrete counts may prevent exact equality, especially for rare classes. A necessary practical condition for every fold to contain class $c$ is $n_c \ge K$. Under grouping, the analogous requirement depends on the number of groups containing the class, not the number of rows.

Repeated k-fold redraws the fold partition $R$ times and performs $RK$ fits. It provides several held-out predictions per observation and describes partition sensitivity more thoroughly than one partition. The repetitions remain dependent because they use the same dataset, and all predictions from one source subject must remain group-consistent. Kohavi's [16] influential empirical comparison supported stratified 10-fold for the studied classification settings, but it should not be converted into a universal rule independent of sample size, groups, learner stability, or compute.

### 5.2. LOOCV and leave-p-out

Leave-one-out cross-validation is the limiting k-fold design with $K = N$. Each training set contains $N - 1$ independent observations, and each assessment set contains one. Its estimate is

$$\hat{\mathcal{R}}_{\text{LOOCV}} = \frac{1}{N} \sum_{i=1}^{N} L\left(y_i, \mathcal{A}(D_N \setminus \{(x_i, y_i)\})(x_i)\right), \qquad K = N. \tag{10}$$

In Equation (10), observation  is predicted by a model that excludes exactly that observation. LOOCV has little training-size mismatch relative to fitting on all  units. However, its training sets differ by only two observations, its held-out losses can be unstable, and its  fits may be expensive. High computational cost therefore does not guarantee a low-variance estimate; Bengio and Grandvalet [17] show that cross-validation variance cannot be recovered by treating overlapping folds as independent, and LOOCV does not escape that dependence.

Leave-p-out withholds every subset of $p$ independent units and requires $\binom{N}{p}$ fits when exhaustive. LOOCV is its $p = 1$ special case. Exhaustive leave-p-out becomes infeasible quickly: for $N = 100$ and $p = 2$, $\binom{100}{2} = 4{,}950$ fits are needed before tuning is considered. Randomly sampling only some $p$-out subsets turns the method into a form of repeated subsampling and changes its estimator.

**5.3. Choosing k: bias, variance, and cost**

As $K$ increases, each model is trained on a larger fraction of the data, which often reduces training-size bias. The variance need not move monotonically because fold losses and overlapping training sets are dependent. Bengio and Grandvalet [17] showed that no universally unbiased variance estimator for k-fold CV can be obtained from the fold results alone; Arlot and Celisse [18] review the broader trade-offs. Figure 7 is therefore meant as a qualitative guide, not a theorem that selects an optimal $K$.

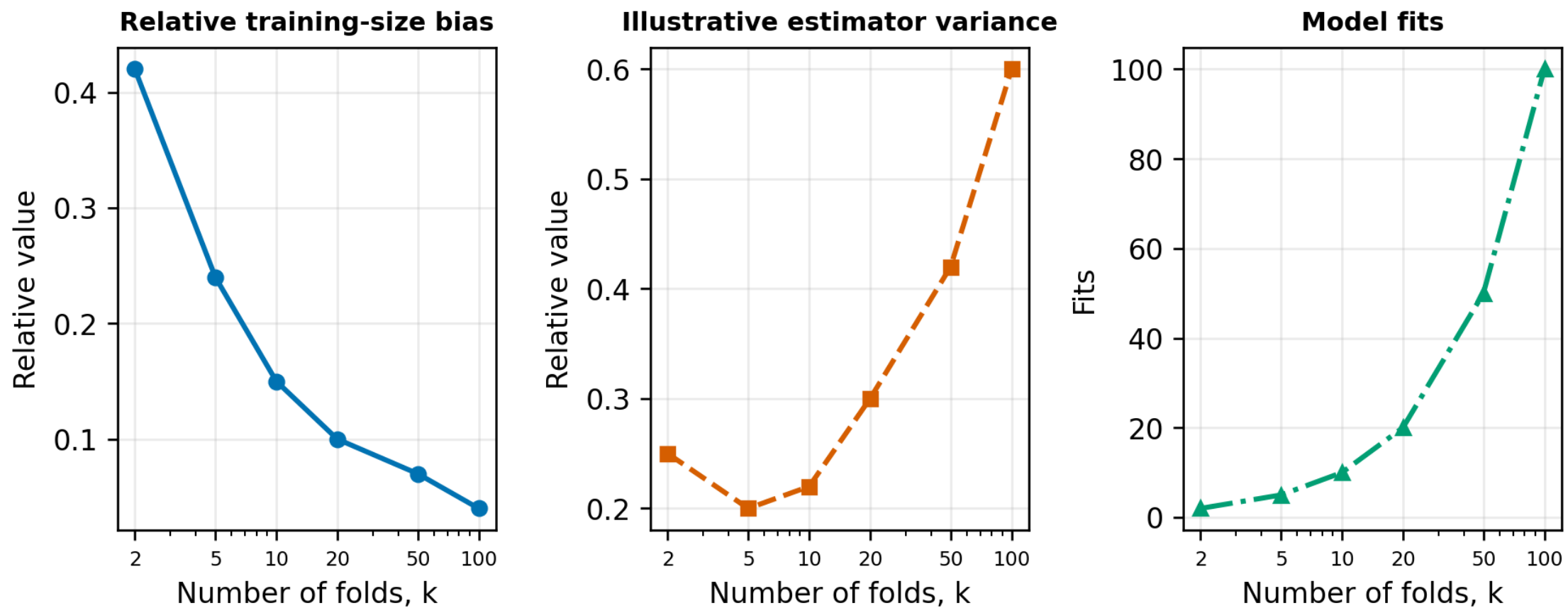

Figure 7. Schematic illustration of the relationship between fold count, training-size bias, estimator variability, and computational cost for $N = 100$. As the training fraction grows, training-size bias may fall, while the number of model fits rises with $K$. For large $K$, assessment folds become small and strongly dependent, which may increase estimator variability. The curves are dimensionless and are intended only to illustrate the trade-off.

For independent-row classification, $K = 5$ or $10$ is a reasonable place to start. A smaller $K$ needs fewer fits and leaves more observations in each assessment fold, at the cost of training each model on less data. A larger $K$ can help when the dataset is very small and fitting is cheap, although repeated five- or 10-fold CV may still be more informative than a single LOOCV run. With grouped data, feasible values of $K$ are further limited by the number of groups and their class balance. Table 6 summarizes these considerations.

Table 6. Practical trade-offs among cross-validation choices. Fit counts exclude hyperparameter search and outer refitting. Training fractions refer to independent units; dependent rows must remain grouped within folds.

| Method | Fits | Training fraction per fit | Main advantage | Main caution |
|---|---|---|---|---|
| k-fold | $K$ | $(K-1)/K$ | Every unit is assessed once | One partition may be unrepresentative |
| Stratified k-fold | $K$ | $(K-1)/K$ | Stabilizes class counts | Row stratification can still leak groups |
| Repeated k-fold | $RK$ | $(K-1)/K$ | Exposes partition sensitivity | Repetitions and predictions are dependent |
| Repeated stratified k-fold | $RK$ | $(K-1)/K$ | Useful for imbalanced independent data | Rare classes still limit feasible $K$ |
| LOOCV | $N$ | $(N-1)/N$ | Minimal training-size mismatch | High fit count and potentially high variance |
| Exhaustive leave-p-out | $\binom{N}{p}$ | $(N-p)/N$ | Systematic assessment of all $p$-out subsets | Combinatorial cost |

Table 6 illustrates why more folds do not automatically mean a better estimate. Subject-level five-fold CV respects the target unit, whereas row-wise LOOCV over 600 epochs does not if one participant's epochs appear on both sides of the split. Repetition is valuable when it exposes partition sensitivity, not when it merely produces more decimal places. The first priorities are the correct independent unit and a clean separation between selection and assessment.

### 5.4. Confusion-matrix foundation

Sensitivity, precision, and F1 are all derived from the same four held-out counts, so the positive class should be stated before these measures are interpreted. Figure 8 provides the actual-by-predicted layout together with the locked S2 noise-feature example. The leaky workflow produces 28 true positives, 2 false negatives, 2 false positives, and 28 true negatives; after feature selection is nested, the corresponding counts are 15, 15, 14, and 16.

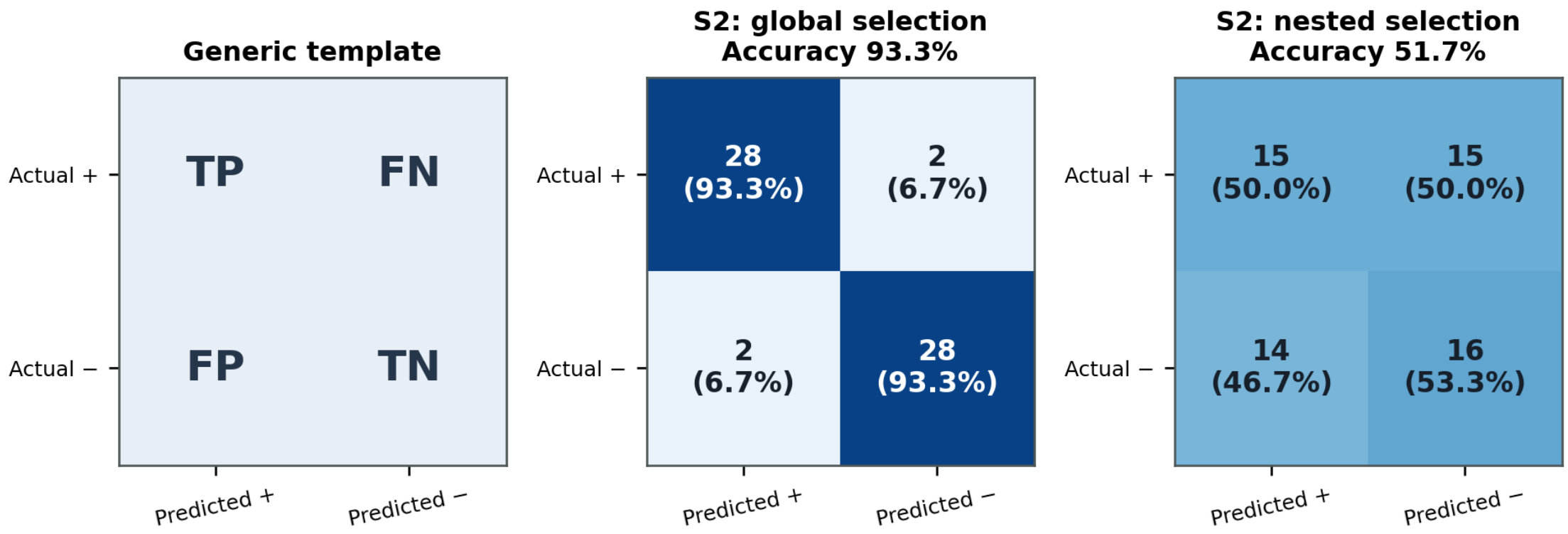

Figure 8. Binary confusion matrices for the fixed S2 selection-leakage example. Rows are actual classes and columns are predictions. With feature selection performed globally before CV, the matrix is [28,2; 2,28] and accuracy reaches 93.3%. Moving selection inside CV gives [15,15; 14,16] and reduces accuracy on random labels to 51.7%.

The side-by-side matrices also show why overall accuracy can be misleading. Because S2 is balanced, the leaky analysis appears strong in both classes, while the nested analysis stays near chance. S1 highlights a different issue: 94.4% accuracy is possible when the assessment set contains only one positive case, making sensitivity hinge on a single observation. Equations (11)-(18) therefore use held-out counts to report class-specific metrics.

$$\text{Accuracy} = \frac{TP + TN}{TP + TN + FP + FN}. \tag{11}$$

In Equation (11), $TP$, $TN$, $FP$, and $FN$ are the four confusion counts and their sum is the assessment size. Accuracy is the fraction of all predictions that are correct.

$$\text{Sensitivity} = \text{Recall} = \frac{TP}{TP + FN}. \tag{12}$$

In Equation (12), $TP + FN$ is the number of actual positive cases. Sensitivity is the fraction of actual positives detected by the model.

$$\text{Specificity} = \frac{TN}{TN + FP}. \tag{13}$$

In Equation (13), $TN + FP$ is the number of actual negative cases. Specificity is the fraction of actual negatives correctly rejected.

$$\text{Precision} = \frac{TP}{TP + FP}. \tag{14}$$

In Equation (14), $TP + FP$ is the number of cases predicted as positive. Precision is therefore the proportion of positive predictions that are correct, and its value depends on class prevalence.

$$F_1 = \frac{2\,\text{Precision Recall}}{\text{Precision} + \text{Recall}} = \frac{2TP}{2TP + FP + FN}. \tag{15}$$

In Equation (15), $F_1$ is the harmonic mean of precision and recall. It does not include $TN$ directly and is undefined under some empty-class or empty-prediction cases unless a reporting convention is declared.

$$\text{Balanced Accuracy} = \frac{1}{2}\left(\frac{TP}{TP+FN} + \frac{TN}{TN+FP}\right). \tag{16}$$

In Equation (16), the two terms are sensitivity and specificity. Balanced accuracy gives equal weight to the positive and negative classes regardless of their observed counts.

$$\text{MCC} = \frac{TP\,TN - FP\,FN}{\sqrt{(TP+FP)(TP+FN)(TN+FP)(TN+FN)}}. \tag{17}$$

Equation (17) defines MCC as the Matthews correlation coefficient. It ranges from −1 for perfectly reversed labels, through 0 when there is no linear association, to +1 for perfect classification, provided that the denominator is nonzero. The metric is particularly useful when the classes are imbalanced [19].

$$\kappa = \frac{p_o - p_e}{1 - p_e}, \qquad p_o = \frac{TP+TN}{N_{\text{eval}}}, \qquad p_e = \frac{(TP+FP)(TP+FN) + (FN+TN)(FP+TN)}{N_{\text{eval}}^2}. \tag{18}$$

In Equation (18), $p_o$ denotes observed agreement, $p_e$ is the agreement expected from the observed actual and predicted marginals, $N_{\text{eval}} = TP + TN + FP + FN$ gives the number of evaluated predictions, and $\kappa$ is Cohen's kappa. Kappa reaches 1 with perfect agreement and equals 0 when observed agreement is no better than the marginal expectation. Prevalence and marginal imbalance can still affect how the value is interpreted.

Table 7 links each formula to the question it is intended to answer. Whatever metric is chosen, its denominator and uncertainty still need to be reported. In multiclass settings, one-versus-rest counts change with the class under consideration, so both the selected class and the macro or micro aggregation rule must be made explicit.

Table 7. Interpretation of binary confusion-matrix metrics and threshold-free summaries. The first eight rows are calculated from a single confusion matrix at a fixed threshold, whereas the final two summarize score ranking before a threshold is selected. Every quantity must come from genuinely held-out predictions or scores. If a denominator is zero, the result is reported as undefined unless a prespecified convention is required for macro averaging.

| Metric | Main question | Important limitation |
|---|---|---|
| Accuracy | What fraction of all cases is correct? | Can be dominated by the majority class |
| Sensitivity / recall | What fraction of actual positives is detected? | Ignores true negatives |
| Specificity | What fraction of actual negatives is rejected? | Ignores true positives |
| Precision | How many positive predictions are correct? | Changes with prevalence |
| F1 | How well are precision and recall balanced? | Omits true negatives and depends on positive-class choice |
| Balanced accuracy | What is the mean recall of the two classes? | Treats both classes as equally important by definition |

| Metric | Main question | Important limitation |
|---|---|---|
| MCC | How strongly do binary predictions and outcomes agree? | Undefined when a marginal factor is zero |
| Cohen's kappa | How much agreement exceeds marginal expectation? | Sensitive to prevalence and marginal distributions |
| ROC-AUC | How well are positive and negative cases separated at any threshold? | Invariant to prevalence; says nothing about calibration or about any particular operating point |
| PR-AUC (average precision) | How well are positives ranked when they are rare? | Depends on prevalence; its baseline is the positive rate, not 0.5 |

The split design and the evaluation metric answer different questions. Balanced accuracy can address class imbalance, but it cannot repair subject leakage. LOSO preserves subject independence, yet a single overall accuracy can still hide weak sensitivity or poor performance at one center. A convincing report therefore pairs a deployment-aligned split with class-specific counts, complementary metrics, and an explicit uncertainty analysis.

### 5.5. Pooling held-out predictions versus averaging fold metrics

Cross-validation yields a prediction each time an observation is held out. These predictions can be combined into one confusion matrix, or a metric can be calculated within each fold before the fold values are averaged. The answers can differ when folds have unequal sizes. They can also differ for nonlinear measures such as F1.

Pooling the held-out predictions makes the primary result easy to audit because every metric can be recomputed directly from TP, FN, FP, and TN. Fold means and standard deviations remain useful for showing how sensitive the result is to partitioning. They should not, however, be interpreted as independent experimental replicates, because the folds share much of their training data.

With grouped data, aggregation also determines whose observations carry more weight. A pooled row-level metric favors groups contributing many records; averaging group metrics gives each group equal weight. Both can be useful, but they address different targets. Section 9.10 works through an example. Reports with unequal group sizes should make their weighting choice explicit.

### 5.6. Threshold-free summaries and where they belong

Equations (11)–(18) apply after continuous scores have been turned into class labels. Many pipelines generate scores first and choose a threshold later. Table 4 places threshold selection in training or inner validation. Before that threshold is fixed, a threshold-free summary is often more informative, provided that its scores were obtained from genuinely held-out observations.

As the decision threshold moves, the ROC curve traces sensitivity against one minus specificity. ROC-AUC can be interpreted as the probability that a randomly chosen positive receives a higher score than a randomly chosen negative. Because both axes are conditional on the true class, changing prevalence does not alter them. This is useful when comparing discrimination across cohorts, but ROC-AUC alone does not tell us how trustworthy a positive alert will be when prevalence changes.

Precision-recall analysis makes the effect of prevalence easier to see. Precision is the fraction of positive predictions that are correct, so it may fall when the negative population grows even if specificity barely changes. Average precision summarizes the curve, but its natural reference level is the prevalence of the positive class. An AP of 0.30 therefore means something different in a balanced cohort than in one with 2% positives, and the baseline prevalence should be reported with the score.

Table 7b makes the difference concrete with a locked example that can be checked by hand. Two cohorts share exactly the same class-conditional score distribution over ten score bins, so the classifier discriminates identically in both. The first cohort contains 100 positive and 100 negative cases; the second contains the same 100 positives and nineteen times as many negatives, giving a five percent prevalence. As Figure 9 shows, the two ROC curves coincide point for point and both give ROC-AUC 0.934. The precision–recall curves diverge immediately: average precision is 0.918 in the balanced cohort and 0.463 in the rare-positive cohort.

Table 7b. Prevalence invariance of ROC and prevalence dependence of precision. Both cohorts contain the same 100 positive cases and the same class-conditional score distribution. The operating point predicts positive when the score falls in the top three bins.

| **Quantity** | **Balanced cohort (prevalence 50%)** | **Rare-positive cohort (prevalence 5%)** |
|---|---|---|
| Positive / negative counts | 100 / 100 | 100 / 1,900 |
| ROC-AUC | 0.934 | 0.934 |
| PR-AUC (average precision) | 0.918 | 0.463 |
| PR baseline (positive rate) | 0.500 | 0.050 |
| Confusion counts at the operating point | TP 68, FN 32, FP 4, TN 96 | TP 68, FN 32, FP 76, TN 1,824 |
| Sensitivity | 68.0% | 68.0% |
| Specificity | 96.0% | 96.0% |
| Precision | 94.4% | 47.2% |
| F1 | 79.1% | 55.7% |
| MCC | 0.667 | 0.540 |

The last three rows show the clinical consequence of prevalence. At 68% sensitivity and 96% specificity, 94 of 100 positive predictions are correct in the balanced cohort, but fewer than 48 of 100 are correct when positives are rare. Neither the classifier nor the threshold has changed. Reporting sensitivity, specificity, and ROC-AUC alone leaves the likely number of false positive alerts in the intended population unclear.

Three containment rules apply to threshold-free summaries without modification. First, the curve must be built from held-out scores, exactly as the confusion counts are; a curve drawn from training scores is optimistic for the same reason a training accuracy is. Second, under cross-validation the choice between one curve from pooled held-out scores and a mean of per-fold curves is the aggregation decision analysed in Section 9.10, and the pooled curve is generally the more transparent primary statement because per-fold curves are defined on different threshold grids. Third, a threshold-free score is a selection criterion like any other: choosing the operating point that maximises it on assessment data reintroduces exactly the optimism demonstrated in S7. The threshold belongs in the training partition or the inner loop, as Table 4 and Table 9 already require.

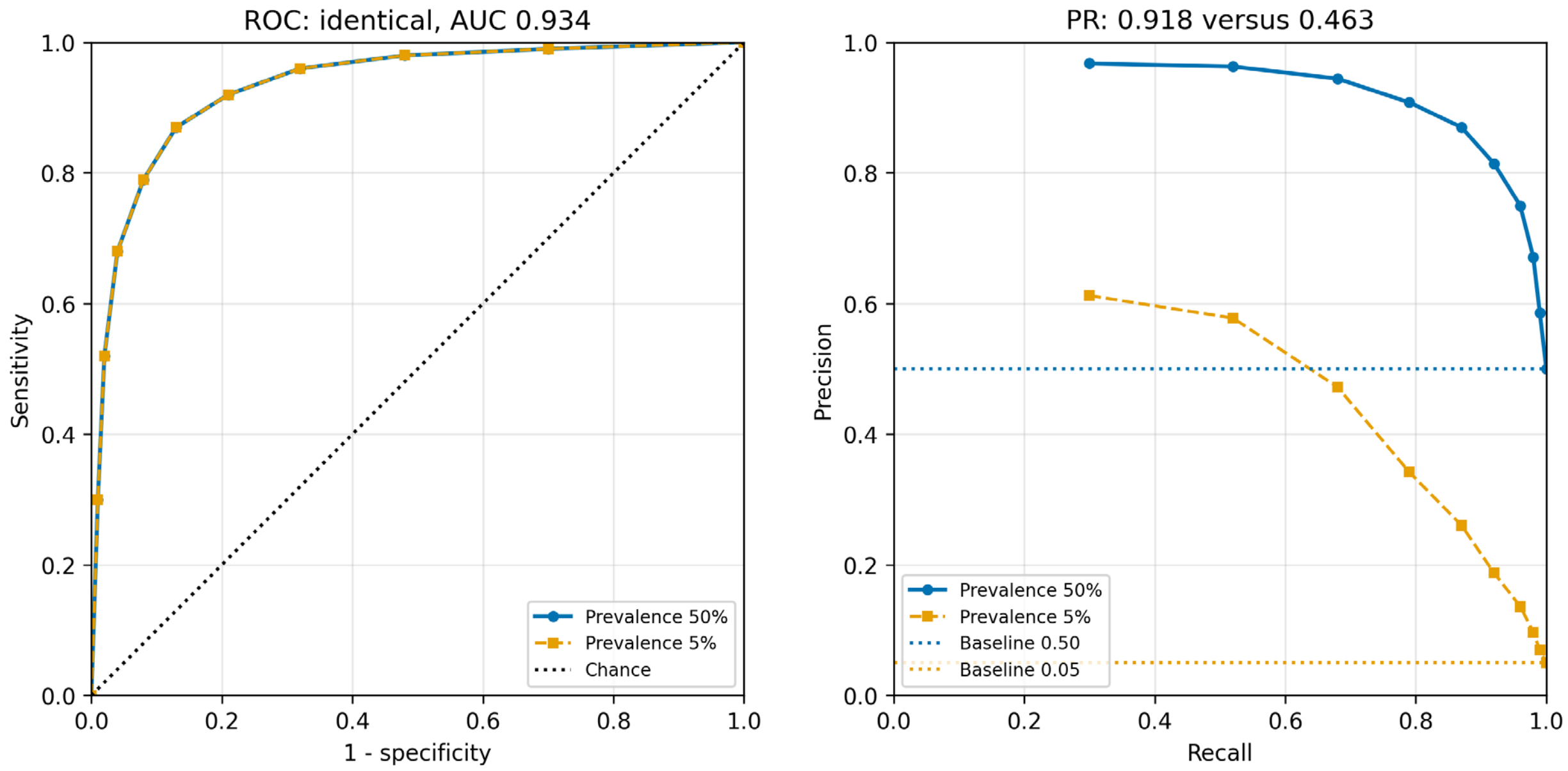

Figure 9. Threshold-free summaries under two prevalences. Both cohorts contain the same 100 positive cases and the same class-conditional score distribution over ten score bins; the second cohort simply contains nineteen times as many negatives. The receiver operating characteristic curves for that reason coincide point for point and share ROC-AUC 0.934, because both of its axes are conditional on the true class. The precision–recall curves diverge, with average precision 0.918 against 0.463, and their baselines are the respective positive rates of 0.50 and 0.05. Discrimination is unchanged; the trustworthiness of a positive prediction is not.

Finally, discrimination is not calibration. Two models with identical ROC-AUC can produce very different predicted probabilities, and a model that ranks cases perfectly can still report a ten percent risk for patients whose observed event rate is forty percent. Where predicted probabilities will be shown to a decision maker, agreement between predicted and observed risk should be reported alongside discrimination, for example through a calibration plot with its slope and intercept, or through the Brier score. This tutorial keeps its worked examples in the confusion-matrix domain because those examples remain auditable by hand, but a deployment-facing report should contain a discrimination summary, a calibration summary, and the confusion counts at the operating point that will actually be used.

These methods answer different practical needs. k-fold CV rotates held-out units, stratification limits class-composition swings, and repetition shows how sensitive the estimate is to the partition. LOOCV uses almost all data for training but is expensive, whereas exhaustive leave-p-out becomes infeasible quickly. Pooled held-out predictions and fold-level averages also weight observations differently. Threshold-free measures should be reported before a decision threshold is fixed, with the caveat that ROC-AUC is prevalence-invariant while precision is not. State the positive class, aggregation rule, and denominator convention, and compute reported metrics only from held-out predictions or scores.

## 6. Group-aware and subject-aware validation

### 6.1. Defining the independent unit for generalization

Group-aware validation asks whether the model still works on units it never saw during training. If records from the same person appear on both sides of a split, part of the apparent performance may come from recognizing that person. EEG epochs can share anatomy and recording conditions, fellow-eye OCT images share patient-specific structure, and repeated blood panels reflect the same underlying physiology and care history. The group identifier should therefore be defined before the folds are created.

The rule is simple: rows that could reveal the same identity or acquisition context must remain together whenever that identity or context is intended to be new. This rule does not imply that every possible hierarchy must always be held out. A personalized EEG model may legitimately train on earlier epochs from the same person if its deployment target is a later epoch for that person. A diagnostic model for new patients requires subject-level separation. The split is correct only relative to the stated use case.

### 6.2. LOSO, LOGO, GroupKFold, and StratifiedGroupKFold

Let subjects be indexed by $s = 1, \dots, G$, let $I_s = \{i: g(i) = s\}$ contain the $n_s$ rows of subject $s$, and let $D_{-s} = D_N \setminus I_s$. A pooled held-out LOSO loss that permits unequal subject sizes is

$$\hat{\mathcal{R}}_{\text{LOSO,pool}} = \frac{1}{\sum_{s=1}^{G} n_s} \sum_{s=1}^{G} \sum_{i \in I_s} L\left(y_i, \mathcal{A}(D_{-s})(x_i)\right). \tag{19}$$

In Equation (19), $G$ is the number of subjects, $n_s$ is the number of rows from subject $s$, and each row is predicted by a model trained without any row from that subject. Subjects with more rows receive more weight because the denominator is $\sum_s n_s$. This estimator answers a random-row-from-a-new-subject question when the observed row-frequency pattern is relevant. An equal-group alternative is

$$\hat{\mathcal{R}}_{\text{LOSO,subject}} = \frac{1}{G} \sum_{s=1}^{G} \left[ \frac{1}{n_s} \sum_{i \in I_s} L\left(y_i, \mathcal{A}(D_{-s})(x_i)\right) \right]. \tag{20}$$

In Equation (20), every subject contributes the same weight $1/G$ regardless of $n_s$. The bracketed term is the mean loss for subject $s$. Equations (19) and (20) coincide when all subjects contribute the same number of rows, as in S3. When group sizes differ, both may be informative and should be labeled as pooled held-out and equal-group results rather than treated as interchangeable.

LOSO is a special case of Leave-One-Group-Out (LOGO) in which the group is a subject. GroupKFold allocates several complete groups to each of $K$ assessment folds and is cheaper when $G$ is large. StratifiedGroupKFold adds an approximate class-balance objective while keeping each group indivisible. Exact balance may be impossible if a class occurs in fewer than $K$ groups or if a few large groups dominate the rows. Table 8 compares these choices and the domain-specific variants.

Table 8. Comparison of group- and subject-aware validation methods. The indivisible unit is kept entirely in training or assessment. Stratification is secondary to group integrity: class proportions may be approximate, but a group must never be split to make them exact.

| Method | Indivisible unit | Fits | Target question | Main limitation |
|---|---|---|---|---|
| LOSO | Subject | $G$ | Performance for a new subject | Expensive and variable when $G$ is small |
| LOGO | Any declared group | $G$ | Performance for a new group of the same type | Group definition must match deployment |
| GroupKFold | Complete group | $K$ | New groups drawn from a similar group population | Class and row counts can be uneven |
| StratifiedGroupKFold | Complete group with approximate class balance | $K$ | New groups with more stable fold composition | Exact stratification may be infeasible |
| Leave-one-session/trial/record-out | Acquisition unit | Number of units | Transfer to a new session, trial, or recording | May retain subject identity unless subjects are also grouped |
| Patient-level paired-eye split | Patient and both eyes | $K$ or patients | Performance for a new patient | Fewer independent units than eye count suggests |
| Leave-one-device-out | Device | Number of devices | Transfer across instruments | Few devices give imprecise transport evidence |
| Leave-one-center-out | Clinical center | Number of centers | Transport to an unseen center | Very wide uncertainty with few centers; center may confound class |

GroupKFold is often a better computational choice than LOSO when hundreds of subjects are available, provided its folds contain enough groups and classes. LOSO gives a clinically interpretable subject-by-subject distribution and exposes influential subjects. Leave-one-center-out addresses a harder target than mixed-center GroupKFold because the assessment environment itself is unseen. Roberts et al. [20] discuss the broader need for structure-aware CV under temporal, spatial, and hierarchical dependence.

### 6.3. Sessions, trials, eyes, and centers

A label such as 'leave-one-record-out' says little about independence. Withholding one recording while training on another from the same participant may still reveal that participant's identity. Map the hierarchy explicitly, such as center -> patient -> eye -> B-scan or subject -> session -> trial -> epoch.

Choose the outer split at the level of the scientific claim and keep every descendant with its parent. Section 7 extends this requirement to nested designs.

For paired-eye OCT, a claim about performance in new patients generally requires both eyes of the assessment patient to be kept out of training. A random split at the eye level may place the fellow eye in the training set and allow the model to exploit shared anatomy or acquisition artifacts. Eye-level separation can still be meaningful if the intended use is prediction for a second eye after the first eye has already been observed. The key is to state that target explicitly instead of treating the two validation schemes as competing algorithms.

In multicenter studies, center identity may represent both clustering and genuine distribution shift. Mixed-center folds estimate performance within the mixture of centers already observed. Leave-one-center-out instead asks how the procedure transfers to each available unseen center, although results from only a few centers should not be treated as a precise estimate for all possible sites.

**6.4. Subject-wise distributions**

A pooled LOSO score can hide clinically important heterogeneity. The locked S3 example contains 30 subjects with 20 epochs each; standard epoch-level 10-fold produces 97.0% accuracy, while leakage-free LOSO gives 426 correct epochs out of 600, or 71.0%. The positive-subject folds contain 216/300 correct positive epochs and the negative-subject folds contain 210/300 correct negative epochs. Figure 10 displays the fixed per-subject accuracies whose mean is exactly 71.0%.

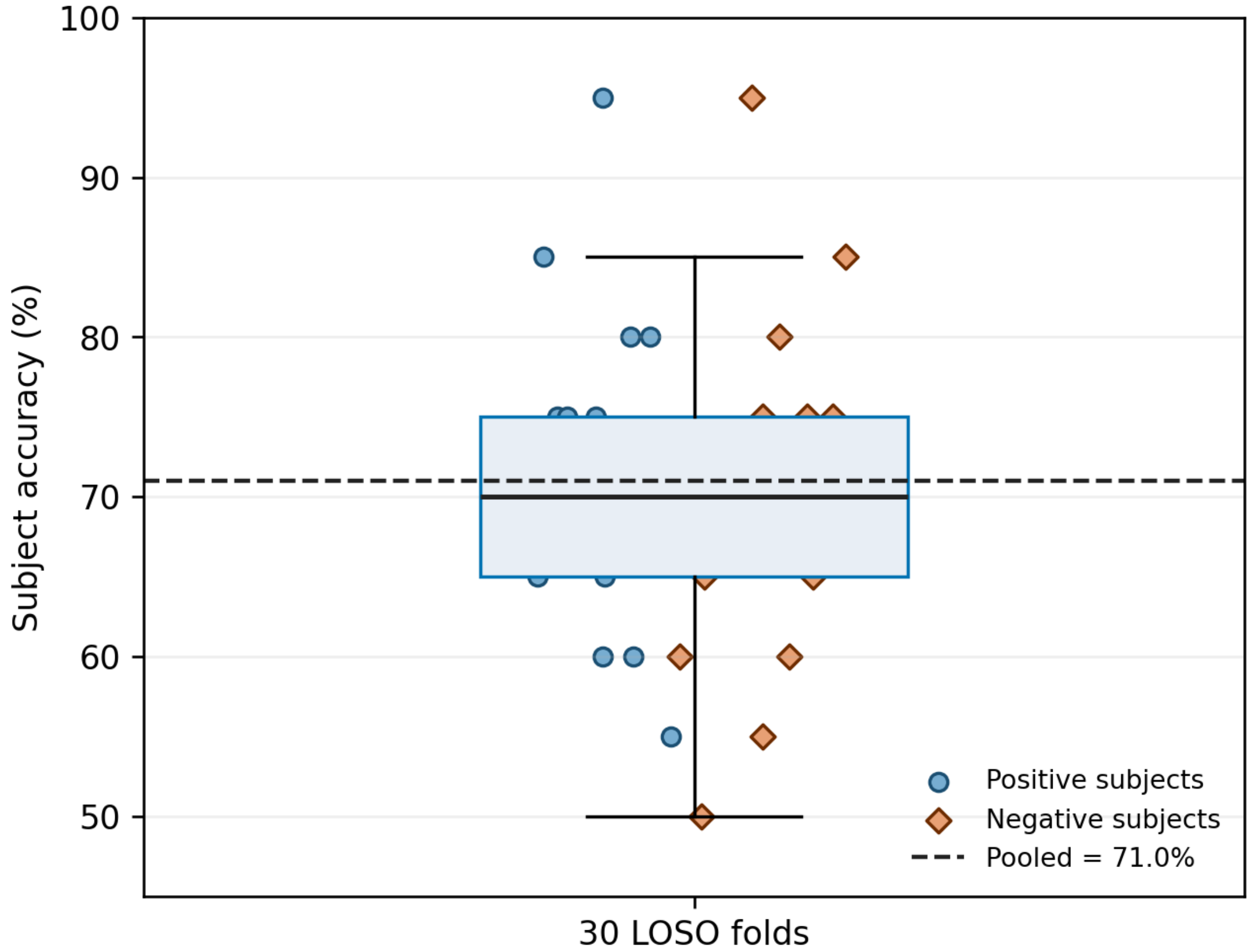

Figure 10. Fixed per-subject LOSO accuracies for the 30-subject S3 EEG-like scenario. Subjects 1–15 are positive and subjects 16–30 are negative; every subject contributes 20 epochs. The 30 accuracies range from 50% to 95%, their equal-subject mean is 71.0%, and their pooled epoch accuracy is also 71.0% because group sizes are equal.

The distribution shows that the pooled score describes no single subject. Some subjects are near chance and two reach 95%, which invites investigation of signal quality, phenotype, and acquisition effects. Because every S3 subject contributes 20 epochs, pooled and equal-group accuracy coincide; this equality will not hold when  varies. Reporting all group points prevents a large number of easy rows from concealing difficult subjects or centers.

**6.5. Choosing the group-aware target**

LOSO, GroupKFold, leave-one-session-out, patient-level splitting, and leave-one-center-out are not stronger and weaker versions of one method. They estimate different deployment conditions. LOSO estimates a new subject drawn from the observed subject population. GroupKFold estimates new groups in larger assessment blocks. Leave-one-center-out estimates transport to an acquisition environment absent from training.

The correct target can require more than one analysis. A multicenter study may report mixed-center grouped cross-validation for interpolation within established centers and leave-one-center-out for transport to a new center. These estimates should not be averaged because their assessment distributions differ.

When the number of groups is small, uncertainty is governed by group heterogeneity. Ten thousand images from three centers still provide only three center-level transport experiments. In this setting, the result for each held-out center is often more informative than a narrow interval computed from image-level counts.

In summary, Group-aware validation keeps all correlated rows from one independent unit on the same side of every split. LOSO is interpretable for new-subject performance, GroupKFold reduces cost, and stratified grouping balances classes only as far as indivisible groups permit. The held-out level must match the deployment target; multiple analyses may be needed when interpolation within known centers and transport to an unseen center are both relevant. Report pooled held-out and equal-group results when group sizes differ, and show group-wise performance.

## 7. Nested cross-validation and correct pipeline construction

### 7.1. Why a second loop is needed

Ordinary cross-validation can either assess a prespecified pipeline or help choose among alternatives, but using the same fold results for both purposes reuses information. Once many configurations are compared, the best observed CV error reflects a mixture of genuine signal and favorable validation noise. Presenting that selected value as an independent estimate is therefore optimistic. Nested cross-validation separates the roles: the inner loop handles preprocessing choices, feature selection, hyperparameters, thresholds, and even model-family selection, whereas the outer loop evaluates the procedure produced by those choices.

This separation is particularly important in high-dimensional biomedical studies. If an SVM kernel, regularization value, selected-feature count, and probability threshold are jointly chosen, the object under evaluation is their selection algorithm rather than one predeclared SVM. Varma and Simon [7] demonstrated the bias created when the same CV is used for selection and error estimation, and Cawley and Talbot [8] showed that the selection criterion itself can be overfit. Nested CV reduces this source of optimism, although it cannot repair a nonrepresentative cohort, incorrect labels, or a group-leaking outer split.

### 7.2. Inner selection and outer assessment

Let $O_o$ be outer assessment fold $o$, let $D_{-o}$ be its outer training data, and let $V_{o,j}$ be inner validation fold $j$ drawn only from $D_{-o}$. For candidate configuration $\lambda \in \Lambda$, the inner loop selects

$$\hat{\lambda}_o = \underset{\lambda \in \Lambda}{\operatorname{argmin}} \left\{ \frac{1}{K_i} \sum_{j=1}^{K_i} \frac{1}{|V_{o,j}|} \sum_{i \in V_{o,j}} L\left(y_i, \mathcal{A}_\lambda(D_{-o} \setminus V_{o,j})(x_i)\right) \right\}. \tag{21}$$

In Equation (21), $K_i$ is the number of inner folds, $\Lambda$ is the candidate set, and $\mathcal{A}_\lambda$ is the complete pipeline under configuration $\lambda$. The set $V_{o,j}$ is never used to fit the corresponding inner model. The selected configuration $\hat{\lambda}_o$ may differ across outer folds because selection is repeated independently. Ties require a prespecified rule, such as choosing the simpler configuration.

After selection, $\mathcal{A}_{\hat{\lambda}_o}$ is refitted on all of $D_{-o}$ and predicts $O_o$ once. With disjoint outer folds whose union contains all $N$ independent units, the nested estimate is

$$\hat{\mathcal{R}}_{\text{nested}} = \frac{1}{N}\sum_{o=1}^{K_o}\sum_{i\in O_o} L\left(y_i, \mathcal{A}_{\hat{\lambda}_o}(D_{-o})(x_i)\right). \tag{22}$$

In Equation (22), $K_o$ is the number of outer folds, $O_o$ is the outer assessment set, and $D_{-o}$ is the corresponding outer training set. Each loss is produced by a model whose hyperparameters were selected without the outer observation. The factor $1/N$ gives an observation-pooled estimate; equal-group aggregation can replace it when groups have unequal row counts. Outer predictions, not inner selection scores, form the final confusion matrix.

### 7.3. Nested group cross-validation

When multiple rows belong to one subject, grouping is required at both levels whenever the inner candidate pipelines could exploit identity. Outer LOSO or GroupKFold estimates performance for new groups. Within each outer training set, inner StratifiedGroupKFold can approximate class balance while preserving group exclusivity. The required inner constraint is

$$\{g(i): i \in D_{o,j}^{\text{inner train}}\} \cap \{g(i): i \in V_{o,j}\} = \varnothing \quad \text{for every } o, j. \tag{23}$$

In Equation (23), $g(i)$ is the group of row $i$, $D_{o,j}^{\text{inner train}}$ is the inner training subset, and $V_{o,j}$ is inner validation. The empty intersection prevents a group from crossing the inner boundary. Ordinary stratified k-fold is valid inside an outer group loop only when each inner row is already an independent unit, such as one subject-level feature vector per person. Otherwise, “outer group-aware, inner row-stratified” design can select hyperparameters that exploit within-subject leakage even if the outer score remains group-clean.

Every learned stage of the pipeline obeys the same containment rule. As detailed in Table 9, preprocessing and feature selection are refitted for every inner candidate fit, then refitted on the complete outer training data after $\hat{\lambda}_o$ is selected. No state learned in one outer fold is reused in another.

Table 9. Responsibilities of each data partition in nested cross-validation. Data-dependent stages are fitted within the current training boundary. Inner validation chooses configurations, whereas outer assessment estimates the complete selection pipeline and must not alter it.

| Pipeline operation | Inner training fold | Inner validation fold | Full outer training data | Outer assessment fold |
|---|---|---|---|---|
| Imputation, normalization, PCA | Fit state | Apply stored state | Refit after selection | Apply outer-training state |
| Supervised feature selection | Fit/rank/select | Evaluate selected features | Refit selected rule | Apply without reranking |
| Class weights or resampling | Estimate/apply | Never resample | Re-estimate/refit | Preserve natural assessment distribution |

| Pipeline operation | Inner training fold | Inner validation fold | Full outer training data | Outer assessment fold |
|---|---|---|---|---|
| Hyperparameters and model family | Fit candidate $\lambda$ | Produce selection loss | Fit $\hat{\lambda}_o$ | Predict once |
| Calibration and decision threshold | Fit within inner training | Choose if tuned | Refit/freeze | Apply unchanged |
| Reported performance | Training diagnostic only | Selection criterion only | Not reported as test performance | Final outer prediction and metric |

Table 9 also explains the proper final refit. After nested assessment, one configuration can be selected by a prespecified rule based on inner selections or by rerunning the selection procedure on all development data. The resulting deployment model may use all development observations, but its unbiased internal estimate remains the earlier outer result. If an untouched external test set exists, it can assess that final refit once.

### 7.4. Reporting nested results

Report the outer independent unit together with $K_o$, $K_i$, the number of repetitions, the candidate space, scoring and tie rules, and every transformation fitted inside the folds. Keep the outer-fold predictions so pooled confusion matrices and subject-level results can be reconstructed. How often a particular $\hat{\lambda}_o$ is selected is useful for judging selection stability, but it is not a performance estimate. When only a few groups or centers are available, the individual outer-fold results are often more informative than a narrow interval that treats rows as independent.

Nested CV does not make an unlimited search space harmless. Increasing $|\Lambda|$ raises both computational cost and the number of chances to fit noise in the inner folds. Candidate settings should therefore be restricted to choices that are scientifically and algorithmically plausible. With small datasets, a modest, stable pipeline may be more credible than an extensive search whose selected configuration changes from one outer fold to another.

### 7.5. Final refitting after nested assessment

Nested cross-validation evaluates the full model-selection procedure, and its outer predictions are the observations that support the reported performance. Once that assessment has been completed, a deployment model can be trained on all available development data. Its final configuration may be chosen using the selection rule learned during nested CV, but the deployment refit is distinct from the outer-fold performance estimate.

The final refit does not create a new performance estimate. Its expected performance is represented by the outer results. A separate lockbox may be opened once after the pipeline and decision threshold are frozen. If the lockbox changes the model, it has become development data and a new independent assessment set is required.

### 7.6. When nesting is unnecessary

Nesting adds little when the entire pipeline was fixed before the data were examined and no hyperparameter, feature set, model family, threshold, calibration rule, or preprocessing alternative is selected from assessment results. It may also be unnecessary when a sufficiently large independent test set is used exactly once after all development choices have been frozen.

The decision should be based on selection burden, not on whether the software function is called cross-validation. A single data-driven choice can create optimism, while a fully prespecified model can be assessed honestly by a simple group-correct hold-out.

In summary, Nested validation separates model selection from performance estimation by enclosing every data-driven choice inside each outer training partition. Grouped problems require group integrity in both loops, and thresholds, calibration, feature selection, and preprocessing must be refitted at the correct level. A locked external test set can replace the outer loop only when it remains untouched throughout development.

## 8. Computational complexity analysis

### 8.1. Counting model fits

A useful first approximation to validation cost is the number of models that must be fitted. Let $R$ be repetitions, $K$ an ordinary fold count, $G$ a group count, $K_o$ and $K_i$ outer and inner fold counts, and $|\Lambda|$ candidate configurations. Leaving aside an optional final deployment fit, the principal counts are

$$M_V = \begin{cases} 1, & \text{hold-out or one fixed train/validation/test assessment,} \\ R, & \text{repeated hold-out,} \\ K, & \text{k-fold,} \\ RK, & \text{repeated k-fold,} \\ N, & \text{LOOCV,} \\ \binom{N}{p}, & \text{exhaustive leave-p-out,} \\ G, & \text{LOSO or LOGO,} \\ K_o(K_i \mid \Lambda \mid +1), & \text{nested CV with one outer refit per fold.} \end{cases} \tag{24}$$

In Equation (24), $M_V$ is the number of training calls made by validation procedure $V$. The nested expression contains $K_o K_i \mid \Lambda \mid$ inner candidate fits and $K_o$ outer refits. GroupKFold uses $K$ fits rather than $G$, and nested group CV has the same symbolic count as nested CV after its group-aware fold numbers are substituted. Repeating the entire nested procedure multiplies its count by the number of repetitions. The nested expression $K_o(K_i \mid \Lambda \mid +1)$ expands to $K_o K_i \mid \Lambda \mid$ inner candidate fits plus $K_o$ outer refits.

Fit count alone does not determine time because the cost of one fit changes with training size and learner. Table 10 gives a transparent numerical comparison using hypothetical per-fit times of 0.01 s at $N = 100$, 0.20 s at $N = 1{,}000$, and 5 s at $N = 10{,}000$. The nested row uses $K_o = 10$, $K_i = 5$, and $|\Lambda| = 20$; the LOSO row assumes $G = 20$, 100, and 500 groups.

Table 10. Symbolic fit counts translated into concrete counts and hypothetical serial runtimes. Per-fit times are illustrative and already include sample-size dependence; they are not hardware benchmarks. Leave-2-out uses N(N-1)/2 fits, LOSO uses the stated group counts, and nested CV includes ten outer refits.

| **Method** | **Fits at** $N = 100$ | **Fits at** $N = 1{,}000$ | **Fits at** $N = 10{,}000$ |
|---|---|---|---|
| Hold-out | 1 | 1 | 1 |
| 5-fold | 5 | 5 | 5 |
| 10-fold | 10 | 10 | 10 |
| Repeated 10 × 10-fold | 100 | 100 | 100 |
| LOOCV | 100 | 1,000 | 10,000 |
| Leave-2-out | 4,950 | 499,500 | 49,995,000 |
| LOSO under stated $G$ | 20 | 100 | 500 |
| Nested 10 × 5 with 20 candidates | 1,010 | 1,010 | 1,010 |

| **Method** | **Hypothetical time at** $N = 100$ | **At** $N = 1{,}000$ | **At** $N = 10{,}000$ |
|---|---|---|---|
| Hold-out | 0.01 s | 0.20 s | 5.00 s |
| 5-fold | 0.05 s | 1.00 s | 25.00 s |
| 10-fold | 0.10 s | 2.00 s | 50.00 s |
| Repeated 10 × 10-fold | 1.00 s | 20.00 s | 8.33 min |
| LOOCV | 1.00 s | 3.33 min | 13.89 h |
| Leave-2-out | 49.50 s | 1.16 d | 7.92 y |
| LOSO under stated $G$ | 0.20 s | 20.00 s | 41.67 min |
| Nested 10 × 5 with 20 candidates | 10.10 s | 3.37 min | 1.40 h |

Table 10 exposes two different explosions. LOOCV grows linearly in $N$, while exhaustive leave-p-out grows combinatorially. A fixed-fold nested design has a fit count independent of $N$ but each individual fit becomes more expensive as $N$ grows. Wall time can also be dominated by feature extraction or deep-network initialization rather than the classifier call itself.

### 8.2. Base-learner complexity

If fit $m$ uses $n_m$ training observations and $D$ features, total training cost is more accurately written as

$$\mathcal{C}_{\text{total}} = \sum_{m=1}^{M_V} \mathcal{C}_{\mathcal{A}}(n_m, D) \approx M_V\, \mathcal{C}_{\mathcal{A}}(\bar{n}, D), \tag{25}$$

where $\mathcal{C}_{\mathcal{A}}(n_m, D)$ is the cost of the complete pipeline in fit $m$ and $\bar{n}$ is a representative training size. Equation (25) is an approximation when fold sizes differ or early stopping changes the number of epochs. Feature selection, calibration, and inference on held-out observations must be added when material. Table 11 gives commonly used order-level learner costs under explicit simplifying assumptions.

Table 11. Approximate learner complexity within a validation design. Here n is the training size in one fit, D is the feature dimension, T is the number of trees, E is the epoch count, and C_net is the fixed per-example forward/backward cost of the network. Constants, sparsity, solver convergence, hardware, and indexing can substantially change realized time.

| Learner | Approximate cost for one fit or evaluation | Validation implication |
|---|---|---|
| Kernel SVM | Training often between $O(n^2)$ and $O(n^3)$, with feature/kernel constants and approximately $O(n^2)$ kernel storage | Nested grids become prohibitive quickly; cache reuse and smaller candidate sets matter |
| kNN | Minimal model fitting beyond storing/indexing $O(nD)$; naive prediction is $O(nD)$ per query | Validation time is dominated by predicting all held-out rows, approximately $O(n_{te}nD)$ per fold |
| Random forest | About $O\left(Tn\log n\sqrt{D}\right)$ for $T$ balanced classification trees with $\sqrt{D}$ candidate features per split | Multiply by folds and candidates; tree and feature-level parallelism are available |
| Deep network | For fixed architecture, one epoch is $O(n\,C_{net})$; $E$ epochs cost $O(En\,C_{net})$ | Nested CV multiplies complete training trajectories; warm starts can bias comparisons if state crosses folds |

Table 11 illustrates why an identical fit count can correspond to milliseconds for a linear classifier but months for a three-dimensional network. kNN is not cost-free either; much of its computation is deferred from fitting to prediction on held-out cases. Early stopping can shorten deep-model training, but it also makes the validation curve part of the learning procedure. Parallel execution reduces elapsed time, not the statistical number of models that were fitted.

### 8.3. Worked nested-grid calculation and acceleration

Consider five hyperparameters, each with four candidate values, searched over the complete Cartesian product. The total number of configurations is

$$|\Lambda| = 4 \times 4 \times 4 \times 4 \times 4 = 4^5 = 1{,}024. \qquad (26)$$

Summing the five grid sizes gives 20 only if the parameters are varied one at a time. Evaluating the full Cartesian product is a different calculation and requires 1,024 distinct configurations.

$$M_{\text{nested}} = 10(5 \times 1{,}024 + 1) = 51{,}200 + 10 = 51{,}210. \qquad (27)$$

Equation (27) corresponds to 51,200 inner-loop fits for comparing candidate configurations, plus 10 refits of the selected configurations on the outer training folds. Training one final deployment model on all development data would increase the grand total to 51,211, but that last fit does not contribute to the nested performance estimate. Table 12 therefore uses 51,210 assessment-related fits in the computational comparison.

Table 12. Computational shortcuts for the five-hyperparameter nested-grid example. Random- and Bayesian-search counts are shown per outer fold and include the ten outer refits. Percentage reductions refer to fitted models; parallel execution changes elapsed time, not the number of fits.

| Strategy | Fits | Reduction from 51,210 | Main trade-off |
|---|---|---|---|
| Full grid, 5 inner folds | 51,210 | 0% | Exhaustive but often wasteful |
| Random search, 100 configurations | 5,010 | 90.2% | May miss narrow useful regions |
| Bayesian search, 60 evaluations | 3,010 | 94.1% | Sequential overhead and optimizer assumptions |
| Coarse-to-fine, 64 + 16 configurations | 4,010 | 92.2% | Refinement rule must be confined to the inner loop |
| Full grid with 3 inner folds | 30,730 | 40.0% | Noisier inner comparison |
| Prespecified screening to 128 candidates | 6,410 | 87.5% | Screening must not use outer outcomes |
| Eight parallel workers | 51,210 | 0% fits; ideally 87.5% wall-time reduction | Real speedup is commonly below 8× owing to overhead and unequal jobs |

Random search is attractive when only some hyperparameters strongly affect performance, as discussed by Bergstra and Bengio [21]. Bayesian optimization may reduce expensive evaluations by modeling the inner objective, but it must restart or remain statistically isolated within each outer training set; Snoek et al. [22] provide a foundational treatment. Coarse-to-fine search and candidate pre-screening are safe when refinement decisions use inner results or prespecified domain knowledge. Reusing outer-fold outcomes to prune the search would reduce cost by reintroducing leakage.

### 8.4. Fast LOOCV for selected models

The count $M_V = N$ is not always equivalent to $N$ complete refits. For ordinary least squares with full-rank design matrix $X$, fitted values $\hat{y} = Hy$, hat matrix $H = X(X^\top X)^{-1}X^\top$, residual $e_i = y_i - \hat{y}_i$, and leverage $h_{ii} = H_{ii}$, the exact deleted residual and LOOCV mean squared error are

$$e_{(-i)} = \frac{e_i}{1 - h_{ii}}, \qquad \hat{\mathcal{R}}_{\text{LOO,MSE}} = \frac{1}{N}\sum_{i=1}^{N}\left(\frac{e_i}{1 - h_{ii}}\right)^2. \tag{28}$$

In Equation (28), $e_{(-i)}$ is the prediction residual for observation $i$ from the model fitted without it, and $h_{ii}$ is its leverage in the full fit. After one factorization and the leverage diagonal are available, each deleted residual costs $O(1)$ additional arithmetic and all $N$ values cost $O(N)$ rather than $N$ independent factorizations. This PRESS identity is exact for ordinary linear regression [23]. Related inverse-update identities exist for selected ridge and kernel methods, but they do not make arbitrary feature selection or deep pipelines free.

### 8.5. Scaling and the value of nesting

Figure 11 compares fit-count growth on logarithmic axes. Hold-out, fixed 10-fold, repeated 10 × 10-fold, and a fixed 1,010-fit nested design form horizontal lines; LOOCV grows linearly, and leave-2-out approaches quadratic growth.

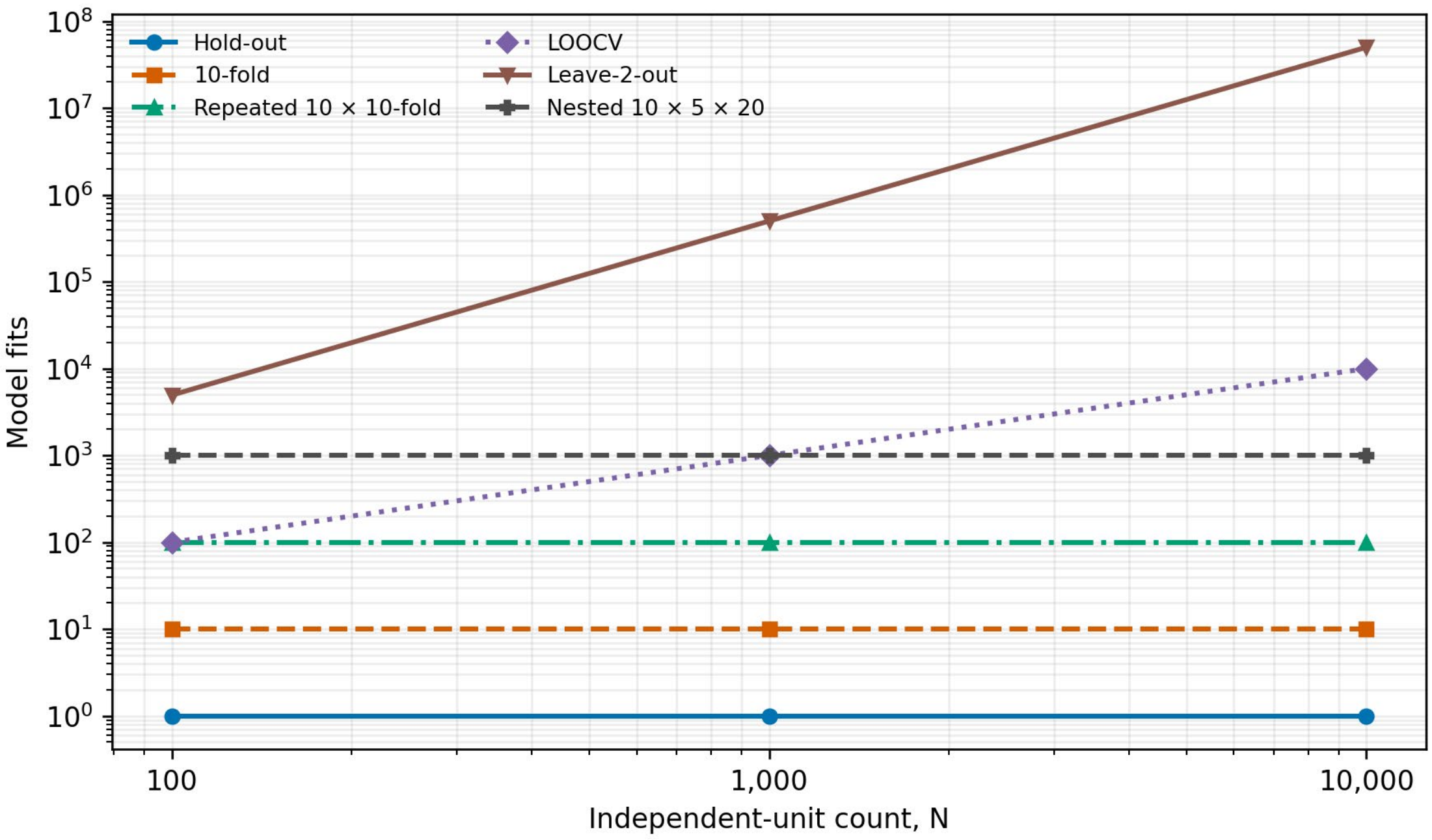

Figure 11. Log-scale model-fit growth for hold-out, 10-fold, repeated 10 × 10-fold, LOOCV, leave-2-out, and nested 10 × 5 CV with 20 candidates. Curves are listed in that order. Values are exact base-10 logarithms of fit counts for $N = 100$, 1,000, and 10,000; per-fit learner cost is analyzed separately in Table 11.

Nested CV earns its computational cost when selection is extensive, the data set is small or high-dimensional, or the claim is important enough to require a careful internal estimate. It adds less value when the pipeline is already frozen and a large, representative, untouched test set is available. For expensive deep models, one group-correct train/validation/test split with a limited number of optimization seeds may be a reasonable compromise. The appropriate choice depends on the scientific cost of optimism as well as the computational cost of preventing it.

### 8.6. Practical strategies for reducing cost

The practical goal is to reduce computation without allowing information to cross the outer assessment boundary.

- Parallel execution can shorten wall-clock time because folds and candidate configurations are usually independent jobs. The number of statistical fits, however, is unchanged.
- A large Cartesian grid can often be replaced by random search or Bayesian optimization, provided that the search remains entirely inside each outer training partition.
- A three-fold inner loop is often adequate for ranking candidates while the outer loop remains at five or 10 folds. Moving from five inner folds to three reduces the inner fit count by 40%.

• Coarse-to-fine search, successive halving, early stopping, warm starts, and regularization paths can discard weak candidates before they incur the cost of full training.

• Feature extraction may be reused when the extractor was fixed independently of the current labels. If the representation is supervised or fine-tuned, it must be learned again inside the appropriate training boundary.

• With deep learning, a frozen backbone, a small number of optimization seeds, and one group-correct train/validation/test split may offer a defensible compromise when full nested assessment is computationally prohibitive.

### 8.7. When computational cost is justified

Many of the most important corrections cost very little. Replacing row-wise folds with GroupKFold generally leaves the number of model fits unchanged, and fitting a scaler separately inside each fold adds almost no computational burden. Leave-one-center-out can even require fewer fits than mixed-center 10-fold CV. In each case, bias is reduced by repairing the information boundary rather than by enlarging the search.

Nested cross-validation can be costly when many candidates must be evaluated on a small number of independent units. That cost should be considered alongside the risk of selecting a model on the same data used to assess it. With limited compute, a focused search is often sensible; assessment data still need to remain outside selection.

The dominant cost is usually the number of times the complete learner is fitted and the cost of each fit. Correct grouping and fold-specific preprocessing are comparatively cheap. A large nested search can often be reduced by narrowing the candidate space or using random search, early stopping, warm starts, or parallel execution, as long as the outer assessment data remain isolated from selection. Under tight compute budgets, preserving that boundary matters more than adding extra folds.

## 9. Scenario-based experimental study

Section 9 uses seven fixed, hand-checkable examples and one reproducible simulation. They are teaching constructions specified in Table 13, not clinical findings or benchmark results. Fixed confusion counts reproduce all values for S1-S7; S8 instead evaluates newly generated datasets and is implemented in Listings A4 and B2. In the binary cases, the minority or clinically relevant condition is treated as positive. Each scenario follows the same structure: setup, flawed procedure, corrected procedure, change in results, and the lesson it illustrates.

Table 13 collects the dimensions, class counts, grouping variables, fixed confusion counts or simulation seed, and deployment target for each scenario. N_row is the number of records supplied to the software and can be much larger than the number of independent units. S2 deliberately uses 10,000 noise features to create many chances for accidental association; the other examples use more moderate feature sets. The scenario descriptions and appendices follow these specifications.

Table 13. Specifications for scenarios S1-S8. Class counts are negative/positive unless stated otherwise. S7 uses 600 development observations, a repeatedly inspected 200-observation test set, and a separate 200-observation lockbox. S1-S7 use fixed confusion counts; bootstrap seeds apply only to the reported intervals. S8 is simulation-based.

| Scenario | Domain and row unit | N_row | D | Class distribution | Independent groups or batch structure | Locked object | Intended target |
|---|---|---|---|---|---|---|---|
| S1 | Generic tabular; observation | 120 | 20 | 100 negative / 20 positive | Not applicable | Fold/repetition tables | New independent observation |
| S2 | High-dimensional screening; observation | 60 | 10,000 | 30 negative / 30 positive; random labels | Not applicable | Fold tables | New observation after complete selection pipeline |
| S3 | EEG-like; epoch | 600 | 64 | 300 negative / 300 positive epochs | 30 subjects × 20 epochs | Fold/subject tables + bootstrap seed 20260803 | New subject |
| S4 | OCT-like; eye | 400 | 128 | 200 negative / 200 positive eyes | 200 patients × 2 eyes | Fold tables | New patient |
| S5 | Multicenter imaging; examination | 600 | 40 | 300 negative / 300 positive | 3 centers × 200 examinations | Fold/center tables | Unseen center |
| S6 | Batch-shifted signals; observation | 240 | 50 | 120 negative / 120 positive | 4 batches × 60 rows (batch structure; not used as the partitioning unit in S6) | Fold tables | New observations after fold-local normalization |
| S7 | Repeated model selection; observation | 1,000 | 50 | Development 300/300; test 100/100; lockbox 100/100 | Not applicable | Two matrices + bootstrap seed 20260807 | Fresh locked test population |

| Scenario | Domain and row unit | N_row | D | Class distribution | Independent groups or batch structure | Locked object | Intended target |
|---|---|---|---|---|---|---|---|
| S8 | Small independent-row study; observation | 80 | 20 | 40 negative / 40 positive per replicate | Not applicable (replicate studies) | Simulation seed 20260808 | Expected performance across regenerated studies |

The values in Table 13 show why stored rows should not be read as independent sample size. S3 has 600 rows but only 30 independent units, while S5 represents only three deployment environments. Each fixed object was built to demonstrate one mechanism rather than the full sampling behavior of a validation method. Listings A3 and A4 make the reported values directly auditable.

Panel B gives the mean, SD, and observed range across assessment partitions. We do not convert the fold SD into a t confidence interval because the folds share training data and are not independent replicates. Table 35 identifies this exact conversion as a reviewer warning sign. For grouped designs, any reported interval is obtained by resampling the independent units (Section 9.11).

### 9.1. S1—small and imbalanced data

Setup. S1 represents a binary screening task in which only 20 of 120 observations are positive. A logistic classifier uses 20 standardized predictors, and the learner, threshold, and preprocessing recipe are otherwise fixed so that split randomness is the variable under study. The incorrect design repeats an unstratified 15% hold-out, producing test sets of only 18 observations with an unstable number of positives. The correct design uses 20 repetitions of stratified five-fold CV, preserving approximately four positives and 20 negatives in every test fold, as summarized in Table 14.

Table 14. Setup for S1, a small imbalanced binary-classification example. Both designs use leakage-safe preprocessing, so the comparison isolates instability introduced by a very small, unstratified assessment sample.

| Component | Incorrect design | Correct design |
|---|---|---|
| Split rule | 20 unstratified random 85/15 hold-outs | 20 repetitions of stratified five-fold CV |
| Assessment size | 18 observations per repetition | All 120 observations tested once per repetition |
| Expected positive count per elementary test set | 3 on average, but not constrained | 4 in each fold |
| Preprocessing | Training-set mean and SD | Fold-training mean and SD |
| Selection | None; fixed logistic model | None; fixed logistic model |
| Primary comparison | Split-to-split metric spread | Repetition-to-repetition metric spread |

Incorrect method, correct method, and result difference. Across the 20 hold-outs, accuracies of 77.8%, 83.3%, 88.9%, and 94.4% occur 2, 5, 8, and 5 times, respectively, yielding a mean of 87.8% and an SD of 5.3 points. Table 15 also shows that the best split contains only one positive assessment case. Repeated stratified five-fold CV reduces the accuracy SD to 1.2 points. Panel B keeps the pooled results separate from the repetition-level mean, SD, and observed min-max range. These repetition summaries are descriptive because the resamples overlap.

Table 15. Confusion matrices and performance comparison for S1. Panel A contrasts the lowest-accuracy hold-out, the highest-accuracy hold-out, and one representative repetition of stratified five-fold CV. Panel B separates metrics from pooled held-out predictions from the repetition-level mean ± SD and observed [min, max].

| Actual class | Worst hold-out: predicted + | Worst hold-out: predicted − | Best hold-out: predicted + | Best hold-out: predicted − | Stratified five-fold: predicted + | Stratified five-fold: predicted − |
|---|---|---|---|---|---|---|
| Positive | 3 (TP) | 2 (FN) | 1 (TP) | 0 (FN) | 12 (TP) | 8 (FN) |
| Negative | 2 (FP) | 11 (TN) | 1 (FP) | 16 (TN) | 6 (FP) | 94 (TN) |

| Design and summary | Accuracy, % | Sensitivity, % | Specificity, % | F1, % | MCC |
|---|---|---|---|---|---|
| Unstratified hold-out — pooled held-out predictions | 87.8 | 59.6 | 93.1 | 60.7 | 0.535 |
| Unstratified hold-out — per-repetition mean ± SD [min, max] | 87.8 ± 5.3 [77.8, 94.4] | 60.9 ± 18.7 [25.0, 100.0] | 92.9 ± 5.7 [84.6, 100.0] | 61.0 ± 9.2 [33.3, 75.0] | 0.565 ± 0.118 [0.236, 0.721] |
| Stratified five-fold — pooled held-out predictions | 88.3 | 60.0 | 94.0 | 63.2 | 0.564 |
| Stratified five-fold — per-repetition mean ± SD [min, max] | 88.3 ± 1.2 [85.8, 90.8] | 60.0 ± 4.3 [50.0, 65.0] | 94.0 ± 0.9 [93.0, 96.0] | 63.1 ± 4.0 [54.1, 70.3] | 0.564 ± 0.048 [0.460, 0.652] |

Inference. Stratification cannot add positive cases or resolve biological uncertainty. In S1, its narrower role is to reduce avoidable variation in class composition while allowing every observation to be assessed across repetitions. The gain is a sensitivity estimate that depends less on one favorable partition, not a higher average accuracy. The isolated 94.4% hold-out result is especially fragile because its sensitivity is determined by a single positive assessment case.

### 9.2. S2 - feature-selection leakage in a random-label control

**Setup.** S2 is intentionally designed as a strong negative control. Sixty balanced labels are generated independently of 10,000 noise features, so the data contain no reproducible predictive signal. A univariate score is used to rank features, a linear SVM is the classifier, and the candidate feature counts are $m \in \{10,20,50\}$ with $C_{SVM} \in \{0.01,0.1,1,10\}$. In the flawed workflow, all 10,000 variables are ranked once using the full set of labels, the most favorable variables are retained, and the SVM is then cross-validated. The leakage-safe workflow repeats both ranking and scaling inside every inner training partition of nested 10 × 5 CV, as defined in Table 16.

Table 16. Setup of S2, a high-dimensional random-label negative control. The only substantive difference is whether feature selection and model tuning are enclosed within the resampling boundary used to estimate performance.

| Component | Leaky workflow | Leakage-free workflow |
|---|---|---|
| Data | $N = 60, D = 10{,}000$, 30/30 classes; $X \perp Y$ | Identical fixed data and seed |
| Feature ranking | Once on all 60 labeled observations | Recomputed using inner-training observations only |
| Hyperparameters | Favorable $m$ and $C_{SVM}$ effectively judged with reused data | $m$ and $C_{SVM}$ selected in inner five-fold CV |
| Performance estimate | Stratified 10-fold CV after global screening | Outer stratified 10-fold predictions, untouched by inner selection |
| Null expectation | Should be near 50%, but leakage exploits extreme noise correlations | Approximately 50% over repeated null data sets |

**Incorrect method, correct method, and result difference.** With 10,000 chances to correlate with 60 random labels, global screening can select features whose accidental association already incorporates every future test label. The leaky matrix in Table 17 reports 56 correct predictions and an apparently impressive 93.3% accuracy, despite the construction $X \perp Y$. Once screening and tuning are nested, the outer predictions contain 31 correct cases, giving 51.7% accuracy and MCC 0.033. This collapse is the numerical signature expected when selection bias, rather than signal, generated the original result [6, 7].

Table 17. Side-by-side confusion matrices and metric comparison for S2. Panel A contains pooled outer predictions for the leaky and nested workflows. Panel B reports pooled metrics separately from the per-fold mean ± SD and observed [min, max] across the ten outer folds. Because the labels and predictors are independent by construction, the 93.3% result is necessarily artifactual, while the nested result is compatible with chance.

| Actual class | Leaky: predicted + | Leaky: predicted − | Nested: predicted + | Nested: predicted − |
|---|---|---|---|---|
| Positive | 28 (TP) | 2 (FN) | 15 (TP) | 15 (FN) |
| Negative | 2 (FP) | 28 (TN) | 14 (FP) | 16 (TN) |

| Workflow and summary | Accuracy, % | Sensitivity, % | Specificity, % | F1, % | MCC |
|---|---|---|---|---|---|

| Workflow and summary | Accuracy, % | Sensitivity, % | Specificity, % | F1, % | MCC |
|---|---|---|---|---|---|
| Feature selection before CV — pooled held-out predictions | 93.3 | 93.3 | 93.3 | 93.3 | 0.867 |
| Feature selection before CV — per-fold mean ± SD [min, max] | 93.3 ± 8.6 [83.3, 100.0] | 93.3 ± 14.1 [66.7, 100.0] | 93.3 ± 14.1 [66.7, 100.0] | 93.1 ± 9.1 [80.0, 100.0] | 0.883 ± 0.151 [0.707, 1.000] |
| Nested feature selection and tuning — pooled outer predictions | 51.7 | 50.0 | 53.3 | 50.8 | 0.033 |
| Nested feature selection and tuning — per-fold mean ± SD [min, max] | 51.7 ± 9.5 [33.3, 66.7] | 50.0 ± 17.6 [33.3, 66.7] | 53.3 ± 23.3 [33.3, 100.0] | 49.9 ± 10.8 [33.3, 66.7] | 0.045 ± 0.211 [−0.333, 0.447] |

Inference. The large gap in this high-dimensional example comes from feature selection being performed with all labels before cross-validation. That lets assessment information shape the feature set, and cross-validating only the classifier cannot remove the contamination. The selector itself must be refitted inside each training fold. If accuracy remains high after label permutation, both the full pipeline and the permutation procedure should be checked. Figure 8 contains the fixed matrices used in this comparison.

### 9.3. S3—epoch leakage in subject-dependent EEG

Setup. S3 has 30 subjects, 20 EEG-like epochs per subject, and 64 features. Fifteen subjects per class produce 600 balanced rows but only 30 independent units. Epoch-level 10-fold CV mixes subjects across training and test. LOSO withholds all 20 epochs of one subject and fits every learned step on the other 29, as Table 18 shows.

Table 18. Setup of S3, the EEG-like repeated-measure scenario. Epoch-level CV answers an easier, subject-overlapping question; LOSO aligns the independent unit with deployment to a previously unseen subject.

| Component | Epoch-level 10-fold | LOSO |
|---|---|---|
| Unit used for partitioning | Individual epoch | Subject identifier |
| Elementary test set | 60 epochs, usually drawn from all subjects | 20 epochs from one unseen subject |
| Training/test subject overlap | Extensive | None |
| Number of outer fits | 10 | 30 |
| Deployment question answered | New epoch from a largely known subject population | New subject |

| Component | Epoch-level 10-fold | LOSO |
|---|---|---|
| Uncertainty summary | Fold metrics, used descriptively | Pooled matrix plus subject-level distribution and stratified subject bootstrap |

Incorrect method, correct method, and result difference. Subject-specific spectral baselines and acquisition signatures allow an epoch classifier to recognize a person and then reuse that person's label when sibling epochs appear in training. Accordingly, the epoch-level matrix in Table 19 contains only 18 errors and reports 97.0% accuracy. Under LOSO, false negatives rise to 84 and false positives to 90, producing 71.0% accuracy and MCC 0.420. The 26-point gap does not show that LOSO damages the model; it shows that epoch-level CV estimated a different, easier prediction task [3].

Table 19. Confusion matrices and performance results for S3. Epoch-level performance is reported both from pooled predictions and as the per-fold mean ± SD with the observed [min, max]. For LOSO, the table gives the pooled point estimate with a stratified subject-bootstrap interval; the distribution of subject-level accuracies is reported separately.

| Actual class | Epoch 10-fold: predicted + | Epoch 10-fold: predicted − | LOSO: predicted + | LOSO: predicted − |
|---|---|---|---|---|
| Positive epochs | 291 (TP) | 9 (FN) | 216 (TP) | 84 (FN) |
| Negative epochs | 9 (FP) | 291 (TN) | 90 (FP) | 210 (TN) |

| Design and summary | Accuracy, % | Sensitivity, % | Specificity, % | F1, % | MCC |
|---|---|---|---|---|---|
| Epoch-level 10-fold — pooled held-out predictions | 97.0 | 97.0 | 97.0 | 97.0 | 0.940 |
| Epoch-level 10-fold — per-fold mean ± SD [min, max] | 97.0 ± 0.7 [96.7, 98.3] | 97.0 ± 1.1 [96.7, 100.0] | 97.0 ± 1.1 [96.7, 100.0] | 97.0 ± 0.7 [96.7, 98.4] | 0.940 ± 0.014 [0.933, 0.967] |
| LOSO — pooled estimate ± subject-bootstrap SD [95% percentile CI] | 71.0 ± 1.96 [67.3, 75.0] | 72.0 ± 2.62 [67.0, 77.3] | 70.0 ± 2.89 [64.3, 75.7] | 71.3 ± 1.96 [67.7, 75.2] | 0.420 ± 0.039 [0.347, 0.500] |

Inference. LOSO accuracy for positive subjects ranges from 55% to 95%, and for negative subjects from 50% to 95%. Across all 30 subjects, the mean is 71.0%, the SD is 10.94 percentage points, and the observed range is 50.0-95.0%. This between-subject spread disappears in the pooled confusion matrix; Figure 10 shows it directly with individual accuracies and a box plot. For a new-subject claim, those 30 people are the independent evidence. Splitting their recordings into more epochs does not create more subjects.

### 9.4. S4—two eyes, one patient

Setup. S4 contains 400 OCT eye records from 200 patients, with two eyes and 128 derived image features per patient. Diagnosis is assigned at the patient level, and fellow eyes share anatomy, genetics, disease history, and usually the acquisition setting. In the flawed analysis, folds are assigned eye by eye, so a patient's other eye can remain in training. The corrected analysis keeps both eyes together and balances patient labels across ten folds, as specified in Table 20.

Table 20. Setup for S4, the paired-eye OCT example. Under patient-grouped validation, the two correlated eyes remain an indivisible unit, while both eye-level predictions are still retained in the pooled performance matrix.

| Component | Eye-level split | Patient-grouped split |
|---|---|---|
| Row/class counts | 400 eyes; 200 positive and 200 negative | Same |
| Independent units | Incorrectly treated as 400 | 200 patients |
| Fold assignment | Each eye assigned independently | Both eyes assigned through one patient ID |
| Fellow-eye overlap | Possible and frequent | Impossible |
| Intended use emulated | A new eye from a possibly represented patient | A new patient with one or two eyes |
| Recommended implementation | Not appropriate for new-patient claims | GroupKFold or StratifiedGroupKFold by patient |

Incorrect method, correct method, and result difference. Shared patient signatures let an eye-level model partially recognize a patient's fellow eye, yielding 95.0% accuracy in the leaky matrix of Table 21. Patient grouping removes that shortcut and produces 81.5% accuracy, 80.0% sensitivity, 83.0% specificity, and MCC 0.630. The 13.5-point accuracy difference is compatible with strong within-patient resemblance rather than a change in the classifier itself. Because both matrices contain the same 200 positive and 200 negative eyes, their metric contrast is not driven by class prevalence.

Table 21. Confusion matrices and performance comparison for S4. Panel A pools all 400 eye-level predictions. Panel B distinguishes pooled metrics from the per-fold mean ± SD and observed [min, max] across ten folds. Grouping by patient reduces the apparent performance obtained when fellow eyes are allowed to cross the split.

| Actual class | Eye-level: predicted + | Eye-level: predicted − | Patient-grouped: predicted + | Patient-grouped: predicted − |
|---|---|---|---|---|
| Positive eyes | 191 (TP) | 9 (FN) | 160 (TP) | 40 (FN) |
| Negative eyes | 11 (FP) | 189 (TN) | 34 (FP) | 166 (TN) |

| Design and summary | Accuracy, % | Sensitivity, % | Specificity, % | F1, % | MCC |
|---|---|---|---|---|---|
| Random eye-level 10-fold — pooled held-out predictions | 95.0 | 95.5 | 94.5 | 95.0 | 0.900 |
| Random eye-level 10-fold — per-fold mean ± SD [min, max] | 95.0 ± 1.2 [92.5, 97.5] | 95.5 ± 1.6 [95.0, 100.0] | 94.5 ± 1.6 [90.0, 95.0] | 95.0 ± 1.2 [92.7, 97.6] | 0.900 ± 0.024 [0.851, 0.951] |
| Patient-grouped 10-fold — pooled held-out predictions | 81.5 | 80.0 | 83.0 | 81.2 | 0.630 |
| Patient-grouped 10-fold — per-fold mean ± SD [min, max] | 81.5 ± 4.4 [75.0, 90.0] | 80.0 ± 5.8 [70.0, 90.0] | 83.0 ± 3.5 [80.0, 90.0] | 81.2 ± 4.8 [73.7, 90.0] | 0.631 ± 0.088 [0.503, 0.800] |

Inference. Leaving out one eye is defensible only for the narrow target of predicting another eye from a patient already represented in training, and that target should be stated plainly. Most diagnostic generalization claims, however, concern patients who were never represented in training; for that target, the patient or patient group is the appropriate split boundary. Metrics may still be calculated per eye, but interval estimation should respect patient clustering or use patient-level resampling. The same reasoning extends to bilateral images, multiple lesions, repeated blood smears, and multiple scans from a single encounter.

### 9.5. S5—multicenter device shift

Setup. S5 contains three balanced centers with 200 examinations each. Several of the 40 features are shifted by center-specific device effects. Mixed-center 10-fold CV is therefore unsuitable for a claim about deployment in a new center, because every assessment fold is trained on all three acquisition environments. Leave-one-center-out provides the appropriate stress test for that target, although three centers are still too few to represent the full population of possible sites. Table 22 defines the two validation designs and their center-level independent units.

Table 22. Setup for S5, the three-center distribution-shift example. Each center contributes 100 positive and 100 negative examinations, so equal center size removes prevalence weighting from the pooled-accuracy comparison.

| Component | Mixed-center 10-fold | Leave-one-center-out |
|---|---|---|
| Rows and classes | 600 examinations; 300 positive / 300 negative | Same |
| Environment in training | All three centers | Two centers |

| Component | Mixed-center 10-fold | Leave-one-center-out |
|---|---|---|
| Environment in assessment | Same center mixture as training | One unseen center |
| Outer fits | 10 | 3 |
| Main risk | Learns device/site shortcuts | High uncertainty from only three centers |

Result difference. Mixed-center CV yields 93.0% pooled accuracy. When each center is held out in turn, the accuracies are 89.0%, 80.0%, and 61.0%, and Table 23 gives the corresponding matrices. Pooling the leave-one-center-out predictions gives 76.7% accuracy. Because only three centers are available, the center-level result is summarized by mean ± SD and the observed min–max range. A Student-t interval is unstable with so few centers and can extend beyond the valid range of the metric, making the observed range more informative for this example.

Table 23. Confusion matrices and metric comparison for S5. Panel A reports the mixed-center, center-specific, and pooled counts. Panel B distinguishes pooled metrics from the mean ± SD and observed [min, max] across the ten mixed folds or the three held-out centers. Pooled leave-one-center-out sensitivity, specificity, F1, and MCC are 74.3%, 79.0%, 76.1%, and 0.534, respectively.

| Assessment | TP | FN | FP | TN | Accuracy, % |
|---|---|---|---|---|---|
| Mixed-center 10-fold, pooled | 276 | 24 | 18 | 282 | 93.0 |
| Center 1 held out | 88 | 12 | 10 | 90 | 89.0 |
| Center 2 held out | 78 | 22 | 18 | 82 | 80.0 |
| Center 3 held out | 57 | 43 | 35 | 65 | 61.0 |
| Leave-one-center-out, pooled | 223 | 77 | 63 | 237 | 76.7 |

| Design and summary | Accuracy, % | Sensitivity, % | Specificity, % | F1, % | MCC |
|---|---|---|---|---|---|
| Mixed-center 10-fold — pooled held-out predictions | 93.0 | 92.0 | 94.0 | 92.9 | 0.860 |
| Mixed-center 10-fold — per-fold mean ± SD [min, max] | 93.0 ± 1.1 [91.7, 95.0] | 92.0 ± 1.7 [90.0, 93.3] | 94.0 ± 1.4 [93.3, 96.7] | 92.9 ± 1.1 [91.5, 94.9] | 0.860 ± 0.021 [0.834, 0.901] |
| Leave-one-center-out — pooled held-out predictions | 76.7 | 74.3 | 79.0 | 76.1 | 0.534 |

| Design and summary | Accuracy, % | Sensitivity, % | Specificity, % | F1, % | MCC |
|---|---|---|---|---|---|
| Leave-one-center-out — per-center mean ± SD [min, max] | 76.7 ± 14.3 [61.0, 89.0] | 74.3 ± 15.8 [57.0, 88.0] | 79.0 ± 12.8 [65.0, 90.0] | 76.0 ± 15.1 [59.4, 88.9] | 0.534 ± 0.286 [0.221, 0.780] |

Inference. The 32-point difference across centers exposes a transport problem hidden by the 93.0% mixed-center result. Center 3 should therefore be examined on its own, including device, protocol, population, and labeling differences, rather than absorbed into a pooled score. Leave-one-center-out does not prove performance at an unseen fourth center; it only measures sensitivity to the three observed center shifts. Multicenter reports should show center-specific matrices and acquisition metadata.

### 9.6. S6—normalization leakage

Setup. S6 uses four 60-row acquisition batches with feature-mean shifts and class counts of 45/15, 35/25, 25/35, and 15/45 negative/positive. The incorrect pipeline computes each z-score mean and SD from all 240 rows before CV, so held-out covariates help define the training coordinate system. The correct pipeline estimates these parameters in each training fold and applies them unchanged to its test fold. Table 24 isolates normalization scope while holding the classifier and ten stratified folds fixed.

Table 24. Setup of S6, the normalization-leakage scenario. Batch shifts are correlated with class prevalence, making the otherwise unsupervised global transformation informative about the held-out distribution.

| Component | Global z-score | Fold-local z-score |
|---|---|---|
| Mean and SD source | All 240 observations | Current 216-row training fold only |
| Transformation of test fold | Uses statistics partly influenced by itself | Uses stored training statistics |
| Classifier | Fixed linear SVM | Same |
| Hyperparameter search | None | None |
| Target | Artificially harmonized observed cohort | New observation processed without future data |

S6 deliberately isolates one variable. Batch membership is correlated with class prevalence, so a batch-aware split would also be defensible and would change the estimate further; that mechanism is the subject of S5. Here the classifier, the fold structure, and the stratification are held fixed so that the only difference between the two columns is the partition on which the z-score parameters are estimated. The corrected 82.5% should therefore be read as “normalization leakage removed,” not as “all leakage removed.”

Result difference. Table 25 reports 88.3% accuracy when normalization is performed globally and 82.5% when the normalization parameters are learned within each fold. That 5.8-point gap corresponds to 28 errors in the global pipeline and 42 in the leakage-safe pipeline. Because the synthetic batches were intentionally small and strongly shifted, the example illustrates the mechanism rather than predicting the size of the effect in a large homogeneous cohort.

Table 25. Confusion matrices and performance for S6. Panel A pools all held-out predictions. Panel B compares metrics from all 240 predictions with the mean ± SD and observed [min, max] across ten overlapping CV folds.

| Actual class | Global z-score: predicted + | Global z-score: predicted − | Fold-local z-score: predicted + | Fold-local z-score: predicted − |
|---|---|---|---|---|
| Positive | 108 (TP) | 12 (FN) | 99 (TP) | 21 (FN) |
| Negative | 16 (FP) | 104 (TN) | 21 (FP) | 99 (TN) |

| Pipeline and summary | Accuracy, % | Sensitivity, % | Specificity, % | F1, % | MCC |
|---|---|---|---|---|---|
| Global z-score before CV — pooled held-out predictions | 88.3 | 90.0 | 86.7 | 88.5 | 0.767 |
| Global z-score before CV — per-fold mean ± SD [min, max] | 88.3 ± 2.6 [83.3, 91.7] | 90.0 ± 3.5 [83.3, 91.7] | 86.7 ± 4.3 [83.3, 91.7] | 88.5 ± 2.6 [83.3, 91.7] | 0.768 ± 0.052 [0.667, 0.833] |
| Training-fold z-score — pooled held-out predictions | 82.5 | 82.5 | 82.5 | 82.5 | 0.650 |
| Training-fold z-score — per-fold mean ± SD [min, max] | 82.5 ± 1.8 [79.2, 83.3] | 82.5 ± 2.6 [75.0, 83.3] | 82.5 ± 2.6 [75.0, 83.3] | 82.5 ± 1.8 [78.3, 83.3] | 0.650 ± 0.034 [0.585, 0.667] |

Inference. Labels are not required for leakage to occur. An unsupervised transformation can still learn from the assessment distribution if it is fitted before the split. The safe approach is to estimate the transformation from the active training rows and apply that fitted state to the held-out rows. The same rule covers harmonization, imputation, PCA, and image-intensity calibration. A marked change after correcting the boundary is a reason to inspect data provenance and the entire preprocessing chain.

### 9.7. S7 - when repeated consultation erodes a test set

Setup. S7 contains 600 development observations, a balanced 200-observation reusable test set, and a separate balanced 200-observation lockbox. Fifty candidate pipelines are centered around an illustrative accuracy near 86%. In the flawed workflow, scores on the reusable test set influence pipeline selection. In the corrected workflow, selection uses development data only, and the lockbox is opened once. Table 26 marks where feedback is allowed in each procedure.

Table 26. Setup of S7, the repeated-test-selection scenario. The 91.2% expectation is the exact expected maximum of 50 independent Binomial(200, 0.86) scores; real candidate scores are usually correlated, but test feedback still selects favorable noise.

| Component | Reused-test selection | Locked assessment |
|---|---|---|
| Candidate comparisons | 50 on the same test set | 50 inside development resampling |
| Final reporting set | The same set that selected the winner | Untouched 200-case lockbox |
| Feedback from assessment | Model choice uses test scores | None before final freeze |
| Illustrative equal-performance model | True accuracy approximately 86% | Same reference accuracy |
| Expected maximum under 50 independent binomial scores | 91.2% when $N_{te} = 200$ and $p = 0.86$ | Not used for selection |

Result difference. The fixed realization in Table 27 reports 92.0% on the eroded test set but 86.0% under one-shot locked assessment, an optimism of six points. The 92.0% value is not a dishonest calculation; it is a conditional maximum selected from 50 noisy calculations. Intervals use 10,000 class-stratified bootstrap resamples with seed 20260807 and describe each displayed matrix, not the repeated selection process. A new lockbox is required after test feedback has influenced the model.

Table 27. Confusion matrices and performance comparison for S7. Values are pooled point estimates with stratified-bootstrap SD and percentile 95% intervals. Each matrix contains 100 positive and 100 negative observations.

| Actual class | Eroded test: predicted + | Eroded test: predicted − | Locked test: predicted + | Locked test: predicted − |
|---|---|---|---|---|
| Positive | 93 (TP) | 7 (FN) | 86 (TP) | 14 (FN) |
| Negative | 9 (FP) | 91 (TN) | 14 (FP) | 86 (TN) |

| Assessment | Accuracy, % | Sensitivity, % | Specificity, % | F1, % | MCC |
|---|---|---|---|---|---|
| Winner on reused test | 92.0 ± 1.92 [88.0, 95.5] | 93.0 ± 2.56 [88.0, 98.0] | 91.0 ± 2.88 [85.0, 96.0] | 92.1 ± 1.89 [88.2, 95.5] | 0.840 ± 0.038 [0.761, 0.910] |
| Frozen pipeline on lockbox | 86.0 ± 2.45 [81.0, 90.5] | 86.0 ± 3.44 [79.0, 92.0] | 86.0 ± 3.45 [79.0, 92.0] | 86.0 ± 2.48 [81.0, 90.6] | 0.720 ± 0.049 [0.620, 0.811] |

Inference. A test set stops being a test set as soon as one of its results changes a subsequent decision. Inspecting it 50 times creates 50 opportunities to select a favorable sampling fluctuation even though no observation is physically moved between files. Nested CV protects model selection only while the outer predictions remain outside the redesign loop. Version history, together with a prespecified rule for when the lockbox may be opened, makes this separation auditable.

### 9.8. S8 - LOOCV variability in small samples

Simulation setup. S8 regenerates 100 balanced data sets with N = 80, D = 20, and 40 observations per class. Standard-normal features are generated, and the first five features receive a class shift δ = 0.81. This shift was calibrated so that a linear discriminant model trained on 80 observations has 75.51% expected accuracy on a large independent sample. This value is the reference target for Panel A (LDA) only; no 75.51% population-accuracy target is assigned to the unpruned-tree panel, which is used to compare estimator variability under learner instability. The same generator is evaluated with two learners: stable linear discriminant analysis and an unstable unpruned classification tree. No hyperparameter is tuned. The simulation relies on seed 20260808 and compares LOOCV with stratified 10-fold CV across independently regenerated studies.

Table 28. Setup for S8, which compares LOOCV with stratified 10-fold CV in small regenerated samples. Both estimators are applied to exactly the same studies. Panel A uses a stable learner and Panel B an unstable learner, allowing the experiment to ask whether estimator variance changes with learner stability. The 75.51% reference target belongs only to Panel A (LDA); no prespecified population-accuracy target is assigned to Panel B.

| Component | LOOCV | Stratified 10-fold |
|---|---|---|
| Assessment observations per fold | 1 | 8 |
| Training observations per fold | 79 | 72 |
| Fits per data set | 80 | 10 |
| LDA reference target (Panel A) | 75.51% | 75.51% |
| Studies | 100 independently regenerated data sets | Same 100 data sets |
| Learner | Panel A: LDA; Panel B: unpruned tree | Same in both panels. |

Result difference. Across regenerated studies, the stable discriminant learner averages 75.5% with LOOCV and 75.3% with stratified 10-fold CV; the corresponding SDs are 5.5 and 6.0 percentage points. For the unstable tree, the means are 66.4% and 66.0%, while LOOCV varies more across studies (8.8 vs 7.5 percentage points). Table 29 reports observed ranges instead of t intervals. The 100 regenerated studies are independent replicates; folds within a single study are not.

Table 29. Repeated-study comparison for S8 under stable and unstable learners. Entries are mean ± between-study standard deviation [observed minimum, maximum] across 100 independently regenerated studies. Values were produced by the scikit-learn implementation in Listing B2 (scikit-learn 1.8.0, NumPy default_rng, seed 20260808). The MATLAB counterpart in Listing A4 reproduces the pattern and approximate magnitudes but not identical digits, because the random stream and the discriminant solver differ.

| Learner | Estimator | Representative study (closest to panel mean) | Confusion counts [TP FN FP TN] | Accuracy |
|---|---|---|---|---|
| Stable LDA | LOOCV | #36 | [31 9 11 29] | 75.0% |
| Stable LDA | Stratified 10-fold | #11 | [31 9 11 29] | 75.0% |
| Unstable tree | LOOCV | #9 | [28 12 15 25] | 66.3% |

| Learner | Estimator | Representative study (closest to panel mean) | Confusion counts [TP FN FP TN] | Accuracy |
|---|---|---|---|---|
| Unstable tree | Stratified 10-fold | #1 | [23 17 10 30] | 66.3% |

| Learner and estimator | Accuracy, % | Sensitivity, % | Specificity, % | F1, % | MCC |
|---|---|---|---|---|---|
| Panel A — stable LDA: LOOCV | 75.5 ± 5.5 [61.3, 87.5] | 75.7 ± 6.7 [60.0, 90.0] | 75.3 ± 6.3 [62.5, 90.0] | 75.5 ± 5.7 [60.8, 87.2] | 0.511 ± 0.110 [0.225, 0.751] |
| Panel A — stable LDA: stratified 10-fold | 75.3 ± 6.0 [58.8, 90.0] | 75.5 ± 7.0 [60.0, 92.5] | 75.0 ± 7.0 [57.5, 95.0] | 75.3 ± 6.1 [59.3, 89.5] | 0.507 ± 0.121 [0.175, 0.804] |
| Panel B — unstable tree: LOOCV | 66.4 ± 8.8 [45.0, 91.2] | 66.4 ± 9.8 [42.5, 97.5] | 66.4 ± 10.1 [35.0, 85.0] | 66.4 ± 8.9 [43.6, 91.8] | 0.329 ± 0.177 [−0.100, 0.832] |
| Panel B — unstable tree: stratified 10-fold | 66.0 ± 7.5 [50.0, 87.5] | 66.3 ± 9.1 [45.0, 92.5] | 65.8 ± 8.7 [45.0, 85.0] | 66.0 ± 7.7 [48.7, 88.1] | 0.322 ± 0.150 [0.000, 0.754] |

Inference. Under the stable discriminant learner, LOOCV is marginally closer to the 75.5% target and marginally less variable than stratified 10-fold, yet it requires eight times as many model fits to obtain that difference. Under the unstable tree learner fitted to the same data, the ordering of the standard deviations reverses: LOOCV varies more across regenerated studies (8.8 points) than 10-fold does (7.5 points), because each LOOCV training set differs from the full data set by one omitted observation, and an unstable rule can convert a change in which observation is omitted into a different partition of the feature space. The practical conclusion is for that reason conditional instead of categorical. LOOCV is not automatically the most reliable estimator at small sample size; its variance depends on how sensitive the learner is to one observation, and its cost is fixed at N fits regardless. Repeated stratified five- or ten-fold cross-validation gives a comparable estimate for a fraction of the compute, and the independent-unit count must still replace the row count when observations are grouped.

### 9.9. Multiclass macro and micro averaging

For class $c$, let $m_c$ denote a one-versus-rest metric based on $TP_c$, $FP_c$, $FN_c$, and $TN_c$. Macro averaging then gives every class the same weight:

$$m_{\text{macro}} = \frac{1}{C}\sum_{c=1}^{C} m_c. \tag{29}$$

In Equation (29), $C$ is the number of classes and $m_c$ is the metric for class $c$. Macro recall is simply the arithmetic mean of the class recalls, regardless of class size.

Micro averaging starts by pooling the counts:

$$P_{\text{micro}} = \frac{\sum_c TP_c}{\sum_c (TP_c + FP_c)}, \quad R_{\text{micro}} = \frac{\sum_c TP_c}{\sum_c (TP_c + FN_c)}, \quad F_{1,\text{micro}} = \frac{2P_{\text{micro}}R_{\text{micro}}}{P_{\text{micro}} + R_{\text{micro}}}. \tag{30}$$

Equation (30) defines $P_{\text{micro}}$, $R_{\text{micro}}$, and $F_{1,\text{micro}}$ as micro precision, recall, and F1. For single-label classification, all three equal accuracy because each error contributes one false positive and one false negative. Table 30 illustrates the difference using classes of 50, 30, and 20 observations. Class C has 50.0% recall, which lowers macro recall to 74.4% even though the micro value remains 80.0%.

Table 30. Three-class confusion matrix and macro/micro calculations for 100 single-label observations, 80 of which are classified correctly. Class-specific rows use one-versus-rest counts. Macro metrics weight classes equally; micro metrics pool the underlying decisions.

| **Actual  Predicted** | **A** | **B** | **C** | **Row total** |
|---|---|---|---|---|
| A | 45 | 3 | 2 | 50 |
| B | 5 | 25 | 0 | 30 |
| C | 4 | 6 | 10 | 20 |

| **Class/average** | **Precision, %** | **Recall, %** | **F1, %** |
|---|---|---|---|
| A | 83.3 | 90.0 | 86.5 |
| B | 73.5 | 83.3 | 78.1 |
| C | 83.3 | 50.0 | 62.5 |
| Macro | 80.1 | 74.4 | 75.7 |
| Micro | 80.0 | 80.0 | 80.0 |

The matrix shows how a respectable micro F1 can coexist with weak performance in a minority class. Macro F1 makes that imbalance easier to see, but class-specific results are still needed to locate the errors. A weighted macro average is also acceptable when the support weights are reported. In biomedical applications, the full multiclass confusion matrix is particularly useful because different misclassifications may have different consequences.

### 9.10. Pooled versus fold-averaged metrics

Let $C_k$ denote the confusion-count vector in fold $k$ and let $m(\cdot)$ be a metric function. Two common summaries are

$$\bar{m} = \frac{1}{K}\sum_{k=1}^{K} m(C_k), \qquad m_{\text{pool}} = m\left(\sum_{k=1}^{K} C_k\right). \tag{31}$$

In Equation (31), $K$ is the fold count, $\bar{m}$ weights folds equally, and $m_{\text{pool}}$ weights the observations that form the summed count vector. They coincide only for selected linear metrics under equal fold sizes. Table 31 uses folds of 10 and 90 observations, so unweighted average accuracy is 85.0% but pooled accuracy is 81.0%. Nonlinearity also makes averaged F1 73.3% and pooled F1 79.6%.

Table 31. Numerical difference between equal-fold averaging and pooled held-out predictions. Unequal fold sizes make the unweighted accuracy overrepresent the 10-case fold, while F1 additionally differs because it is nonlinear in the confusion counts.

| Fold/summary | TP | FN | FP | TN | Accuracy, % | F1, % |
|---|---|---|---|---|---|---|
| Fold 1, $n = 10$ | 1 | 0 | 1 | 8 | 90.0 | 66.7 |
| Fold 2, $n = 90$ | 36 | 9 | 9 | 36 | 80.0 | 80.0 |
| Equal-fold average | — | — | — | — | 85.0 | 73.3 |
| Pooled matrix | 37 | 9 | 10 | 44 | 81.0 | 79.6 |

Pooled metrics answer a random-held-out-observation question. Equal-fold or equal-group averages answer a random-fold or random-group question and may be appropriate when groups deserve equal weight. The report should state both the unit and the aggregation operator. Silent switching between them can produce differences larger than a proposed algorithmic improvement.

### 9.11. Uncertainty and inferential examples

For $R$ independent replicate metrics with mean $\bar{m}$ and sample SD $s_m$, the usual standard error and interval are

$$\mathrm{SE}(\bar{m}) = \frac{s_m}{\sqrt{R}}, \qquad \mathrm{CI}_{95\%} = \bar{m} \pm t_{0.975,R-1}\,\mathrm{SE}(\bar{m}). \tag{32}$$

In Equation (32), $R$ is the number of independent replicates and $t_{0.975,R-1}$ is a Student-$t$ quantile. CV folds violate independence because their training sets overlap, so Equation (32) is descriptive when applied directly to folds.

For $R$ paired resampled differences $d_r$ with mean $\bar{d}$, SD $s_d$, training size $n_{tr}$, and test size $n_{te}$, the Nadeau–Bengio corrected statistic is

$$t_{\mathrm{corr}} = \frac{\bar{d}}{\sqrt{\left(\frac{1}{R} + \frac{n_{te}}{n_{tr}}\right) s_d^2}}, \qquad \mathrm{df} = R - 1. \tag{33}$$

In Equation (33), $d_r$ compares two algorithms on the same split and the ratio $n_{te}/n_{tr}$ inflates variance for training-set overlap. The correction is approximate and does not create independent data [24, 25].

For a score whose larger values are better, a finite permutation test uses

$$p = \frac{1 + \sum_{b=1}^{B} \mathbf{1}\left(T_b \geq T_{\mathrm{obs}}\right)}{B + 1}. \tag{34}$$

In Equation (34), $T_{\mathrm{obs}}$ is the observed complete-pipeline score, $T_b$ is its score under permutation $b$, $B$ is the number of permutations, and $\mathbf{1}$ is an indicator. The full pipeline must be retrained for every permutation, and labels must be permuted at the independent-unit level [15]. Table 32 quantifies the difference between a naive and corrected comparison and gives a finite permutation example. With 50 paired 10-fold differences, $\bar{d} = 0.025$ and $s_d = 0.040$, the nominally decisive naive result becomes nonsignificant after correction.

Table 32. Worked uncertainty, corrected-comparison, and permutation calculations. The first row uses genuinely regenerated studies; the next two expose the effect of CV dependence; the final row uses the plus-one finite-permutation convention.

| Calculation | Inputs | SE or denominator | Statistic | Two-sided or permutation p |
|---|---|---|---|---|
| Independent-replicate CI for S8 stratified 10-fold accuracy | R = 100, mean = 0.753, SD = 0.060 | SE = 0.0060 | 95% CI [0.741, 0.765] | — |
| Naive paired t-test | R = 50, d_bar = 0.025, s_d = 0.040 | 0.00566 | t = 4.42 | 0.000055 |
| Nadeau–Bengio corrected t-test | Above; n_te / n_tr = 1 / 9 | 0.01448 | t_corr = 1.73 | 0.091 |
| Permutation test | B = 999; b = 4 permuted scores at least as large as T_obs | — | p = (b + 1) / (B + 1) | 0.005 |

The corrected test compares the average difference between algorithms, whereas the permutation test asks whether the complete pipeline performs better than a null distribution obtained by exchangeable labels. Neither test, on its own, establishes clinical usefulness, transportability, or freedom from leakage. If several comparisons or metrics are chosen after inspecting the results, those choices need their own safeguards. Effect sizes, held-out confusion matrices, and uncertainty should therefore accompany $p$-values.

### 9.12. Cross-scenario synthesis

Table 33 brings the eight scenarios together, but they should not be interpreted as estimates of a single bias mechanism. S1 concerns sensitivity to partition choice, whereas S8 examines estimator variation across independently regenerated studies. The other cases demonstrate information crossing the assessment boundary or a split performed at a smaller unit than the deployment target.

Table 33. Main results for the eight controlled scenarios. S1 illustrates partition instability; S8 compares mean accuracy and estimator variability rather than systematic bias.

| ID | Failure mechanism | Flawed or unstable result | Safer result | Main lesson |
|---|---|---|---|---|
| S1 | Tiny, imbalanced hold-out | 87.8 ± 5.3%; range 77.8–94.4% | Repeated stratified 5-fold: 88.3 ± 1.2% | Stratification and repetition reduce split sensitivity but do not add cases. |
| S2 | Feature selection before CV | 93.3% under random labels | Nested selection and tuning: 51.7% | Global screening can manufacture strong performance from noise. |
| S3 | Epoch-level subject leakage | 97.0% sample-level CV | LOSO: 71.0% | A stable row-level score may estimate recognition of known subjects. |
| S4 | Fellow-eye patient leakage | 95.0% eye-level CV | Patient-grouped CV: 81.5% | Both eyes from one patient must remain in the same partition. |

| ID | Failure mechanism | Flawed or unstable result | Safer result | Main lesson |
|---|---|---|---|---|
| S5 | Mixed-center site leakage | 93.0% mixed-center CV | Leave-one-center-out: 76.7%; centers 61.0–89.0% | Pooled CV can conceal transport failure at an unseen center. |
| S6 | Global normalization | 88.3% | Fold-local normalization: 82.5% | Label-free transformations can still use assessment information. |
| S7 | Repeated test-set use | 92.0% after 50 choices | Untouched lockbox: 86.0% | A test set becomes development data after it guides a decision. |
| S8 | LOOCV variance depends on learner stability | Stable learner: 75.5 ± 5.5% at 80 fits | 10-fold: 75.3 ± 6.0% at 10 fits; with an unstable learner LOOCV becomes noisier (8.8 vs 7.5) | More fits do not buy a better estimate; LOOCV variance is learner-dependent. |

The largest changes in these constructed examples appear when feature selection is performed globally, when records from the same subject or patient are split across training and assessment, and when centers are mixed. Fixing these cases requires either a different split or moving data-dependent operations inside the training boundary. S8 addresses a different issue: LOOCV uses eight times as many fits, has almost the same mean as stratified 10-fold CV for the stable LDA learner, and is more variable for the unstable tree.

Across S1-S8, misleading estimates arise for different reasons: unstable class counts, pre-split selection, repeated identities, paired organs, center effects, global preprocessing, repeated test-set feedback, and learner instability. The practical response is consistent: split at the independent unit, fit every data-dependent step using training data only, and do not let assessment results guide later development.

## 10. Method selection by data size

Sample-size planning should begin with the units that will actually be new at deployment. For a new-subject claim, 600 epochs collected from 30 people provide evidence that is much closer to 30 independent observations than to 600. The appropriate design also depends on class support, group sizes, dimensionality, learner stability, and available computation. The ranges below are intended as working guides rather than universal thresholds.

### 10.1. Very small data: $N < 100$

Below 100 independent units, one split is usually too sensitive to a few cases. Repeated stratified five-fold CV is a practical default for independent rows; when deployment targets a new subject, patient, or other group, repeated grouped five-fold CV or LOSO is required to preserve the relevant independence unit. LOOCV trains on nearly all observations, but S8 shows that its advantage over stratified 10-fold CV is small for a stable learner and reverses for an unstable one, while its cost is fixed at N fits. Taken together with the partition-stability illustration in S1, these considerations support repeated stratified five-fold CV as a practical default below 100 independent units, with LOOCV reserved for stable or closed-form learners and for sensitivity analysis. Sample-size planning should be tied to the target model and outcome frequency rather than a universal row-count threshold [26].

**10.2. Small to moderate data: $100 \leq N \leq 1,000$**

Repeated stratified five- or 10-fold CV is usually practical in this range and gives a useful view of partition-to-partition variation. Nested CV is warranted when the data are used to choose features, model families, thresholds, or hyperparameters; keeping the candidate set modest limits both computation and selection variance. A 20% hold-out with $N = 120$ leaves only 24 observations for assessment and is therefore fragile, whereas the same fraction with $N = 1{,}000$ leaves 200. Subject- or patient-level grouping is still required whenever rows are dependent, regardless of the nominal row count.

**10.3. Moderate to large data: $1,000 < N \leq 10,000$**

When each assessment fold still contains hundreds of observations from the important classes, a single stratified five- or 10-fold run may be sufficient for a prespecified classical pipeline. Repetition becomes more informative when the learner is unstable or when the groups are heterogeneous. A hold-out design can also be appropriate if it contains enough independent cases from each class and is kept untouched throughout development. A broad model search may justify nested assessment, whereas a large independent lockbox can provide a direct final check once the system has already been selected.

**10.4. Large data and deep learning: $N > 10,000$**

For expensive models, a group-correct train/validation/test split may offer the best compromise between rigor and cost. The validation set is still development data because it informs early stopping, architecture, and thresholds; only the locked test set evaluates the selected system. Multiple training seeds can describe optimization variability without repeated test-set consultation. Even millions of images say little about new-patient or new-center performance if they come from only 40 patients or three centers.

**10.5. High-dimensional data with few independent units**

Feature selection is especially vulnerable to chance associations when the candidate feature set is much larger than the number of independent units. Any supervised screening must therefore be refitted within each inner training fold. A feature list chosen once from all observations leaks information into the outer assessment, even if the classifier itself is simple.

A safer strategy is to combine nested cross-validation with regularization, a limited candidate space, and a permutation-based negative control. Feature stability across outer folds is also informative: strong predictive performance can coexist with a biomarker set that changes substantially from fold to fold. Random search, sparse models, three-fold inner validation, and prespecified feature families can reduce computation without relaxing the information boundary.

The D/N ratio alone is not a complete description of selection risk. What also matters is how many data-driven comparisons are made relative to N_eff. Searching 10,000 candidate features in 30 patients is therefore a high-risk selection problem even if those patients generate hundreds of images.

Table 34 relates the sample-size ranges to independent rows, repeated observations within subjects, and multicenter data. Its suggestions assume sufficient class support and processing fitted within each fold. 'Nested' places selection in the inner loop; 'grouped' keeps records attached to their subject, patient, or center. Neither technique can compensate for too few independent units to support the claim.

Table 34. Validation choices by independent-unit count and dependence structure. N counts the units expected to be new at deployment. Additional images, epochs, windows, or measurements from the same unit do not increase N.

| Independent-unit size | Independent observations | Subject/patient-dependent rows | Multicenter or multi-device data |
|---|---|---|---|
| $N < 100$ | Repeated stratified 5-fold; LOOCV as sensitivity analysis | LOSO plus subject distribution; repeated grouped 5-fold if class/group counts permit | Leave-one-center-out with every center shown; conclusions exploratory when centers are very few |
| $100 \le N \le 1{,}000$ | Repeated stratified 5/10-fold; nested CV for selection | StratifiedGroupKFold or LOSO; nested group CV for tuning | Report mixed-center and leave-one-center-out targets separately |
| $1{,}000 < N \le 10{,}000$ | One stratified 5/10-fold often sufficient; hold-out if absolute counts are large | GroupKFold; patient-level hold-out can be efficient | Grouped center-aware folds; reserve a center if transport is central |
| $N > 10{,}000$ | Locked train/validation/test; repeat optimization seeds if needed | Subject/patient-level train/validation/test; never split descendants | Center-stratified development plus untouched external centers when available |

With fewer than 100 independent units, Table 34 generally favors repeated five-fold CV over routine LOOCV; using almost every observation for training adds little if the resulting estimate is unstable or hard to interpret. For roughly 100-1,000 units, nested assessment is often feasible for classical models. Above about 10,000 genuinely independent units, one untouched split may already give narrow uncertainty without repeated full training. Rare classes, distribution shift, and high-consequence deployment can still justify stronger safeguards at any size. Count the target units before choosing folds, repetitions, or nesting.

A useful sequence is therefore to count the independent target units first and decide on folds, repetitions, and nesting only afterward. Very small datasets often benefit from repeated stratified or grouped CV, whereas genuinely large independent samples can support a locked three-way split. A large number of stored rows cannot compensate for having only a small number of subjects or centers.

## 11. Decision tree and reporting checklist

### 11.1. Selecting a design

Validation design is easier to choose when it is treated as a sequence of deployment questions rather than a list of software options. Figure 12 begins by asking whether the rows are dependent, then asks whether the target is a new group or center, whether substantial model selection is being performed, and whether a sufficiently large independent test set is available. Dependence moves the split boundary to the relevant group level. Model selection adds an inner loop, but it does not change that outer boundary.

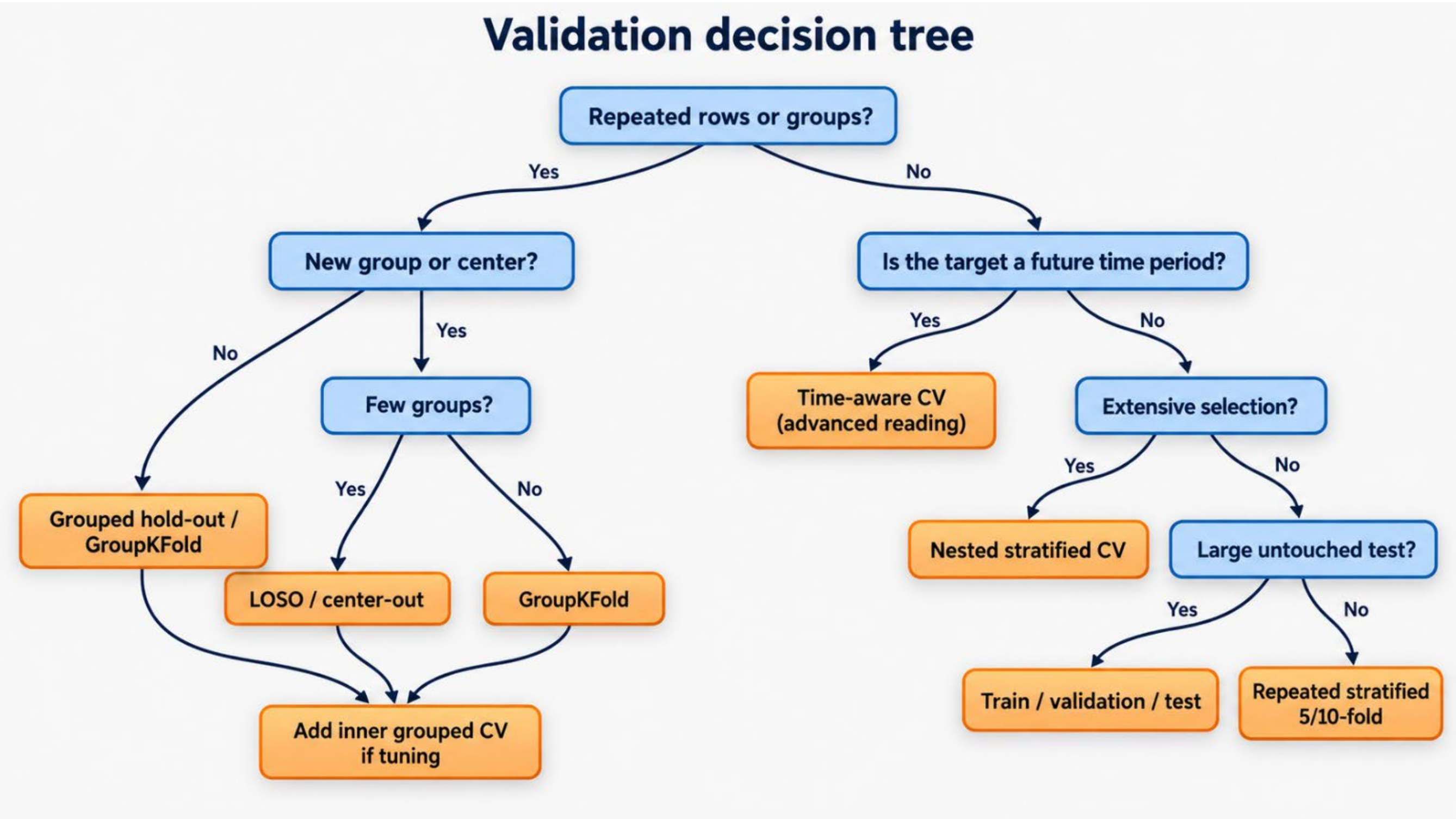

Figure 12. Decision tree for choosing a validation design from the intended independent unit, dependence structure, temporal ordering, amount of model selection, and available assessment size. Group integrity and time order are resolved before stratification or nesting, and inner-loop selection is kept separate from outer performance estimation.

The decision tree does not always lead to a single reportable number. A multicenter study, for example, may report mixed-center GroupKFold for interpolation within the observed center mixture and leave-one-center-out for transport to an unseen center, provided that those targets are named separately. Likewise, a deep model may use one grouped train/validation/test split while a cheaper frozen classifier is assessed with repeated CV. Whether the resulting estimates are comparable depends on the scientific target, not on whether the studies use the same validation style.

### 11.2. Reporting and reviewer checklist

Table 35 turns the design principles into a checklist that an author or reviewer should be able to complete from the manuscript, supplementary material, or released code. The most consequential omissions are those that prevent the information boundary from being reconstructed. Extra decimal places add little when the independent unit itself has not been defined.

Table 35. Checklist for reporting and reviewing machine-learning validation. It is intended for both manuscript preparation and peer review. A red flag indicates that clarification or reanalysis is needed; by itself, it does not prove that the analysis is invalid.

| No. | Reporting item | Minimum information | Reviewer red flag |
|---|---|---|---|
| 1 | Target claim | New row, subject, patient, eye, session, device, center, or time point | Target population is unstated |
| 2 | Independent unit | Stored row count, independent-unit count, group hierarchy, and group sizes | Rows are called subjects |
| 3 | Split design | Hold-out ratio or fold/repetition counts and seed | "10-fold CV" without independent unit |
| 4 | Class composition | Counts per class in every final partition; rare-class handling | Ratios only; zero-class folds possible |
| 5 | Group integrity | Identifier used and assertion of zero overlap | Epochs, eyes, or scans split independently |
| 6 | Preprocessing scope | Fit/apply boundary for imputation, scaling, PCA, harmonization | Transformation fitted before splitting |
| 7 | Feature selection | Method, candidate feature counts, and fold where ranking occurs | One global feature list precedes CV |
| 8 | Hyperparameter selection | Candidate space, inner folds, metric, tie rule | Best CV score also reported as final performance |
| 9 | Threshold/calibration | Data used to choose threshold and calibration model | Test-set threshold optimization |
| 10 | Final assessment | Untouched outer/test predictions and number of openings | Test results guided revisions |
| 11 | Metrics | Positive class, confusion counts, macro/micro and pooling rule | Accuracy alone for imbalance |
| 12 | Uncertainty | Unit resampled, SD/SE distinction, CI method | Fold SD treated as independent-subject evidence |
| 13 | Statistical comparison | Paired splits, corrected variance or permutation unit | Naive fold-wise -test |
| 14 | Reproducibility | Seeds, software versions, fold IDs, code, and failure handling | Folds cannot be reconstructed |

This checklist is best completed before the final test set is opened. Saving fold identifiers and held-out predictions is especially useful because it allows the reported metrics to be reconstructed independently. Biomedical papers should place patient and center counts alongside image or epoch counts so that the true unit of evidence is visible. If the reported split evaluates an easier problem than the claimed deployment setting, a target-aligned reanalysis is warranted.

### 11.3. Template methods paragraph

For the repeated-measure EEG example, the methods could be described more naturally as follows:

Performance was estimated using leave-one-subject-out cross-validation across 30 participants. At each outer iteration, all 20 epochs from one participant were withheld together as the assessment set. Within the remaining participants, five-fold GroupKFold was used to choose the hyperparameters and the number of selected features. Standardization and feature selection were refitted in every inner training fold and then applied, without refitting, to the corresponding validation fold. After a configuration had been selected, the full pipeline was refitted on the complete outer training set and used once on the held-out participant. Final metrics were calculated from the pooled outer predictions, and per-participant accuracy was reported separately. Confidence intervals came from stratified bootstrap resampling at the participant level. A 999-permutation negative control repeated the complete pipeline after permuting labels by subject. Random seeds, fold identifiers, software versions, and all held-out predictions were archived.

In summary, Decide the deployment unit before selecting a resampling function. Use the checklist to make information boundaries, selection roles, group integrity, aggregation, and uncertainty reconstructable. A validation result is credible when its target and pipeline can be audited from counts and indices.

## 12. Limitations and discussion

### 12.1. Limits of internal validation

No internal validation method can repair a nonrepresentative cohort, systematic label error, inappropriate outcome definition, or deployment shift absent from the data. Nested CV controls reuse during selection but does not guarantee calibration, fairness, causal validity, or clinical benefit. Group-aware CV protects declared identifiers but may miss latent families, duplicate sources, or device correlations. External evidence and prospective monitoring remain necessary for consequential deployment. Small-sample image and biomedical studies should therefore combine resampling with explicit representativeness and dataset-quality arguments [27].

The numerical cutoffs in Section 10 and the S1–S8 outcomes are teaching devices rather than universal performance laws. Uncertainty estimation from CV folds remains difficult because the training sets overlap, and corrected tests are approximations rather than exact fixes. With only a handful of centers, the individual center results can be more informative than a narrow population interval obtained from strong assumptions. In addition, computational constraints may reasonably favor a simpler locked design when a sufficiently large independent assessment sample is already available.

### 12.2. Methods outside the main scope

Several related strategies are left to advanced reading because they rely on assumptions and deployment targets outside this tutorial's main scope. These include bootstrap, .632, and .632+ estimators [28], out-of-bag error, forward-chaining and purged time-series CV, and external validation across datasets [18, 20]. Time-dependent tasks must preserve temporal order, and external datasets should be evaluated without retrospective adaptation. The mechanics differ, but the rule is the same: assessment information must not feed back into learning.

Here, probability calibration is discussed only as a learned pipeline step that must remain inside the training boundary (Tables 4 and 9). A full treatment would require worked examples of calibration plots, slope and intercept, Brier-score analysis and decomposition, and recalibration on external data. Bootstrap, temporal, and cross-dataset validation are likewise beyond the detailed examples used here.

### 12.3. Open questions

Biomedical data often have several nested levels at once—windows within recordings, recordings within sessions, sessions within subjects, and subjects within centers. No single grouping label captures all of these relationships. The practical challenge is to choose an outer unit that matches deployment while still respecting the lower-level structure.

A related open question is how quickly selection bias grows as the number of candidate pipelines increases relative to N_eff. More quantitative guidance is needed to separate settings where a large untouched test set is sufficient from those where nested assessment remains necessary. Related work has shown that hyperparameter-selection criteria can themselves be overfit [29], while reviews of machine learning in genomics document analogous evaluation pitfalls in high-dimensional biological data [30].

Foundation models create an additional provenance problem before downstream validation even starts. Pretraining data may already contain an evaluation sample, a near-duplicate, or records from the same patient or institution. A later train/test split cannot remove that earlier overlap. Whenever possible, document pretraining sources, check benchmark overlap, and disclose unresolved contamination when interpreting downstream performance.

Resampling controls a specified form of data reuse; it does not create representativeness or new independent patients. Numerical recommendations should therefore be treated as conditional, heterogeneity should be shown when groups are few, and transport claims should rely on external validation. Bootstrap, temporal, and cross-dataset designs warrant their own dedicated treatment.

## 13. Conclusion

Validation estimates the performance of a complete learning procedure under a declared future-use condition. Hold-out, k-fold, LOOCV, LOSO, GroupKFold, leave-one-center-out, and nested cross-validation differ in training fraction, variance, computational cost, and the unit withheld from learning. The decisive design choice is the unit that will be new at deployment.

The controlled scenarios showed that plausible scores can change sharply when the information boundary is corrected. Random-label accuracy fell from 93.3% to 51.7% after feature selection was nested. EEG epoch-level accuracy fell from 97.0% to 71.0% under LOSO. Eye-level OCT accuracy fell from 95.0% to 81.5% after both eyes of each patient were kept together. Mixed-center accuracy fell from 93.0% to 76.7% when centers were held out. These reductions are not penalties imposed by stricter software. They are corrections to the question being estimated.

The examples repeatedly point to three practices. Split at the level of the unit that must be new at deployment; fit learned transformations and data-driven choices only within the active training data; and keep final assessment results out of model, feature, threshold, and reporting decisions. Save confusion matrices, per-unit outcomes, fold identifiers, aggregation rules, and uncertainty procedures so the reported estimate can be reconstructed.

A sophisticated learner is still weak evidence of deployment performance if information crosses the assessment boundary. Validation should therefore be planned with the experiment, before model fitting. The split and the allowed information flow define what the final performance estimate actually means.

## Appendix A. MATLAB templates for implementation

### A.1. Repeated stratified CV, LOSO, and nested group CV in MATLAB

Listing A1 contains the complete MATLAB function validation_tutorial_demo.m. It generates grouped synthetic data, fits preprocessing only within active folds, and runs repeated CV, LOSO, and nested group CV. In the nested routine, the supervised univariate selector is reranked within every inner training fold and again on the complete outer training partition after a configuration is chosen; outer assessment records never enter that ranking. Listing B1 follows the same separation in scikit-learn by placing SelectKBest inside a Pipeline. Undefined metric denominators return NaN, and summary means use omitnan rather than substituting zero.

```
function validation_tutorial_demo
% VALIDATION_TUTORIAL_DEMO  Leakage-safe validation templates (Listing A1).
% Requires Statistics and Machine Learning Toolbox.
%
% Changes relative to the earlier draft of this listing:
%   (1) The nested loop now contains a SUPERVISED FEATURE SELECTOR that is
%       refitted inside every inner training fold and re-selected on the
%       full outer training set after the configuration is chosen. Without
%       this, the listing did not demonstrate the containment rule that
%       scenario S2 is built to teach, and it did not match the
%       scikit-learn counterpart in Listing B1.
%   (2) binaryMetrics returns NaN for undefined quantities instead of
%       silently substituting zero, matching the convention declared in
%       Table 7.
%   (3) Both the row-level and the grouped routines now report pooled
%       held-out metrics as the primary figure, so the MATLAB and Python
%       listings use the same aggregation rule (see Section 9.10).

rng(20260803,"twister");
G=30; nPerGroup=20; D=12;
group=repelem((1:G)',nPerGroup);
yGroup=[zeros(G/2,1);ones(G/2,1)]; y=repelem(yGroup,nPerGroup);
subjectOffset=0.70*randn(G,D);
X=randn(G*nPerGroup,D)+subjectOffset(group,:);
X(:,1:3)=X(:,1:3)+0.55*y;

rowOut   = repeatedStratifiedCV(X,y,5,2,20260803);
losoOut  = losoSVM(X,y,group);
nestedOut= nestedGroupCVSVM(X,y,group,5,3,[0.1 1],[0.5 1],[4 8],20260803);

fprintf("Pooled held-out accuracy\n");
fprintf("  row-level repeated 5-fold : %.3f  (mean of repetition-pooled)\n", rowOut.meanAccuracy);
```

```
fprintf("  LOSO                       : %.3f\n", losoOut.metrics.accuracy);
fprintf("  nested group CV            : %.3f\n", nestedOut.metrics.accuracy);
fprintf("  equal-subject LOSO mean    : %.3f\n", mean(losoOut.subjectAccuracy));
end

% ----------------------------------------------------------------------
function out=repeatedStratifiedCV(X,y,K,R,seed)
rng(seed,"twister"); n=size(X,1); allPred=nan(n,R);
repAccuracy=nan(R,1); repMetrics=cell(R,1);
for r=1:R
    cv=cvpartition(y,'KFold',K); pred=nan(n,1);
    for k=1:K
        tr=training(cv,k); te=test(cv,k);
        [Xtr,Xte]=trainZscore(X(tr,:),X(te,:));
        mdl=fitcsvm(Xtr,y(tr),'KernelFunction','linear', ...
            'BoxConstraint',1,'Standardize',false,'ClassNames',[0;1]);
        pred(te)=predict(mdl,Xte);
    end
    assert(all(~isnan(pred)),'every row must receive one held-out prediction');
    allPred(:,r)=pred;
    repMetrics{r}=binaryMetrics(y,pred);          % pooled over the repetition
    repAccuracy(r)=repMetrics{r}.accuracy;
end
out=struct('predictionByRepetition',allPred,'repetitionMetrics',{repMetrics}, ...
    'repetitionAccuracy',repAccuracy,'meanAccuracy',mean(repAccuracy), ...
    'sdAccuracy',std(repAccuracy));
end

% ----------------------------------------------------------------------
function out=losoSVM(X,y,group)
u=unique(group,'stable'); pred=nan(size(y)); subjectAccuracy=nan(numel(u),1);
for s=1:numel(u)
    te=group==u(s); tr=~te;
    assert(isempty(intersect(unique(group(tr)),unique(group(te)))), ...
        'group crossed the LOSO boundary');
    [Xtr,Xte]=trainZscore(X(tr,:),X(te,:));
    mdl=fitcsvm(Xtr,y(tr),'KernelFunction','rbf','BoxConstraint',1, ...
        'KernelScale',1,'Standardize',false,'ClassNames',[0;1]);
    pred(te)=predict(mdl,Xte); subjectAccuracy(s)=mean(pred(te)==y(te));
end
assert(all(~isnan(pred)));
out=struct('prediction',pred,'metrics',binaryMetrics(y,pred), ...
    'subjectID',u,'subjectAccuracy',subjectAccuracy);
end

% ----------------------------------------------------------------------
function out=nestedGroupCVSVM(X,y,group,Kouter,Kinner,boxGrid,scaleGrid,kGrid,seed)
% Candidate space Lambda = boxGrid x scaleGrid x kGrid. Scaling AND the
% supervised feature ranking are refitted inside every inner training fold.
u=unique(group,'stable'); G=numel(u); pred=nan(size(y));
if Kouter>=G
    [~,~,outerFold]=unique(group,'stable'); Kouter=G;    % outer LOSO
else
    outerFold=stratifiedGroupFolds(y,group,Kouter,seed);
end
selected=nan(Kouter,3);
for o=1:Kouter
    teOuter=outerFold==o; trOuter=~teOuter;
    assert(isempty(intersect(unique(group(trOuter)),unique(group(teOuter)))), ...
        'group crossed the outer boundary');
    Xouter=X(trOuter,:); youter=y(trOuter); innerGroups=group(trOuter);
    innerFold=stratifiedGroupFolds(youter,innerGroups,Kinner,seed+o);

    bestLoss=inf; best=[NaN NaN NaN];
    for b=1:numel(boxGrid)
      for q=1:numel(scaleGrid)
        for kk=1:numel(kGrid)
            nSel=min(kGrid(kk),size(X,2));
            foldLoss=nan(Kinner,1);
            for j=1:Kinner
                trInner=innerFold~=j; vaInner=innerFold==j;
                assert(isempty(intersect(unique(innerGroups(trInner)), ...
                    unique(innerGroups(vaInner)))),'group crossed the inner boundary');
                [Xtr,Xva]=trainZscore(Xouter(trInner,:),Xouter(vaInner,:));
                idx=selectFeatures(Xtr,youter(trInner),nSel);   % fitted on inner training only
                mdl=fitcsvm(Xtr(:,idx),youter(trInner),'KernelFunction','rbf', ...
                    'BoxConstraint',boxGrid(b),'KernelScale',scaleGrid(q), ...
                    'Standardize',false,'ClassNames',[0;1]);
                m=binaryMetrics(youter(vaInner),predict(mdl,Xva(:,idx)));
                foldLoss(j)=1-m.balancedAccuracy;
            end
            candidateLoss=mean(foldLoss,'omitnan');
            if candidateLoss<bestLoss
                bestLoss=candidateLoss; best=[boxGrid(b) scaleGrid(q) nSel];
            end
```

```
        end
      end
    end

    % refit the selected configuration on the complete outer training set
    [Xtr,Xte]=trainZscore(X(trOuter,:),X(teOuter,:));
    idx=selectFeatures(Xtr,y(trOuter),best(3));    % ranking refitted, never on outer test
    finalModel=fitcsvm(Xtr(:,idx),y(trOuter),'KernelFunction','rbf', ...
        'BoxConstraint',best(1),'KernelScale',best(2), ...
        'Standardize',false,'ClassNames',[0;1]);
    pred(teOuter)=predict(finalModel,Xte(:,idx));  % predicted once, no reranking
    selected(o,:)=best;
end
assert(all(~isnan(pred)));
out=struct('prediction',pred,'metrics',binaryMetrics(y,pred), ...
    'selected',array2table(selected,'VariableNames', ...
    {'BoxConstraint','KernelScale','NumFeatures'}),'outerFold',outerFold);
end

% ----------------------------------------------------------------------
function idx=selectFeatures(Xtr,ytr,nSel)
% Univariate two-sample t statistic computed from training rows only.
mu1=mean(Xtr(ytr==1,:),1); mu0=mean(Xtr(ytr==0,:),1);
v1=var(Xtr(ytr==1,:),0,1);  v0=var(Xtr(ytr==0,:),0,1);
n1=sum(ytr==1); n0=sum(ytr==0);
score=abs(mu1-mu0)./sqrt(max(v1/max(n1,1)+v0/max(n0,1),eps));
[~,order]=sort(score,'descend');
idx=sort(order(1:min(nSel,numel(order))));
end

% ----------------------------------------------------------------------
function fold=stratifiedGroupFolds(y,group,K,seed)
% Greedy binary-class allocation; groups remain indivisible.
rng(seed,"twister"); [~,~,gi]=unique(group,'stable'); G=max(gi);
assert(G>=K,'fewer groups than folds');
counts=[accumarray(gi,y==0,[G 1]) accumarray(gi,y==1,[G 1])];
priority=max(counts,[],2)+1e-6*rand(G,1); [~,order]=sort(priority,'descend');
target=sum(counts,1)/K; foldCounts=zeros(K,2); groupFold=zeros(G,1);
for z=1:G
    g=order(z);
    if z<=K
        f=z;
    else
        score=inf(K,1);
        for k=1:K
            candidate=foldCounts; candidate(k,:)=candidate(k,:)+counts(g,:);
            score(k)=sum(((candidate-target).^2)./max(target,1),'all');
        end
        f=find(score==min(score),1,'first');
    end
    groupFold(g)=f; foldCounts(f,:)=foldCounts(f,:)+counts(g,:);
end
fold=groupFold(gi);
for k=1:K
    assert(isempty(intersect(unique(group(fold~=k)),unique(group(fold==k)))));
end
end

% ----------------------------------------------------------------------
function [XtrZ,XteZ]=trainZscore(Xtr,Xte)
mu=mean(Xtr,1,'omitnan'); sigma=std(Xtr,0,1,'omitnan');
sigma(~isfinite(sigma)|sigma==0)=1;
XtrZ=(Xtr-mu)./sigma; XteZ=(Xte-mu)./sigma;
end

% ----------------------------------------------------------------------
function m=binaryMetrics(y,p)
% Undefined quantities are returned as NaN. A zero denominator means the
% metric has no value for this partition; replacing it with zero would
% silently bias any average that includes it (Table 7).
y=double(y(:)); p=double(p(:));
tp=sum(y==1&p==1); fn=sum(y==1&p==0); fp=sum(y==0&p==1); tn=sum(y==0&p==0);
und=@(a,b) undefinedSafe(a,b);
m=struct('TP',tp,'FN',fn,'FP',fp,'TN',tn, ...
    'accuracy',und(tp+tn,numel(y)), ...
    'sensitivity',und(tp,tp+fn), ...
    'specificity',und(tn,tn+fp), ...
    'precision',und(tp,tp+fp), ...
    'F1',und(2*tp,2*tp+fp+fn));
m.balancedAccuracy=0.5*(m.sensitivity+m.specificity);
m.MCC=und(tp*tn-fp*fn, sqrt((tp+fp)*(tp+fn)*(tn+fp)*(tn+fn)));
end

function v=undefinedSafe(a,b)
if b>0, v=a/b; else, v=NaN; end
```

```
end
```

Final metrics use only the outer prediction vector. Inner-loop scores select BoxConstraint and KernelScale but never contribute to the reported confusion matrix. If group labels are strongly imbalanced, inspect the greedy allocation and replace it with a more suitable grouped splitter when needed. Save fold identifiers with the predictions so the assessment can be reconstructed later.

### A.2. Reproducing figures 1–12

Listing A2 uses the fixed arrays reported in the article. The make_validation_figures.m script exports vector PDFs and 300-dpi PNGs. MATLAB releases may wrap long labels differently, but the nodes, edges, matrices, and numerical values are unchanged. For Figure 2, the code now builds row-wise folds without subject identifiers and explicitly checks that the k-fold and LOSO panels are different.

```
function make_validation_figures(outputDir)
% MAKE_VALIDATION_FIGURES  Regenerates Figures 1-12 (Listing A2).
%
% Notes specific to this listing:
%   (1) Figure 2 now separates row-wise k-fold from group-aware LOSO. In the
%       older code, kfold(k,(3*k-2):(3*k))=2 and loso(k,groups==k)=2 could
%       produce the same pattern when fold size matched group size. Here the
%       k-fold panel deliberately mixes records across subject boundaries,
%       while LOSO keeps each subject intact. The hold-out example also crosses
%       one subject boundary.
%   (2) exportgraphics writes a vector PDF for arXiv and also saves a 300-dpi
%       PNG for quick visual inspection.
%   (3) Figure 12 includes the temporal branch discussed in Section 3.3.
%   (4) Figure 9 is the threshold-free example from Section 5.6; the former
%       Figures 9-11 therefore appear as Figures 10-12.
%   (5) The arrays below are the same values printed in the article so the
%       generated figures remain synchronized with the reported results.

if nargin==0, outputDir="validation_figures"; end
if ~isfolder(outputDir), mkdir(outputDir); end
close all; rng(20260801,"twister");
oi=[0 114 178;230 159 0;0 158 115;204 121 167;86 180 233;213 94 0]/255;

% ------------------------------------------------------------------ Fig 1
s=["Validation" "Validation" "Validation" "Validation" ...
   "Basic splits" "Cross-validation" "Group-aware" "Nested"];
t=["Basic splits" "Cross-validation" "Group-aware" "Nested" ...
   "Hold-out / three-way / repeated" "k-fold / LOOCV / leave-p-out" ...
   "Subject / eye / session / center" "Nested CV / nested group CV"];
figure('Color','w'); gp=plot(digraph(s,t),'Layout','layered','Direction','down', ...
    'NodeColor',oi(1,:),'MarkerSize',8,'LineWidth',1.2);
highlight(gp,["Hold-out / three-way / repeated" "k-fold / LOOCV / leave-p-out"], ...
    'NodeColor',oi(2,:));
highlight(gp,["Subject / eye / session / center" "Nested CV / nested group CV"], ...
    'NodeColor',oi(3,:)); title('Validation taxonomy');
saveFigure(1,outputDir);

% ------------------------------------------------------------------ Fig 2
% 4 subjects x 3 records; 1 = training, 2 = assessment.
groups=repelem(1:4,3); n=numel(groups); nGroup=4;

holdout=ones(1,n); holdout(8:12)=2;          % crosses the subject-3 boundary

kfold=ones(nGroup,n);                        % interleaved, group-naive
for k=1:nGroup, kfold(k,k:nGroup:n)=2; end

loocv=ones(n,n); for k=1:n, loocv(k,k)=2; end

loso=ones(nGroup,n);                         % the only group-clean design
for k=1:nGroup, loso(k,groups==k)=2; end

assert(~isequal(kfold,loso), ...
    'Figure 2 must contrast group-naive k-fold with group-aware LOSO.');

figure('Color','w'); tl=tiledlayout(2,2,'TileSpacing','compact');
names={"Hold-out","4-fold (group-naive)","LOOCV","LOSO (group-aware)"};
mats={holdout,kfold,loocv,loso};
for j=1:4
    nexttile; imagesc(mats{j}); hold on; xline([3.5 6.5 9.5],'k','LineWidth',1.4);
    title(names{j}); xlabel('Records ordered by subject'); ylabel('Split iteration');
```

```
    set(gca,'YTick',1:size(mats{j},1));
end
colormap(tl,[oi(1,:);oi(2,:)]); saveFigure(2,outputDir);

% ---------------------------------------------------------------- Fig 3
x=1:5; bias=[1.15 .78 .48 .26 .12]; variance=[.10 .18 .34 .62 1.02];
total=[1.35 1.06 .92 .98 1.24]; figure('Color','w');
plot(x,bias,'-o',x,variance,'--s',x,total,':^','LineWidth',1.8); grid on;
xlabel('Model complexity'); ylabel('Relative expected error');
legend('Bias^2','Variance','Total','Location','best'); saveFigure(3,outputDir);

% ---------------------------------------------------------------- Fig 4
figure('Color','w'); tiledlayout(1,2,'TileSpacing','compact');
nexttile; plot(digraph(["All data" "Global scale" "Global selection" "Split"], ...
    ["Global scale" "Global selection" "Split" "Score"]), ...
    'Layout','layered','Direction','down','NodeColor',oi(2,:)); title('Incorrect');
nexttile; plot(digraph(["All data" "Split" "Train-only scale" "Train-only selection"], ...
    ["Split" "Train-only scale" "Train-only selection" "Score held-out once"]), ...
    'Layout','layered','Direction','down','NodeColor',oi(1,:)); title('Correct');
saveFigure(4,outputDir);

% ---------------------------------------------------------------- Fig 5
s=["Outer data" "Outer data" "Outer train" "Inner folds" "Select lambda" "Outer refit"];
t=["Outer train" "Outer test" "Inner folds" "Select lambda" "Outer refit" "Predict outer test"];
figure('Color','w'); gp=plot(digraph(s,t),'Layout','layered','Direction','down', ...
    'NodeColor',oi(1,:),'LineWidth',1.2);
highlight(gp,["Outer test" "Predict outer test"],'NodeColor',oi(2,:));
title('Nested cross-validation'); saveFigure(5,outputDir);

% ---------------------------------------------------------------- Fig 6
% Both arrays are the locked S1 repetition accuracies of Table 15.
a=repelem([77.8 83.3 88.9 94.4],[2 5 8 5]);
b=repelem([85.8 86.7 87.5 88.3 89.2 90.0 90.8],[1 2 4 6 4 2 1]);
assert(numel(a)==20 && numel(b)==20);
figure('Color','w'); tiledlayout(1,2,'TileSpacing','compact');
nexttile; histogram(a,8,'FaceColor',oi(1,:)); xline(mean(a),'--k');
title(sprintf('Repeated hold-out (SD %.2f)',std(a))); xlabel('Accuracy (%)');
nexttile; histogram(b,7,'FaceColor',oi(2,:)); xline(mean(b),'--k');
title(sprintf('Repeated stratified 5-fold (SD %.2f)',std(b))); xlabel('Accuracy (%)');
saveFigure(6,outputDir);

% ---------------------------------------------------------------- Fig 7
k=[2 5 10 20 50 100]; rb=[.42 .24 .15 .10 .07 .04]; rv=[.25 .20 .22 .30 .42 .60];
figure('Color','w'); tiledlayout(1,3,'TileSpacing','compact');
nexttile; plot(k,rb,'-o','Color',oi(1,:)); title('Relative bias'); grid on;
nexttile; plot(k,rv,'--s','Color',oi(2,:)); title('Relative variance'); grid on;
nexttile; plot(k,k,':^','Color',oi(3,:)); title('Model fits'); grid on;
for j=1:3, nexttile(j); xlabel('k'); end
saveFigure(7,outputDir);

% ---------------------------------------------------------------- Fig 8
figure('Color','w'); tl=tiledlayout(1,3,'TileSpacing','compact');
plotCM([1 1;1 1],{'TP','FN';'FP','TN'},'Template');
plotCM([28 2;2 28],[],'Global selection');
plotCM([15 15;14 16],[],'Nested selection');
colormap(tl,flipud(gray(256))); saveFigure(8,outputDir);

% ---------------------------------------------------------------- Fig 9
% Threshold-free comparison for the two prevalence settings in Section 5.6.
% The locked scores use ten bins (bin 10 is highest). The two cohorts have
% identical class-conditional score distributions; only the number of
% negatives changes. Their ROC curves therefore coincide.
posCount =[30 22 16 11 8 5 4 2 1 1];        % 100 positives in both cohorts
negBal   =[ 1  1  2  4 5 8 11 16 22 30];    % 100 negatives, prevalence 50%
negRare  =negBal*19;                        % 1900 negatives, prevalence 5%
assert(sum(posCount)==100 && sum(negBal)==100 && sum(negRare)==1900);
[fprBal,tprBal,recBal,preBal,aucBal,apBal]=rocpr(posCount,negBal);
[fprRar,tprRar,recRar,preRar,aucRar,apRar]=rocpr(posCount,negRare);
assert(abs(aucBal-aucRar)<1e-12,'ROC must be prevalence-invariant here');
figure('Color','w'); tiledlayout(1,2,'TileSpacing','compact');
nexttile; plot(fprBal,tprBal,'-o','Color',oi(1,:),'LineWidth',1.8); hold on;
plot(fprRar,tprRar,'--s','Color',oi(2,:),'LineWidth',1.4);
plot([0 1],[0 1],':k'); grid on; axis square;
xlabel('1 - specificity'); ylabel('Sensitivity');
title(sprintf('ROC: identical, AUC %.3f',aucBal));
legend(sprintf('prevalence 50%%'),sprintf('prevalence 5%%'),'chance','Location','southeast');
nexttile; plot(recBal,preBal,'-o','Color',oi(1,:),'LineWidth',1.8); hold on;
plot(recRar,preRar,'--s','Color',oi(2,:),'LineWidth',1.4);
yline(0.50,':','Color',oi(1,:)); yline(0.05,':','Color',oi(2,:));
grid on; axis square; ylim([0 1]);
xlabel('Recall'); ylabel('Precision');
title(sprintf('PR: %.3f versus %.3f',apBal,apRar));
legend('prevalence 50%','prevalence 5%','baseline 0.50','baseline 0.05','Location','southwest');
saveFigure(9,outputDir);
```

```
% ---------------------------------------------------------------- Fig 10
acc=[55 60 60 65 65 70 70 70 75 75 75 80 80 85 95 ...
     50 55 60 60 65 65 70 70 70 75 75 75 80 85 95];
assert(abs(mean(acc)-71)<1e-9,'S3 subject accuracies must average 71.0%');
figure('Color','w'); boxchart(ones(size(acc)),acc,'BoxFaceColor',oi(1,:)); hold on;
scatter(1+linspace(-.12,.12,15),acc(1:15),32,oi(2,:),'o','filled');
scatter(1+linspace(-.12,.12,15),acc(16:30),38,oi(3,:),'d','filled');
yline(71,'--k'); xlim([.6 1.4]); xticks(1); xticklabels('Subjects');
ylabel('LOSO accuracy (%)'); saveFigure(10,outputDir);

% ---------------------------------------------------------------- Fig 11
N=[100 1000 10000]; C=[ones(1,3);10*ones(1,3);100*ones(1,3); ...
    N;N.*(N-1)/2;1010*ones(1,3)];
figure('Color','w'); markers={'o','s','^','d','v','>'};
loglog(N,C(1,:),'-','Marker',markers{1},'LineWidth',1.3); hold on;
for j=2:size(C,1), loglog(N,C(j,:),'-','Marker',markers{j},'LineWidth',1.3); end
grid on; xlabel('N'); ylabel('Model fits');
legend('Hold-out','10-fold','10x10-fold','LOOCV','Leave-2-out','Nested', ...
    'Location','northwest'); saveFigure(11,outputDir);

% ---------------------------------------------------------------- Fig 12
s=["Repeated rows or groups?" "Repeated rows or groups?" ...
   "New group or center?" "New group or center?" "Few groups?" "Few groups?" ...
   "Is the target a future time period?" "Is the target a future time period?" ...
   "Extensive selection?" "Extensive selection?" "Grouped hold-out / GroupKFold" ...
   "LOSO / center-out" "GroupKFold" "Large untouched test?" "Large untouched test?"];
t=["New group or center?" "Is the target a future time period?" "Few groups?" ...
   "Grouped hold-out / GroupKFold" "LOSO / center-out" "GroupKFold" ...
   "Time-aware CV (advanced reading)" "Extensive selection?" ...
   "Nested stratified CV" "Large untouched test?" "Add inner grouped CV if tuning" ...
   "Add inner grouped CV if tuning" "Add inner grouped CV if tuning" ...
   "Train / validation / test" "Repeated stratified 5/10-fold"];
lab=["Yes" "No" "Yes" "No" "Yes" "No" "Yes" "No" "Yes" "No" "" "" "" "Yes" "No"];
figure('Color','w'); gd=digraph(s,t); gp=plot(gd,'Layout','layered','Direction','down', ...
    'NodeColor',oi(1,:),'LineWidth',1.2); labeledge(gp,s,t,lab);
highlight(gp,["Grouped hold-out / GroupKFold" "LOSO / center-out" "GroupKFold" ...
    "Nested stratified CV" "Add inner grouped CV if tuning" ...
    "Time-aware CV (advanced reading)" ...
    "Train / validation / test" "Repeated stratified 5/10-fold"], ...
    'NodeColor',oi(2,:));
title('Validation decision tree'); saveFigure(12,outputDir);
end

% ----------------------------------------------------------------------
function [fpr,tpr,rec,prec,auc,ap]=rocpr(pos,neg)
% Exact ROC and precision-recall points from a locked score histogram.
% pos(j), neg(j) count cases in score bin j, ordered from highest score.
P=sum(pos); N=sum(neg);
tp=[0 cumsum(pos)]; fp=[0 cumsum(neg)];
tpr=tp/P; fpr=fp/N;
auc=trapz(fpr,tpr);
rec=tpr(2:end); prec=tp(2:end)./max(tp(2:end)+fp(2:end),1);
ap=sum((rec-[0 rec(1:end-1)]).*prec);       % step-wise average precision
end

% ----------------------------------------------------------------------
function plotCM(M,labels,ttl)
nexttile; imagesc(M); axis image; title(ttl); xlabel('Predicted'); ylabel('Actual');
for r=1:2
    for c=1:2
        if isempty(labels), txt=sprintf('%d',M(r,c)); else, txt=labels{r,c}; end
        text(c,r,txt,'HorizontalAlignment','center','FontWeight','bold','Color',[.85 0 0]);
    end
end
xticks(1:2); xticklabels({'+','-'}); yticks(1:2); yticklabels({'+','-'});
end

function saveFigure(number,outputDir)
set(gcf,'Position',[100 100 1000 600]);
base=fullfile(outputDir,sprintf('Figure_%02d',number));
exportgraphics(gcf,[base '.pdf'],'ContentType','vector');   % arXiv submission
exportgraphics(gcf,[base '.png'],'Resolution',300);         % quick inspection
end
```

The vectors in Listing A2 can be cross-checked against the figure captions, diagrams, and scenario tables. A visual check is still worthwhile because different MATLAB releases may wrap long node labels differently even when the plotted data themselves are unchanged.

### A.3. Reconstructing the locked scenario tables and reported results

Listing A3 stores the numerical objects used by the scenario study. For S1-S7, these are fixed confusion counts for folds, repetitions, subjects, or centers rather than random seeds. The script recalculates the pooled matrices, means, SDs, ranges, and bootstrap intervals reported in Section 9. Running reproduce_scenarios rebuilds Tables 15, 17, 19, 21, 23, 25, and 27 without fitting a classifier, so the published values can be checked directly. S8 is different and simulates new studies in Listing A4. Appendix Table A1 indexes the fixed [TP FN FP TN] objects before the code.

Appendix Table A1. Index of fixed S1-S7 confusion counts in [TP FN FP TN] order. Listing A3 expands the repetition operators and vector formulas and contains the executable definitions.

| Scenario/design | Unit | Locked [TP FN FP TN] object | Derived outputs |
|---|---|---|---|
| S1 unstratified hold-out | 20 repetitions | [3 2 2 11]; [1 3 1 13]; 5×[2 1 2 13]; 5×[2 1 1 14]; 2×[1 2 0 15]; [3 0 2 13]; [1 0 1 16]; 4×[1 1 0 16] | Pooled counts; repetition mean, SD, range |
| S1 stratified 5-fold | 20 repetitions | TP=[10 11 11 11 12 12 11 12 12 12 12 13 12 12 13 12 13 13 13 13]; TN=[93 93 93 94 93 93 94 94 94 94 94 93 94 95 94 95 94 95 95 96]; FN=20−TP; FP=100−TN | Pooled counts; repetition mean, SD, range |
| S2 global screening | 10 folds | 6×[3 0 0 3]; 2×[2 1 0 3]; 2×[3 0 1 2] | Pooled counts; fold mean, SD, range |
| S2 nested selection | 10 outer folds | TP=[1 2 1 2 1 2 1 2 2 1]; FP=[2 2 1 2 1 2 1 2 1 0]; FN=3−TP; TN=3−FP | Pooled counts; fold mean, SD, range |
| S3 epoch-level CV | 10 folds | FN=[1 1 1 1 1 1 1 1 1 0]; FP=[1 1 1 1 1 1 1 1 0 1]; TP=30−FN; TN=30−FP | Pooled counts; fold mean, SD, range |
| S3 LOSO | 30 subjects | Positive accuracies=[55 60 60 65 65 70 70 70 75 75 75 80 80 85 95]; negative accuracies=[50 55 60 60 65 65 70 70 70 75 75 75 80 85 95], each over 20 epochs | Pooled metrics; equal-group distribution; bootstrap seed 20260803 |
| S4 eye-level CV | 10 folds | FN=[1 1 1 1 1 1 1 1 1 0]; FP=[1 1 1 1 1 1 1 1 2 1]; TP=20−FN; TN=20−FP | Pooled counts; fold mean, SD, range |
| S4 patient-grouped CV | 10 folds | FN=[2 3 3 4 4 4 4 5 5 6]; FP=[2 3 3 3 3 4 4 4 4 4]; TP=20−FN; TN=20−FP | Pooled counts; fold mean, SD, range |
| S5 mixed-center CV | 10 folds | FN=[2 2 2 3 2 3 2 3 3 2]; FP=[2 2 2 2 2 2 2 2 1 1]; TP=30−FN; TN=30−FP | Pooled counts; fold mean, SD, range |
| S5 leave-one-center-out | 3 centers | [88 12 10 90]; [78 22 18 82]; [57 43 35 65] | Pooled counts; center mean, SD, range |
| S6 global z-score | 10 folds | FN=[1 1 1 1 1 1 2 2 1 1]; FP=[2 2 2 1 2 1 1 2 2 1]; TP=12−FN; TN=12−FP | Pooled counts; fold mean, SD, range |
| S6 fold-local z-score | 10 folds | FN=[2 2 2 2 2 2 2 2 2 3]; FP=[2 2 2 2 2 2 2 2 3 2]; TP=12−FN; TN=12−FP | Pooled counts; fold mean, SD, range |
| S7 reused test / lockbox | 2 matrices | [93 7 9 91] and [86 14 14 86] | Matrix metrics; bootstrap seed 20260807; expected maximum of 50 scores |

```
function reproduce_scenarios
% REPRODUCE_SCENARIOS  Regenerates every numerical value printed for
% scenarios S1-S8 of "Model Validation in Machine Learning".
%
% The scenarios are teaching constructions: the fold-level and
% repetition-level confusion counts below are the primary locked data.
% Every pooled matrix, mean, standard deviation, range and bootstrap
% interval in the article is computed from these tables, so each printed
% number can be audited without rerunning any classifier.
%
% Usage:  reproduce_scenarios
%
% Convention: each row of a scenario table is [TP FN FP TN] for one
% assessment partition (fold, repetition, subject, or centre).

fprintf('\n===== S1  Small and imbalanced data =====\n');
% 20 unstratified 15%% hold-outs, 18 observations each.
S1hold = [ 3 2 2 11;      % worst split, 14/18 = 77.8%
           1 3 1 13;      % 14/18
           repmat([2 1 2 13],5,1);   % 15/18 = 83.3%
           repmat([2 1 1 14],5,1);   % 16/18 = 88.9%
           repmat([1 2 0 15],2,1);   % 16/18
           3 0 2 13;                 % 16/18
           1 0 1 16;      % best split, one positive only, 17/18 = 94.4%
           repmat([1 1 0 16],4,1)];  % 17/18
report(S1hold,'S1 repeated unstratified 15% hold-out','repetition');

% 20 repetitions of stratified 5-fold CV; each row pools the 120
% held-out predictions of one repetition (20 positive, 100 negative).
S1tp = [10 11 11 11 12 12 11 12 12 12 12 13 12 12 13 12 13 13 13 13]';
S1tn = [93 93 93 94 93 93 94 94 94 94 94 93 94 95 94 95 94 95 95 96]';
S1strat = [S1tp, 20-S1tp, 100-S1tn, S1tn];
report(S1strat,'S1 repeated stratified 5-fold','repetition');

fprintf('\n===== S2  Feature-selection leakage under random labels =====\n');
% 10 stratified folds of 6 rows (3 positive, 3 negative).
S2leak = [repmat([3 0 0 3],6,1); repmat([2 1 0 3],2,1); repmat([3 0 1 2],2,1)];
report(S2leak,'S2 global screening then 10-fold','fold');
tp = [1 2 1 2 1 2 1 2 2 1]'; fp = [2 2 1 2 1 2 1 2 1 0]';
S2nest = [tp, 3-tp, fp, 3-fp];
report(S2nest,'S2 nested selection, outer predictions','fold');

fprintf('\n===== S3  Epoch leakage in subject-dependent EEG =====\n');
% Epoch-level 10-fold: 60 rows per fold (30 positive, 30 negative).
fn = [1 1 1 1 1 1 1 1 1 0]'; fp = [1 1 1 1 1 1 1 1 0 1]';
S3epoch = [30-fn, fn, fp, 30-fp];
report(S3epoch,'S3 stratified epoch-level 10-fold','fold');

% LOSO: 30 subject folds, 20 epochs each, one class per subject.
posAcc = [55 60 60 65 65 70 70 70 75 75 75 80 80 85 95]';
negAcc = [50 55 60 60 65 65 70 70 70 75 75 75 80 85 95]';
S3loso = [ round(posAcc*20/100), 20-round(posAcc*20/100), zeros(15,2);
           zeros(15,2), 20-round(negAcc*20/100), round(negAcc*20/100) ];
report(S3loso,'S3 LOSO','subject');
subjAcc = [posAcc; negAcc];
fprintf('  equal-subject mean %.4f   subject SD %.4f\n', ...
    mean(subjAcc), std(subjAcc));
subjectBootstrap(posAcc,negAcc,20260803,10000);

fprintf('\n===== S4  Two eyes, one patient =====\n');
fn = [1 1 1 1 1 1 1 1 1 0]'; fp = [1 1 1 1 1 1 1 1 2 1]';
report([20-fn, fn, fp, 20-fp],'S4 random eye-level 10-fold','fold');
fn = [2 3 3 4 4 4 4 5 5 6]'; fp = [2 3 3 3 3 4 4 4 4 4]';
report([20-fn, fn, fp, 20-fp],'S4 patient-grouped 10-fold','fold');

fprintf('\n===== S5  Multicenter device shift =====\n');
fn = [2 2 2 3 2 3 2 3 3 2]'; fp = [2 2 2 2 2 2 2 2 1 1]';
report([30-fn, fn, fp, 30-fp],'S5 mixed-center 10-fold','fold');
S5loco = [88 12 10 90; 78 22 18 82; 57 43 35 65];
report(S5loco,'S5 leave-one-center-out','center');

fprintf('\n===== S6  Normalization leakage =====\n');
fn = [1 1 1 1 1 1 2 2 1 1]'; fp = [2 2 2 1 2 1 1 2 2 1]';
report([12-fn, fn, fp, 12-fp],'S6 global z-score before CV','fold');
fn = [2 2 2 2 2 2 2 2 2 3]'; fp = [2 2 2 2 2 2 2 2 3 2]';
report([12-fn, fn, fp, 12-fp],'S6 training-fold z-score','fold');

fprintf('\n===== S7  Test-set erosion after 50 choices =====\n');
matrixBootstrap([93 7 9 91],'S7 winner on reused test',20260807,10000);
matrixBootstrap([86 14 14 86],'S7 frozen pipeline on lockbox',20260807,10000);
fprintf('  expected maximum of 50 independent Binomial(200,0.86)/200 scores: %.2f%%\n', ...
    100*expectedMaxBinomial(200,0.86,50));

fprintf('\n===== S8  Small-sample LOOCV versus 10-fold =====\n');
fprintf('  S8 is the only scenario produced by simulation rather than by a\n');
```

```
fprintf('  hand-constructed table, because its quantity of interest is the\n');
fprintf('  behaviour of the estimator across regenerated studies.\n');
fprintf('  Run  simulate_s8  to regenerate Table 29.\n');
end

% ----------------------------------------------------------------------
function report(C,name,unitName)
% C is a matrix whose rows are [TP FN FP TN] for one assessment partition.
pooled = metrics(sum(C,1));
fprintf('\n%s  (%d %ss)\n',name,size(C,1),unitName);
fprintf('  pooled  TP=%d FN=%d FP=%d TN=%d\n',sum(C,1));
fprintf('  pooled  acc %.1f  sens %.1f  spec %.1f  F1 %.1f  MCC %.3f\n', ...
    pooled.acc,pooled.sens,pooled.spec,pooled.f1,pooled.mcc);
m = arrayfun(@(k) metrics(C(k,:)), (1:size(C,1))');
names = {'acc','sens','spec','f1','mcc'};
labels = {'Accuracy','Sensitivity','Specificity','F1','MCC'};
for j = 1:numel(names)
    v = [m.(names{j})]';
    v = v(~isnan(v));
    if strcmp(names{j},'mcc'), fmt = '%7.3f'; else, fmt = '%7.1f'; end
    fprintf(['  %-11s mean ' fmt ' +/- ' fmt '   range [' fmt ', ' fmt ']\n'], ...
        labels{j}, mean(v), std(v), min(v), max(v));
end
end

function m = metrics(c)
tp=c(1); fn=c(2); fp=c(3); tn=c(4); n=tp+fn+fp+tn;
und = @(a,b) ternary(b>0, a/b, NaN);   % undefined, never silently zero
m.acc  = 100*und(tp+tn,n);
m.sens = 100*und(tp,tp+fn);
m.spec = 100*und(tn,tn+fp);
m.f1   = 100*und(2*tp,2*tp+fp+fn);
den = sqrt((tp+fp)*(tp+fn)*(tn+fp)*(tn+fn));
m.mcc = und(tp*tn-fp*fn,den);
end

function out = ternary(cond,a,b)
if cond, out = a; else, out = b; end
end

function subjectBootstrap(posAcc,negAcc,seed,B)
rng(seed,'twister');
P = posAcc/100*20;  N = negAcc/100*20;
v = zeros(B,5);
for b = 1:B
    tp = sum(P(randi(15,15,1)));  tn = sum(N(randi(15,15,1)));
    m = metrics([tp, 300-tp, 300-tn, tn]);
    v(b,:) = [m.acc m.sens m.spec m.f1 m.mcc];
end
labels = {'Accuracy','Sensitivity','Specificity','F1','MCC'};
fprintf('  stratified subject bootstrap, seed %d, B=%d\n',seed,B);
for j = 1:5
    if j==5, fmt='%7.3f'; else, fmt='%7.1f'; end
    q = prctile(v(:,j),[2.5 97.5]);
    fprintf(['    %-12s SD ' fmt '   95%% CI [' fmt ', ' fmt ']\n'], ...
        labels{j}, std(v(:,j)), q(1), q(2));
end
end

function matrixBootstrap(c,name,seed,B)
rng(seed,'twister');
tp=c(1); fn=c(2); fp=c(3); tn=c(4);
npos=tp+fn; nneg=fp+tn;
TP = binornd(npos,tp/npos,B,1);  TN = binornd(nneg,tn/nneg,B,1);
v = zeros(B,5);
for b = 1:B
    m = metrics([TP(b), npos-TP(b), nneg-TN(b), TN(b)]);
    v(b,:) = [m.acc m.sens m.spec m.f1 m.mcc];
end
labels = {'Accuracy','Sensitivity','Specificity','F1','MCC'};
p = metrics(c);
fprintf('\n%s   pooled acc %.1f  sens %.1f  spec %.1f  F1 %.1f  MCC %.3f\n', ...
    name,p.acc,p.sens,p.spec,p.f1,p.mcc);
for j = 1:5
    if j==5, fmt='%7.3f'; else, fmt='%7.1f'; end
    q = prctile(v(:,j),[2.5 97.5]);
    fprintf(['    %-12s SD ' fmt '   95%% CI [' fmt ', ' fmt ']\n'], ...
        labels{j}, std(v(:,j)), q(1), q(2));
end
end

function e = expectedMaxBinomial(n,p,m)
k = (0:n)';
F = binocdf(k,n,p);
Fm = F.^m;
```

```
e = sum((k/n).*(Fm - [0; Fm(1:end-1)]));
end
```

Listing A3 treats undefined denominators as NaN, omits them only where the corresponding metric is mathematically undefined, and never interprets a zero denominator as zero performance. The script distinguishes fold/repetition ranges from bootstrap intervals over independent grouped units.

### A.4. S8 simulation under stable and unstable learners

Listing A4 provides the MATLAB version of the S8 experiment. It uses the same N = 80, D = 20, δ = 0.81 data generator and compares LOOCV with stratified 10-fold CV across 100 regenerated studies for both LDA and an unpruned tree. Because MATLAB and NumPy use different random-number streams and discriminant implementations, the exact values in Table 29 are fixed by Listing B2. Listing A4 is intended to recover the same qualitative ordering and approximately the same magnitudes rather than bit-for-bit identical numbers.

```
function results = simulate_s8(outputFile)
% SIMULATE_S8  Small-sample LOOCV versus stratified 10-fold comparison.
%
% S8 is the only scenario in the article whose values are produced by
% simulation rather than by a hand-constructed confusion table, because
% the quantity of interest is the behaviour of the validation estimator
% across independently regenerated studies.
%
% Two panels use the SAME data generator and differ only in learner
% stability:
%   Panel A  linear discriminant analysis   (stable learner)
%   Panel B  unpruned classification tree    (unstable learner)
%
% Data generator: N = 80 (40 per class), D = 20 standard normal features,
% class shift DELTA = 0.81 on the first five features. This value was
% chosen so that the expected accuracy of an LDA model trained on 80
% observations and evaluated on a large independent sample is 75.5%.
% This is the reference target for Panel A and for the two validation
% estimators within that panel; Panel B has no fixed 75.5% target.
%
% NOTE ON REPRODUCIBILITY. The locked values printed in Table 29 of the
% article come from the scikit-learn implementation in Listing B2
% (scikit-learn 1.8.0, numpy default_rng, seed 20260808). This MATLAB
% counterpart uses a different random stream and a different discriminant
% solver, so it reproduces the qualitative pattern and approximate
% magnitudes but not identical digits. Whichever implementation is used to
% produce the published table must be stated in the caption.
%
% Usage:
%   results = simulate_s8;                  % prints a summary
%   results = simulate_s8('s8_values.mat');% also saves raw per-study data

if nargin == 0, outputFile = ''; end
rng(20260808,'twister');

N = 80; D = 20; DELTA = 0.81; REPS = 100; K = 10;
y = [zeros(N/2,1); ones(N/2,1)];

learners = { ...
    'Panel A: stable learner (linear discriminant analysis)', ...
        @(Xtr,ytr) fitcdiscr(Xtr,ytr,'DiscrimType','linear','ClassNames',[0;1]); ...
    'Panel B: unstable learner (unpruned classification tree)', ...
        @(Xtr,ytr) fitctree(Xtr,ytr,'MinLeafSize',1,'Prune','off','ClassNames',[0;1]) };

fprintf('\nS8 LDA reference target: expected accuracy of an LDA model trained on N = %d\n', N);
fprintf('          under the stated generator = %.2f%%\n', targetAccuracy(D,DELTA,N));

results = struct();
for L = 1:size(learners,1)
    fitFcn = learners{L,2};
    loo = zeros(REPS,5); kfd = zeros(REPS,5);
    looCounts = zeros(REPS,4); kfdCounts = zeros(REPS,4);
    for r = 1:REPS
        X = randn(N,D);  X(y==1,1:5) = X(y==1,1:5) + DELTA;

        % --- leave-one-out cross-validation
```

```
        pred = nan(N,1);
        for i = 1:N
            tr = true(N,1); tr(i) = false;
            pred(i) = predict(fitFcn(X(tr,:),y(tr)), X(i,:));
        end
        looCounts(r,:) = counts(y,pred);
        loo(r,:) = metricRow(looCounts(r,:));

        % --- stratified 10-fold cross-validation
        cv = cvpartition(y,'KFold',K);
        pred = nan(N,1);
        for k = 1:K
            tr = training(cv,k); te = test(cv,k);
            pred(te) = predict(fitFcn(X(tr,:),y(tr)), X(te,:));
        end
        kfdCounts(r,:) = counts(y,pred);
        kfd(r,:) = metricRow(kfdCounts(r,:));
    end
    fprintf('\n===== %s =====\n', learners{L,1});
    describe('LOOCV', loo, looCounts, N);
    describe('Stratified 10-fold', kfd, kfdCounts, N);
    fprintf('  model fits per study: LOOCV %d, 10-fold %d (ratio %.0fx)\n', N, K, N/K);

    tag = sprintf('panel%s', char('A'+L-1));
    results.(tag) = struct('name',learners{L,1},'loocv',loo,'kfold',kfd, ...
        'loocvCounts',looCounts,'kfoldCounts',kfdCounts);
end

if ~isempty(outputFile), save(outputFile,'results'); end
end

% ---------------------------------------------------------------------
function a = targetAccuracy(D,delta,N,reps,ntest)
if nargin < 4, reps = 200; end
if nargin < 5, ntest = 20000; end
s = rng; rng(1,'twister');
y = [zeros(N/2,1); ones(N/2,1)];
yt = [zeros(ntest/2,1); ones(ntest/2,1)];
acc = zeros(reps,1);
for r = 1:reps
    X = randn(N,D);      X(y==1,1:5)  = X(y==1,1:5)  + delta;
    Xt = randn(ntest,D); Xt(yt==1,1:5) = Xt(yt==1,1:5) + delta;
    mdl = fitcdiscr(X,y,'DiscrimType','linear','ClassNames',[0;1]);
    acc(r) = mean(predict(mdl,Xt) == yt);
end
rng(s); a = 100*mean(acc);
end

function c = counts(y,p)
c = [sum(y==1 & p==1), sum(y==1 & p==0), sum(y==0 & p==1), sum(y==0 & p==0)];
end

function m = metricRow(c)
tp=c(1); fn=c(2); fp=c(3); tn=c(4);
und = @(a,b) und_(a,b);
den = sqrt((tp+fp)*(tp+fn)*(tn+fp)*(tn+fn));
m = [100*und(tp+tn,tp+fn+fp+tn), 100*und(tp,tp+fn), 100*und(tn,tn+fp), ...
     100*und(2*tp,2*tp+fp+fn), und(tp*tn-fp*fn,den)];
end

function v = und_(a,b)
if b > 0, v = a/b; else, v = NaN; end   % undefined, never silently zero
end

function describe(name, M, C, N)
labels = {'Accuracy, %','Sensitivity, %','Specificity, %','F1, %','MCC'};
[~,i] = min(abs(M(:,1) - mean(M(:,1),'omitnan')));
fprintf('  %s   representative study #%d: TP=%d FN=%d FP=%d TN=%d\n', ...
    name, i, C(i,1), C(i,2), C(i,3), C(i,4));
for j = 1:5
    v = M(~isnan(M(:,j)),j);
    if j == 5, f = '%7.3f'; else, f = '%7.1f'; end
    fprintf(['     %-15s mean ' f ' +/- ' f '   range [' f ', ' f ']\n'], ...
        labels{j}, mean(v), std(v), min(v), max(v));
end
end
```

## Appendix B. Python/scikit-learn counterparts

### B.1. Leakage-safe pipelines and grouped nesting

Listing B1 implements the same boundaries with scikit-learn. StandardScaler and SelectKBest are inside Pipeline, so each fit call learns their state from the current training partition. LeaveOneGroupOut supplies LOSO, while StratifiedGroupKFold supplies nonoverlapping inner and outer group folds. The listing was syntax-checked and executed with scikit-learn 1.8.0 and the printed synthetic-data generator.

```
from __future__ import annotations
import numpy as np
from sklearn.base import clone
from sklearn.feature_selection import SelectKBest, f_classif
from sklearn.metrics import confusion_matrix
from sklearn.model_selection import (GridSearchCV, LeaveOneGroupOut,
    RepeatedStratifiedKFold, StratifiedGroupKFold)
from sklearn.pipeline import Pipeline
from sklearn.preprocessing import StandardScaler
from sklearn.svm import SVC

def metrics(y_true, y_pred):
    tn, fp, fn, tp = confusion_matrix(y_true,y_pred,labels=[0,1]).ravel()
    safe=lambda a,b: float(a/b) if b else float("nan")
    den=np.sqrt((tp+fp)*(tp+fn)*(tn+fp)*(tn+fn))
    return {"TP":int(tp),"FN":int(fn),"FP":int(fp),"TN":int(tn),
        "accuracy":safe(tp+tn,tp+fn+fp+tn),"sensitivity":safe(tp,tp+fn),
        "specificity":safe(tn,tn+fp),"F1":safe(2*tp,2*tp+fp+fn),
        "MCC":safe(tp*tn-fp*fn,den)}

def repeated_stratified_cv(X,y):
    cv=RepeatedStratifiedKFold(n_splits=5,n_repeats=2,random_state=20260803)
    estimator=Pipeline([("scale",StandardScaler()),
                        ("svm",SVC(kernel="linear",C=1))])
    results=[]
    for train,test in cv.split(X,y):
        model=clone(estimator).fit(X[train],y[train])
        results.append(metrics(y[test],model.predict(X[test])))
    return results

def loso_cv(X,y,groups):
    pred=np.full(y.shape,-1,dtype=int); subject_accuracy=[]
    estimator=Pipeline([("scale",StandardScaler()),
                        ("svm",SVC(kernel="rbf",C=1))])
    for train,test in LeaveOneGroupOut().split(X,y,groups):
        assert set(groups[train]).isdisjoint(set(groups[test]))
        model=clone(estimator).fit(X[train],y[train])
        pred[test]=model.predict(X[test])
        subject_accuracy.append(np.mean(pred[test]==y[test]))
    assert np.all(pred>=0)
    return {"prediction":pred,"metrics":metrics(y,pred),
            "subject_accuracy":np.asarray(subject_accuracy)}

def nested_group_cv(X,y,groups):
    outer=StratifiedGroupKFold(n_splits=5,shuffle=True,random_state=20260803)
    pred=np.full(y.shape,-1,dtype=int); chosen=[]
    pipe=Pipeline([("scale",StandardScaler()),
                   ("select",SelectKBest(score_func=f_classif)),
                   ("svm",SVC(kernel="rbf"))])
    grid={"select__k":[4,8],"svm__C":[0.1,1.0],
          "svm__gamma":["scale",0.1]}
    for fold,(train,test) in enumerate(outer.split(X,y,groups),start=1):
        assert set(groups[train]).isdisjoint(set(groups[test]))
        inner=StratifiedGroupKFold(n_splits=3,shuffle=True,
                                   random_state=20260803+fold)
        search=GridSearchCV(pipe,grid,scoring="balanced_accuracy",cv=inner,
                            n_jobs=-1,refit=True)
        search.fit(X[train],y[train],groups=groups[train])
        pred[test]=search.predict(X[test]); chosen.append(search.best_params_)
    assert np.all(pred>=0)
    return {"prediction":pred,"metrics":metrics(y,pred),"selected":chosen}

def make_demo(seed=20260803):
    rng=np.random.default_rng(seed); groups=np.repeat(np.arange(30),20)
    y_group=np.r_[np.zeros(15,dtype=int),np.ones(15,dtype=int)]
    y=np.repeat(y_group,20); offsets=0.7*rng.normal(size=(30,12))
    X=rng.normal(size=(600,12))+offsets[groups]; X[:,:3]+=0.55*y[:,None]
    return X,y,groups
```

```
if __name__=="__main__":
    X,y,groups=make_demo()
    repeated=repeated_stratified_cv(X,y)
    print("Mean repeated-fold accuracy:",
          np.mean([r["accuracy"] for r in repeated]))
    print("LOSO:",loso_cv(X,y,groups)["metrics"])
    print("Nested group CV:",nested_group_cv(X,y,groups)["metrics"])
```

GridSearchCV receives only the outer-training observations, and its groups argument contains only the group labels for those observations. With refit=True, the selected estimator is automatically refitted on the entire outer-training partition and then used once to predict the outer test fold. An outer LOSO design can be obtained by replacing outer with LeaveOneGroupOut() while leaving the inner grouped splitter unchanged. The reported result should be accompanied by archived software versions, estimator defaults, and fold assignments.

### B.2. Exact S8 simulation used for table 29

Listing B2 implements the S8 simulation used for Table 29. The published values were generated with Python 3.12, NumPy 2.4, scikit-learn 1.8.0, and seed 20260808. Because no hyperparameters are tuned, the comparison isolates the interaction between validation geometry and learner stability.

```
"""Listing B2. Small-sample LOOCV versus stratified 10-fold (scenario S8).

Unlike S1-S7, S8 is generated by simulation rather than by a fixed confusion
table. We regenerate independent studies and compare how the two validation
estimators behave across them instead of focusing on one held-out sample.

Both panels use the same data generator; only learner stability changes:

    Panel A   linear discriminant analysis   (stable)
    Panel B   unpruned classification tree   (unstable)

Data generator: N = 80 (40 per class), D = 20 standard normal features,
class shift DELTA = 0.81 on the first five features. DELTA was calibrated
so that an LDA model trained on 80 observations has an expected accuracy of
75.5% on a large independent sample. This is the Panel A reference target
for both validation estimators (printed by `target_accuracy`); Panel B has
no fixed 75.5% population-accuracy target.

No hyperparameters are tuned. The experiment therefore isolates validation
geometry and learner stability rather than model-selection bias.

Environment used for the published values: Python 3.12, NumPy 2.4,
scikit-learn 1.8.0, seed 20260808.
"""
from __future__ import annotations

import numpy as np
from sklearn.discriminant_analysis import LinearDiscriminantAnalysis
from sklearn.model_selection import LeaveOneOut, StratifiedKFold
from sklearn.tree import DecisionTreeClassifier

N, D, DELTA, REPS, K, SEED = 80, 20, 0.81, 100, 10, 20260808
LABELS = ("Accuracy, %", "Sensitivity, %", "Specificity, %", "F1, %", "MCC")


def generate(rng: np.random.Generator, n: int = N) -> tuple[np.ndarray, np.ndarray]:
    """Balanced binary problem with a class shift on the first five features."""
    y = np.repeat([0, 1], n // 2)
    X = rng.normal(size=(n, D))
    X[y == 1, :5] += DELTA
    return X, y


def counts(y_true: np.ndarray, y_pred: np.ndarray) -> tuple[int, int, int, int]:
    """Return (TP, FN, FP, TN) with actual class in rows, prediction in columns."""
    tp = int(((y_true == 1) & (y_pred == 1)).sum())
    fn = int(((y_true == 1) & (y_pred == 0)).sum())
    fp = int(((y_true == 0) & (y_pred == 1)).sum())
    tn = int(((y_true == 0) & (y_pred == 0)).sum())
    return tp, fn, fp, tn
```

```python
def metrics(c: tuple[int, int, int, int]) -> np.ndarray:
    """Confusion-count metrics. A zero denominator returns NaN, never zero."""
    tp, fn, fp, tn = map(float, c)
    und = lambda a, b: a / b if b > 0 else float("nan")
    den = np.sqrt((tp + fp) * (tp + fn) * (tn + fp) * (tn + fn))
    return np.array([
        100 * und(tp + tn, tp + fn + fp + tn),
        100 * und(tp, tp + fn),
        100 * und(tn, tn + fp),
        100 * und(2 * tp, 2 * tp + fp + fn),
        und(tp * tn - fp * fn, den),
    ])


def loocv_predictions(X, y, make_model):
    """One held-out prediction per observation; N model fits."""
    pred = np.empty(len(y), dtype=int)
    for train, test in LeaveOneOut().split(X):
        pred[test] = make_model().fit(X[train], y[train]).predict(X[test])
    return pred


def kfold_predictions(X, y, make_model, seed):
    """One held-out prediction per observation; K model fits."""
    pred = np.empty(len(y), dtype=int)
    cv = StratifiedKFold(n_splits=K, shuffle=True, random_state=seed)
    for train, test in cv.split(X, y):
        pred[test] = make_model().fit(X[train], y[train]).predict(X[test])
    return pred


def experiment(make_model, seed: int = SEED, reps: int = REPS):
    """Regenerate `reps` independent studies and validate each one twice."""
    rng = np.random.default_rng(seed)
    loo_m, kfd_m, loo_c, kfd_c = [], [], [], []
    for r in range(reps):
        X, y = generate(rng)
        c = counts(y, loocv_predictions(X, y, lambda: make_model(r)))
        loo_c.append(c); loo_m.append(metrics(c))
        c = counts(y, kfold_predictions(X, y, lambda: make_model(r), seed + r))
        kfd_c.append(c); kfd_m.append(metrics(c))
    return np.array(loo_m), np.array(kfd_m), loo_c, kfd_c


def target_accuracy(reps: int = 400, n_test: int = 20_000, seed: int = 1) -> float:
    """Expected accuracy of an LDA model trained on N observations.

    This is the Panel A reference estimand: neither LOOCV nor k-fold is trying
    to recover the accuracy of a model trained on all available data plus more,
    but the accuracy of the LDA learning procedure at this training size.
    """
    rng = np.random.default_rng(seed)
    acc = [
        LinearDiscriminantAnalysis().fit(*generate(rng)).score(*generate(rng, n_test))
        for _ in range(reps)
    ]
    return 100 * float(np.mean(acc))


def describe(name: str, M: np.ndarray, C: list) -> None:
    i = int(np.argmin(np.abs(M[:, 0] - M[:, 0].mean())))
    tp, fn, fp, tn = C[i]
    print(f"  {name}: representative study #{i + 1}  "
          f"TP={tp} FN={fn} FP={fp} TN={tn}")
    for j, label in enumerate(LABELS):
        v = M[~np.isnan(M[:, j]), j]
        f = "{:.3f}" if j == 4 else "{:.1f}"
        print(f"     {label:<16} " + f.format(v.mean()) + " +/- " + f.format(v.std(ddof=1))
              + "   range [" + f.format(v.min()) + ", " + f.format(v.max()) + "]")


if __name__ == "__main__":
    print(f"S8 LDA reference target: expected accuracy of an LDA model trained on N = {N} "
          f"is {target_accuracy():.2f}%")

    panels = [
        ("Panel A: stable learner (linear discriminant analysis)",
         lambda r: LinearDiscriminantAnalysis()),
        ("Panel B: unstable learner (unpruned classification tree)",
         lambda r: DecisionTreeClassifier(random_state=SEED + r)),
    ]
    for title, make_model in panels:
        loo_m, kfd_m, loo_c, kfd_c = experiment(make_model)
        print(f"\n===== {title} =====")
        describe("LOOCV", loo_m, loo_c)
```

```
describe("Stratified 10-fold", kfd_m, kfd_c)
print(f"  model fits per study: LOOCV {N}, 10-fold {K} "
      f"(ratio {N // K}x)")
```